\documentclass[letterpaper]{article} 
\usepackage{aaai2027}  
\usepackage[hyphens]{url}  
\usepackage{graphicx} 
\usepackage{natbib}  
\usepackage{caption} 
\usepackage{algorithm}
\usepackage{algorithmic}
\usepackage{caption}      
\usepackage{subcaption}   
\usepackage{newfloat}
\usepackage{listings}
\DeclareCaptionStyle{ruled}{labelfont=normalfont,labelsep=colon,strut=off} 
\floatstyle{ruled}
\newfloat{listing}{tb}{lst}{}
\floatname{listing}{Listing}

\usepackage{booktabs}

\usepackage{amsmath}
\usepackage{amsthm}
\usepackage{amssymb}
\usepackage{booktabs}
\usepackage{multirow}
\usepackage{algorithm}
\usepackage{algorithmic}
\usepackage[table]{xcolor}
\definecolor{evagreen}{RGB}{46,125,50}
\definecolor{evablue}{RGB}{52,120,190}
\definecolor{evagold}{RGB}{212,175,55}
\newtheorem{proposition}{Proposition}
\newtheorem{corollary}{Corollary}
\newcommand{\indicator}{\mathbb{I}}
\usepackage[colorlinks,breaklinks,
  linkcolor=black,citecolor=black,urlcolor=blue]{hyperref}
\newcommand{\method}{ABM-LoRA}
\newcommand{\blfootnote}[1]{%
  \begingroup
  \renewcommand\thefootnote{}\footnote{#1}%
  \addtocounter{footnote}{-1}%
  \endgroup
}
 
\title{Activation Boundary Matching: Task-Informed Initialization for \\ Low-Rank Adaptation}
 
\author{
    Dongha Lee\textsuperscript{*1}\quad
    Jinhee Park\textsuperscript{*1,2}\quad
    Minjun Kim\textsuperscript{1}\quad
    Junseok Kwon\textsuperscript{$\dagger$1}
}
\affiliations{
    \textsuperscript{1}Chung-Ang University \quad
    \textsuperscript{2}Korea Electronics Technology Institute (KETI)\\
    \{ia06073, iv4084em, ekdh635, jskwon\}@cau.ac.kr\\
    Code: \url{https://github.com/lee-research/ABM-LORA}
}

\nocopyright
\begin{document}
 
\maketitle

\blfootnote{$^{*}$Equal contribution. $^{\dagger}$Corresponding author.}

\begin{abstract}
Low-Rank Adaptation (LoRA) is highly sensitive to initialization, yet
existing schemes construct the initial subspace from statistics at the
pretrained point, capturing pre-adaptation geometry rather than how the
adapter must move during learning. We examine the early adaptation
trajectory and uncover a temporal asymmetry: task-induced activation
\emph{boundaries}---the signs of layer-wise pre-activations---recover
markedly faster than activation values or effective low-rank updates and
become reusable well before the probe adapter converges. Motivated by
this, we propose Activation Boundary Matching for LoRA (\method{}), which
briefly trains a standard adapter as an \emph{early task probe}, then uses
its pre-activation signs as layer-wise targets for a fresh adapter under a
margin-based hinge objective. Both are discarded before otherwise
unchanged downstream fine-tuning. We show that this
\emph{boundary-supervision lifting} can expose LoRA directions attenuated
or locally unobservable through the downstream Jacobian, and that what
transfers is the boundary side---not the exact activation value or the
output-level signal. Because a partial boundary trace suffices,
\method{} recovers most of the benefit of a fully trained reference at a
fraction of the overhead, requiring only probe forward passes. It improves over standard LoRA on T5-base/GLUE, ConvNeXt-T and Swin-T fine-grained classification, and instruction tuning with Qwen2.5-1.5B and LLaMA2-7B, and matches or surpasses SVD- and gradient-based initializers without their preprocessing.
\end{abstract}

\section{Introduction}
\label{sec:intro}

Large-scale pretrained transformers such as
BERT~\cite{devlin2019bert}, GPT-3~\cite{brown2020language},
T5~\cite{raffel2020exploring}, and Vision Transformer
(ViT)~\cite{dosovitskiy2020image} have become a dominant paradigm for
transfer learning in language and vision.
Despite their strong general-purpose representations, fully fine-tuning these
increasingly large models requires substantial computational and storage
costs.
Low-Rank Adaptation (LoRA)~\cite{hu2022lora} addresses this challenge by
freezing each pretrained weight $W_0\in\mathbb{R}^{d\times k}$ and learning
a low-rank update $BA$, where
$A\in\mathbb{R}^{r\times k}$,
$B\in\mathbb{R}^{d\times r}$, and
$\operatorname{rank}(BA)\leq r$.
This reduces the number of trainable parameters from
$\mathcal{O}(dk)$ to $\mathcal{O}(r(d+k))$.

This efficiency, however, makes LoRA optimization particularly sensitive to
initialization.
Under the standard scheme, $A_0$ is randomly initialized and $B_0$ is set to
zero, ensuring that the initial model exactly recovers the pretrained
function.
The first effective update is consequently restricted by the random
low-rank factors and may filter out task-relevant gradient components.
LoRA initialization therefore determines not only the initial parameter
values but also the update directions accessible during the critical early
stage of adaptation.
Developing task-aware initializations is thus important for improving the
optimization and downstream performance of LoRA.

\begin{figure}[t]
    \centering

    \includegraphics[
        width=\columnwidth,
        trim=0 0 0 0,
        clip
    ]{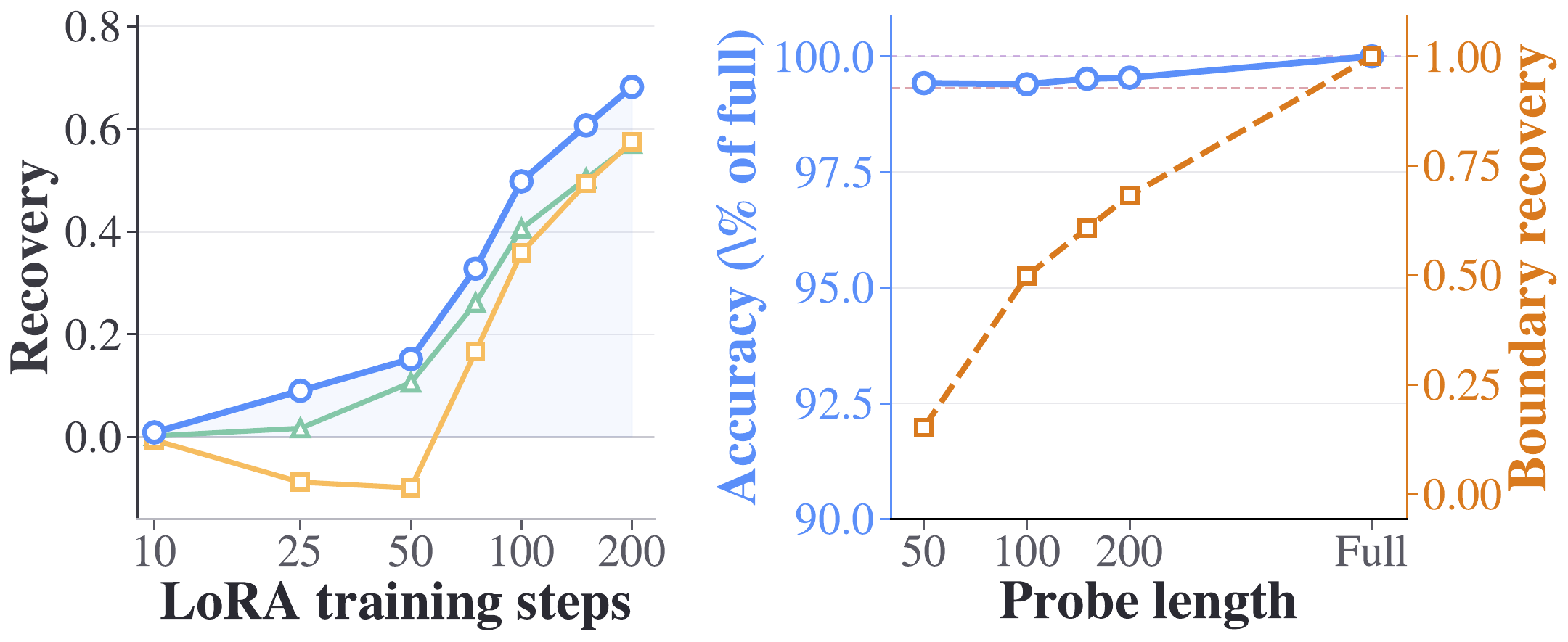}
    \vspace{-7mm}
\caption{            
    \textbf{Partial early boundaries are sufficient for effective LoRA
    initialization.}
    \textbf{(a)} During standard LoRA fine-tuning on MNLI,
    activation-boundary (\textbf{blue}) recovery advances ahead of activation-value (\textbf{yellow}) and
    effective-update (\textbf{green}) recovery, reaching $0.50$ at 100 steps compared with
    $0.36$ and $0.41$, respectively.
    \textbf{(b)} We initialize fresh adapters from different probe checkpoints
    using ABM.
    A 50-step probe attains a boundary-recovery (\textbf{orange}) F1 of only $0.15$ but already
    achieves $99.4\%$ of full-reference ABM accuracy (\textbf{blue}) at a total optimization-step budget of only \textbf{1.11}$\times$ that of standard LoRA, whereas full-reference ABM requires $2.01\times$.
    Details are provided in Appendix~\ref{app:boundary_dynamics}.
}
    \label{fig:motivation}
    \vspace{-5mm}
\end{figure}

Motivated by this sensitivity, recent studies have developed increasingly
sophisticated LoRA initialization strategies.
Weight-aware methods such as PiSSA~\cite{meng2024pissa} and
OLoRA~\cite{buyukakyuz2024olora} construct low-rank factors from the spectral
structure of pretrained weights, while gradient-aware approaches such as
LoRA-GA~\cite{wang2024lora} align the initial update with task gradients to
approximate the early dynamics of full fine-tuning.
Other approaches exploit data-dependent activation statistics, rescale the
initial factors, or correct gradient misalignment
~\cite{chen2025hrp,das2025consnotrainlora,wang2025activation,li2025beyond}.
Together, these methods demonstrate that incorporating task-relevant structure
into the initial low-rank space can substantially improve LoRA optimization.
Despite these advances, initialization is typically constructed from
information evaluated at or near the pretrained starting point.
This leaves the early adaptation trajectory itself largely unexplored as a
source of initialization information.
Although training a fully converged reference adapter would be costly, a brief
interaction with the downstream task may already reveal functional structure
unavailable at the pretrained point.
We therefore ask:
\emph{can the early adaptation trajectory expose partial functional structure
that is already sufficient to initialize a fresh LoRA adapter?}

To answer this question, we examine the training trajectory of standard LoRA
without applying any specialized initialization.
Our analysis reveals two complementary properties of task-induced activation
boundaries.
First, task-induced activation boundaries develop ahead of continuous
numerical quantities.
We measure boundary recovery as the F1 agreement between the current and final
signs over coordinates whose signs change between the pretrained model and
the final adapter.
Activation-value and effective-update recovery are normalized such that their
respective final states equal one; complete definitions are provided in
Appendix~\ref{app:boundary_dynamics}.
As shown in Fig.\ref{fig:motivation}(a), after 100 optimization steps on
MNLI, boundary recovery reaches $0.50$, whereas effective-update and
activation-value recovery reach only $0.41$ and $0.36$, respectively.
At earlier checkpoints, activation-value recovery can even become negative,
indicating that numerical activations temporarily move away from their final
values while task-induced sign structure has already begun to emerge.

Second, initialization utility emerges substantially earlier than complete
boundary recovery.
As shown in Fig.\ref{fig:motivation}(b), a 50-step probe achieves a
boundary-recovery F1 of only $0.15$, yet an adapter initialized from this
partial trace already attains $99.4\%$ of the accuracy obtained by
full-reference ABM.
It requires $1.11\times$ the cost of standard LoRA, compared with
$2.01\times$ for the full-reference setting.
Increasing the probe length continues to improve boundary recovery but yields
little additional initialization benefit.
Thus, a partial task-conditioned boundary trace becomes reusable while the
underlying numerical solution remains far from convergence.

Together, these observations suggest a different use of short task
adaptation.
Rather than treating an early LoRA checkpoint as an incomplete solution to be
continued or copied, we use it as an \emph{early task probe}.
Although its activation configuration is still incomplete, it already
contains sufficient task-conditioned structure to construct layer-wise
boundary targets for a fresh adapter.
For an input $x$, the probe provides a target sign for each monitored
pre-activation, thereby transforming global task supervision into direct
intermediate supervision.

This change in supervision is particularly relevant to low-rank adaptation.
For the LoRA parameters $\phi_l$ associated with an intermediate
pre-activation $z_l$, output-level task supervision and direct boundary
supervision reach the adapter through
$\nabla_{\phi_l}\mathcal{L}_{\mathrm{task}}
    =
    J_l^\top H_l^\top q,
    \nabla_{\phi_l}\mathcal{L}_{\mathrm{B}}
    =
    J_l^\top q_l^{\mathrm{B}}$,
where $J_l=\partial z_l/\partial\phi_l$,
$H_l=\partial o/\partial z_l$, and $q$ and $q_l^{\mathrm{B}}$ denote the
corresponding output- and boundary-level signals.
Thus, task supervision is filtered via the downstream Jacobian $H_l$
before reaching the adapter, whereas a boundary target directly supervises
the corresponding intermediate activation.
Our analysis in Section~\ref{sec:method} shows that this direct pathway can
expose boundary-relevant LoRA directions that are attenuated or locally
unobservable through the output mapping.
We refer to transformation of probe-acquired task information into direct
intermediate targets as \emph{boundary-supervision lifting}.

Based on this insight, we propose
\textbf{Activation Boundary Matching for LoRA (ABM-LoRA)}.
ABM first trains a standard LoRA adapter for a small number of steps and
freezes it as an early task probe.
During boundary matching, representative downstream inputs are forwarded
through the frozen probe and a freshly factorized adapter, and the probe's
layer-wise pre-activation signs directly supervise the corresponding
activations of the fresh adapter.
ABM transfers only these activation sides, rather than the probe weights or
exact activation values.
Its margin-based squared hinge loss supervises violated coordinates; once a
coordinate reaches its target margin, the corresponding loss term and
gradient vanish.

The probe is discarded after boundary matching, and the resulting adapter is
used as the initialization for standard downstream fine-tuning.
Because the boundary objective is removed after initialization, the original 
task objective remains free to preserve useful transferred
boundaries, discover additional boundaries not captured by the short probe,
or correct premature ones.
ABM therefore assumes neither that the probe has recovered the complete final
activation configuration nor that boundary formation and numerical refinement
are independent or strictly sequential.
Across tasks, probes of only 50--200 steps generally approach
full-reference performance, demonstrating that complete boundary recovery is
unnecessary for effective initialization.
Across language understanding, vision, and instruction-tuning tasks with
diverse backbones, ABM provides broad improvements over standard LoRA.
It requires neither pretrained-weight decomposition nor explicit full-gradient
estimation and can be integrated into standard LoRA training without
architectural modifications.
Despite its simple implementation and modest overhead, ABM remains competitive
with specialized initialization methods.
Together, these results show that ABM extracts a partial but reusable
functional trace from early adaptation and turns it into a broadly applicable
LoRA initialization.
Our contributions are summarized as follows:
\begin{itemize}
    \item
    We uncover a temporal asymmetry in standard LoRA adaptation:
    task-induced activation boundaries develop before activation values and
    effective updates numerically converge, while even partially recovered
    boundaries already provide an effective initialization signal.
    \vspace{-1mm}
    \item
    We propose ABM-LoRA, which extracts a partial boundary trace using a short
    task probe and transforms it into direct layer-wise supervision for a
    fresh LoRA initialization.
    Our analysis shows that this supervision can expose boundary-relevant
    directions attenuated through the downstream output mapping.
    \vspace{-1mm}
    \item
    Experiments across language understanding, vision, and instruction-tuning
    tasks with diverse backbones show improvements over standard LoRA with
    limited overhead and distinguish ABM from longer training, weight reuse,
    output distillation, and activation-value matching.
\end{itemize}

\section{Related Work}
\label{sec:related}
 
\textbf{Parameter-Efficient Fine-Tuning.}
Parameter-efficient fine-tuning (PEFT) has emerged as an effective strategy
for adapting large pretrained models across both language and vision domains.
Following the success of large transformer models such as BERT, GPT-style models, and Vision Transformers (ViTs)~\cite{dosovitskiy2020image}, numerous PEFT methods
have been proposed to reduce the cost of downstream adaptation.
Representative approaches include inserting lightweight adapter modules (AdaptFormer~\cite{chen2022adaptformer}), introducing trainable prompts at the input level (Visual Prompt Tuning, VPT~\cite{jia2022visual}), and applying low-rank adaptation mechanisms such as LoRA~\cite{hu2022lora}. These methods primarily focus on architectural modifications, prompt design, or rank allocation to improve adaptation efficiency. In contrast, our work addresses an orthogonal yet underexplored aspect: initialization. We demonstrate that carefully aligning activation boundaries at initialization can further enhance both convergence behavior and final performance of LoRA-based adaptation.
 
\noindent\textbf{Initialization in Low-Rank Adaptation.}
Initialization strongly influences the convergence and final quality of LoRA-based adaptation, and a growing body of work replaces the default random scheme with more structured alternatives. Weight-based methods such as PiSSA~\cite{meng2024pissa} and OLoRA~\cite{buyukakyuz2024olora} derive the low-rank subspace from the singular or orthonormal structure of the pretrained weights. Data-driven methods instead draw on the downstream task itself: EVA~\cite{paischer2026parameter} initializes from the singular directions of activation statistics, and LoRA-One~\cite{zhang2025lora} from those of the one-step fine-tuning gradient.
 
Despite their differences, these methods share a common shape: the initialization is a single precomputed object, evaluated at the pretrained point, that captures the geometry \emph{before} adaptation but not how the adapter must move as the task is learned. ABM-LoRA departs from this pattern. Rather than reading off a static statistic, it obtains its target from a short task-adapted probe and matches the \emph{activation boundaries} the probe induces on the data, transferring the sign structure of the pretrained Jacobian without any singular value decomposition or gradient estimate.
 
\noindent\textbf{Knowledge Distillation and Activation Alignment.}
Knowledge distillation is a widely used transfer-learning paradigm that transfers information from teacher to student models. Activation Boundary Distillation (ABD)~\cite{heo2019knowledge} aligns binary activation masks to transfer structural knowledge across networks. Related approaches include hidden response transfer~\cite{romero2015fitnets}, relational feature distillation~\cite{yim2017gift}, and attention-based distillation~\cite{zagoruyko2017paying}. These methods emphasize the importance of piecewise linear regions induced by ReLU-like activations~\cite{montufar2014number,pascanu2013number}.
Unlike conventional distillation, ABM uses a briefly trained LoRA probe rather
than a converged teacher and applies boundary alignment only to initialize a
fresh adapter.
It transfers layer-wise activation sides without copying probe weights or
exact activation values.
After initialization, both the probe and boundary objective are removed, and
the adapter is trained solely with the downstream loss.
ABM repurposes activation-boundary alignment as a task-informed LoRA
initialization mechanism rather than a persistent training-time supervision
objective.

\begin{figure*}[t]
\centering
\includegraphics[width=\textwidth, height=0.35\textwidth]{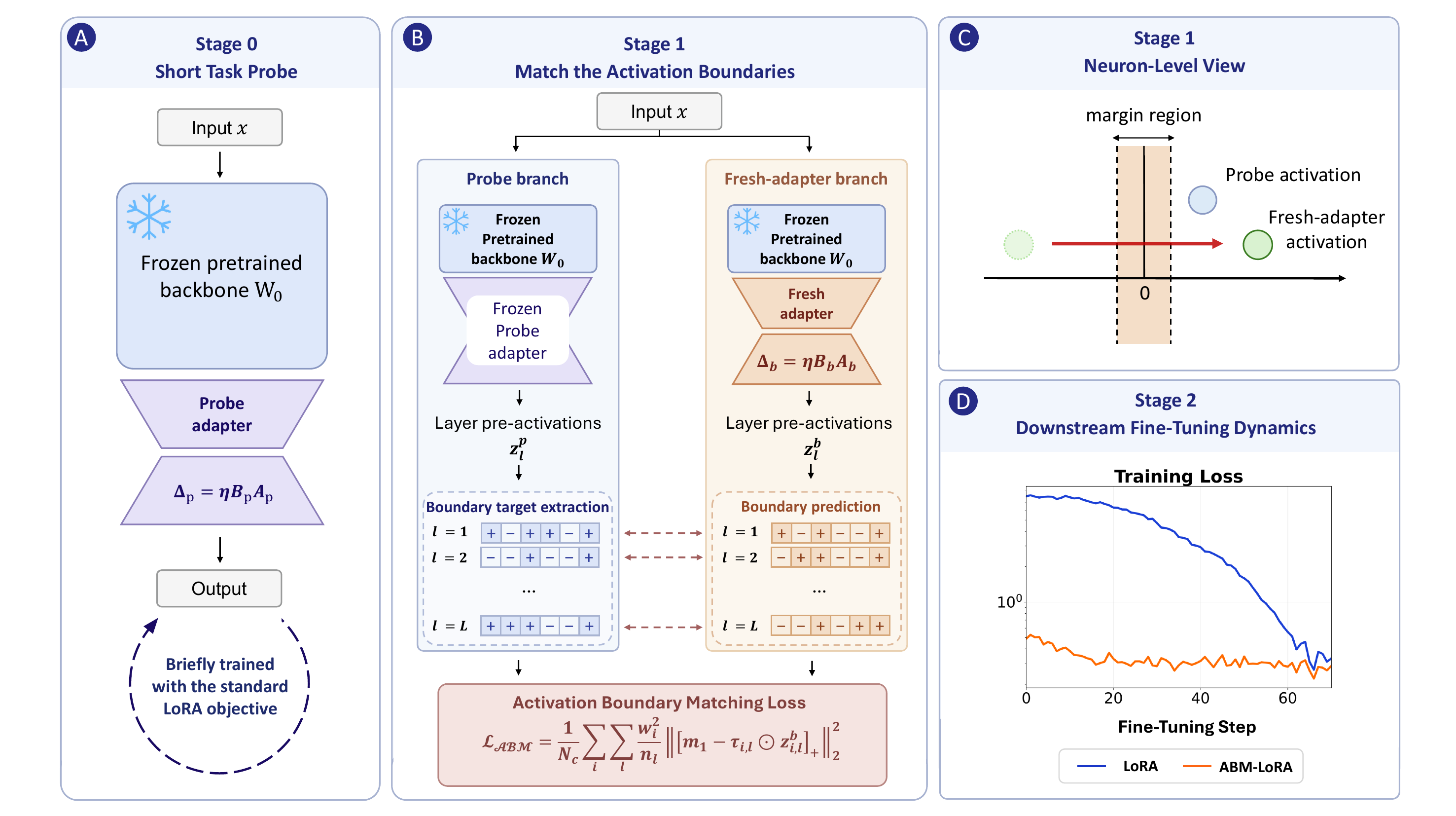}
\vspace{-6mm}
\caption{
    \textbf{Overview of ABM-LoRA.}
    \textbf{(A) Stage 0: Short task probe.}
    A standard LoRA adapter is briefly trained on the downstream objective to
    expose early task-induced activation boundaries.
    \textbf{(B) Stage 1: Boundary matching.}
    The frozen probe provides layer-wise activation signs as direct targets
    for a freshly initialized LoRA adapter.
    Only the fresh adapter is optimized using the weighted squared hinge
    objective.
    \textbf{(C) Neuron-level view.}
    Each probe sign defines a target half-space, and the fresh pre-activation
    is moved beyond margin $m$ without matching the probe's exact value.
    \textbf{(D) Stage 2: Downstream fine-tuning.}
    The probe and boundary objective are removed, and the boundary-matched
    adapter is optimized solely with the original task objective to produce
    the final task adapter. The ABM-initialized adapter begins Stage~2 with substantially lower task loss
than standard LoRA and converges from this task-informed starting point.
}
\label{fig:method_overview}
\vspace{-5mm}
\end{figure*}

\section{Method}
\label{sec:method}

\subsection{Problem Setup and Output-Level Supervision}
\label{sec:problem_setup}

Let $f(x;W_0,\phi)$ denote a pretrained model with frozen parameters $W_0$
and trainable LoRA parameters $\phi$.
For each adapted linear layer $l$, LoRA parameterizes its weight as
\begin{equation}
    W_l
    =
    W_{0,l}
    +
    s_l B_lA_l,
    \quad
    A_l\in\mathbb{R}^{r\times d_l^{\mathrm{in}}},
    \,
    B_l\in\mathbb{R}^{d_l^{\mathrm{out}}\times r},
    \label{eq:lora}
\end{equation}
where $s_l$ is the LoRA scaling factor and
$r\ll\min(d_l^{\mathrm{out}},d_l^{\mathrm{in}})$.
We collect all trainable factors as
$\phi=\{A_l,B_l\}_{l\in\mathcal S}$, where $\mathcal S$ is the set of
LoRA-adapted layers.
Given a downstream dataset
$\mathcal D=\{(x_i,y_i)\}_{i=1}^{N}$, standard LoRA minimizes
\begin{equation}
    \mathcal L_{\mathrm{task}}(\phi)
    =
    \frac{1}{N}
    \sum_{i=1}^{N}
    \ell\!\left(f(x_i;W_0,\phi),y_i\right).
    \label{eq:task_objective}
\end{equation}

Standard LoRA receives supervision exclusively from the final task output.
To characterize how this signal reaches an intermediate adapter, let
$\phi_l=\{A_l,B_l\}$ denote the LoRA parameters associated with
pre-activation $z_l$, and let $o$ denote the model output.
We define
\begin{equation}
    J_l
    =
    \frac{\partial z_l}{\partial\phi_l},
    \qquad
    H_l
    =
    \frac{\partial o}{\partial z_l}.
    \label{eq:local_jacobians}
\end{equation}
Because the effect of $\phi_l$ on the output is transmitted through $z_l$,
the chain rule gives
\begin{equation}
    \nabla_{\phi_l}\mathcal L_{\mathrm{task}}
    =
    J_l^\top H_l^\top q,
    \qquad
    q=\nabla_o\mathcal L_{\mathrm{task}}.
    \label{eq:task_gradient}
\end{equation}
Thus, task supervision is transformed by the downstream Jacobian $H_l$
before reaching the intermediate LoRA module.
Activation directions attenuated or suppressed by this mapping may
consequently provide weak or no local supervision to $\phi_l$.
While this downstream transformation is common to neural-network training,
it is particularly relevant to LoRA because adaptation is additionally
restricted by a low-rank parameterization.

\subsection{ABM-LoRA Overview}
\label{sec:abm_overview}

Fig.\ref{fig:method_overview} presents the overall ABM-LoRA procedure.
Motivated by the early emergence of activation boundaries observed in
Fig.\ref{fig:motivation}, ABM constructs a task-informed LoRA initialization
through three stages.

\noindent\textbf{Stage 0: Short task probing.}
As shown in Fig.\ref{fig:method_overview}A, we briefly train a standard LoRA
adapter on the downstream objective.
This adapter serves as an \emph{early task probe} rather than a fully
converged reference model.
After $T_p$ steps, the probe is frozen and used only to expose the
task-induced layer-wise activation signs that have emerged during early
adaptation.

\noindent\textbf{Stage 1: Activation boundary matching.}
As shown in Figure~\ref{fig:method_overview}B, the same downstream input is
forwarded through the frozen probe branch and a fresh-adapter branch.
The signs of the probe pre-activations are used as direct targets for the
corresponding pre-activations of the fresh adapter.
Only the fresh LoRA factors are optimized; the pretrained backbone and probe
adapter remain frozen.
At the neuron level
(Figure~\ref{fig:method_overview}C), each probe sign defines a target
half-space.
The fresh activation is encouraged to enter this half-space with margin $m$,
without reproducing the probe's exact activation value.

\noindent\textbf{Stage 2: Downstream fine-tuning.}
After boundary matching, we discard the probe and use the boundary-matched
adapter as the initialization for standard downstream fine-tuning.
The boundary objective is removed, and only the original task objective is
optimized.
As shown in Figure~\ref{fig:method_overview}(D), the ABM-initialized adapter
begins this stage with substantially lower task loss than standard LoRA,
indicating that the partial boundary trace already provides a more
task-informed starting point.
Because no boundary constraint is retained, downstream optimization remains
free to preserve, complete, or correct the transferred activation
configuration, producing the final task adapter.

\subsection{Stage 0: Short Task Probe}
\label{sec:probe}

We initialize a standard LoRA adapter $\phi_0^p$ using the conventional
scheme and optimize it using Eq.\eqref{eq:task_objective} for $T_p$
steps:
\begin{equation}
    \phi^p
    =
    \operatorname{Probe}
    \left(
        \phi_0^p,\mathcal D,T_p
    \right).
    \label{eq:probe}
\end{equation}
The probe is not expected to converge to a final task solution.
Its purpose is to expose the task-conditioned activation structure formed
during early adaptation.

Let
$\mathcal M=\{0,\ldots,L-1\}$ denote the ordered set of $L$ monitored layers.
For each $l\in\mathcal M$, let
$z_l(x;\phi^p)\in\mathbb R^{n_l}$ denote the corresponding probe
pre-activation.
For notational simplicity, all monitored token and hidden coordinates at
layer $l$ are flattened into a vector of size $n_l$.
We define the probe-induced activation signature as
\begin{equation}
    \tau_l(x)
    =
    \operatorname{sign}
    \left(
        z_l(x;\phi^p)
    \right)
    \in\{-1,+1\}^{n_l},
    \label{eq:boundary_target}
\end{equation}
where $\operatorname{sign}(u)=+1$ for $u\geq0$ and $-1$ otherwise.
The collection
\begin{equation}
    \mathcal T^p
    =
    \left\{
        \tau_l(x)
        \,\middle|\,
        x\in\mathcal D_c,\;
        l\in\mathcal M
    \right\}
    \label{eq:boundary_collection}
\end{equation}
forms the layer-wise boundary targets, where
$\mathcal D_c\subseteq\mathcal D$ is the calibration data used for boundary
matching.

After $T_p$ steps, we freeze the probe and use it only to produce boundary
targets during Stage~1.
No probe weight, exact activation value, or optimizer state is transferred to
the fresh adapter.

\subsection{Stage 1: Activation Boundary Matching}
\label{sec:boundary_matching}

We instantiate a fresh LoRA adapter $\phi_0^b$ using the standard
initialization: $A_{l,0}^b$ is Kaiming-initialized and $B_{l,0}^b$ is set to
zero.
Thus, its initial effective update is
\begin{equation}
    \Delta_{l,0}^b
    =
    s_lB_{l,0}^bA_{l,0}^b
    =
    0,
    \label{eq:fresh_zero_update}
\end{equation}
and the fresh branch initially recovers the pretrained function.

For the same calibration input $x_i$, let
\begin{equation}
    z_{i,l}^b
    =
    z_l(x_i;\phi^b)
\end{equation}
denote the pre-activation of the fresh branch at monitored layer $l$.
The frozen probe provides the corresponding target
$\tau_{i,l}=\tau_l(x_i)$ from Eq.\eqref{eq:boundary_target}.
Given a margin $m>0$, we define the coordinate-wise boundary residual as
\begin{equation}
    r_{i,l}(\phi^b)
    =
    \left[
        m\mathbf 1
        -
        \tau_{i,l}\odot z_{i,l}^b
    \right]_+,
    \label{eq:boundary_residual}
\end{equation}
where $[\cdot]_+$ is applied element-wise.

The fresh adapter is optimized using the weighted squared hinge objective
\begin{equation}
    \mathcal L_{\mathrm{ABM}}(\phi^b)
    =
    \frac{1}{N_c}
    \sum_{i=1}^{N_c}
    \sum_{l\in\mathcal M}
    \frac{w_l^2}{n_l}
    \left\|
        r_{i,l}(\phi^b)
    \right\|_2^2,
    \label{eq:loss_abm}
\end{equation}
where $N_c=|\mathcal D_c|$, and division by $n_l$ averages the squared
boundary residual over the monitored activation coordinates at layer $l$.
The depth-dependent weight is defined as
\begin{equation}
    w_l
    =
    \frac{l+1}{L},
    \qquad
    l\in\mathcal M=\{0,\ldots,L-1\}.
    \label{eq:layer_weight}
\end{equation}
The increasing weights place greater emphasis on boundary targets from deeper
representations while retaining supervision across all monitored layers.

Starting from $\phi_0^b$, we optimize only the fresh LoRA factors for $T_b$
steps:
\begin{equation}
    \phi^{\mathrm{ABM}}
    =
    \operatorname{BoundaryMatch}
    \left(
        \phi_0^b,
        \phi^p,
        \mathcal D_c,
        T_b
    \right).
    \label{eq:abm_initialization}
\end{equation}
At each matching step, the frozen probe generates the target signs online,
while gradients are computed only for the fresh LoRA factors.
The pretrained backbone and probe adapter remain frozen throughout this
stage.
After boundary matching, the probe is discarded and
$\phi^{\mathrm{ABM}}$ is retained as the initialization for downstream
fine-tuning.

For coordinate $j$, the corresponding boundary term is active only when
\begin{equation}
    \tau_{i,l,j}z_{i,l,j}^b<m.
\end{equation}
Once
$\tau_{i,l,j}z_{i,l,j}^b\geq m$, its residual and gradient vanish.
ABM therefore transfers the probe-induced activation side without requiring
the fresh adapter to reproduce the probe's exact activation value or LoRA
parameters.
Although Stage~1 does not directly use ground-truth labels, it remains
task-conditioned through the boundary targets produced by the short probe.

\subsection{Why Does Boundary Supervision Help?}
\label{sec:boundary_analysis}

ABM does not introduce additional task information: its boundary targets are
extracted from a short probe trained with the same downstream objective.
Its benefit instead comes from changing how the probe-acquired information is
delivered to a fresh LoRA adapter.
While standard task supervision reaches an intermediate adapter only through
the remaining downstream computation, ABM attaches the transferred targets
directly to the corresponding pre-activations.

Using the local Jacobians $J_l$ and $H_l$ in
Eq.\eqref{eq:local_jacobians}, consider a small perturbation
$\phi_l\mapsto\phi_l+\epsilon v$ of the LoRA parameters at layer $l$, with all
other parameters held fixed.
Up to higher-order terms, this perturbation changes the intermediate
pre-activation by $\epsilon J_lv$ and the final output by
$\epsilon H_lJ_lv$.
Thus, output-level supervision observes the parameter direction $v$
only through the downstream-transformed quantity $H_lJ_lv$.

For clarity, we suppress the sample index and flatten the monitored
activation coordinates into a single vector.
For boundary matching, let $D_l$ denote the diagonal mask selecting
pre-activation coordinates that have not yet reached their target-side
margin:
\begin{equation}
    D_l
    =
    \operatorname{Diag}
    \left(
        \indicator
        \left[
            m\mathbf{1}
            -
            \tau_l\odot z_l
            >
            0
        \right]
    \right).
    \label{eq:active_boundary_mask_main}
\end{equation}
The first-order change in the active boundary residual is, up to the
coordinate-wise target signs, $D_lJ_lv$.
This leads to the following local distinction.

\begin{proposition}[Direct boundary observability]
\label{prop:boundary_observability_main}
If a LoRA parameter direction $v$ satisfies
\begin{equation}
    H_lJ_lv=0
    \qquad\text{and}\qquad
    D_lJ_lv\neq0,
    \label{eq:direct_boundary_observability}
\end{equation}
then $v$ is locally unobservable through the model output but observable
through the active boundary constraints.
\end{proposition}

The proof of
Proposition~\ref{prop:boundary_observability_main} is provided in
Appendix~\ref{app:theoretical_proofs}.
Corollary~\ref{cor:downstream_attenuation} extends this result to attenuated,
rather than exactly null, output directions, while
Proposition~\ref{prop:multilayer_observability} shows that adding supervision
at further layers cannot reduce local boundary observability.
These results do not imply that every boundary-relevant direction is
task-optimal or that ABM necessarily produces a larger task gradient.
Instead, they establish the structural role of
\emph{boundary-supervision lifting}: task information discovered by the probe
can directly supervise intermediate LoRA modules without first passing
through the current downstream Jacobian.
ABM thereby converts an early functional trace into layer-wise initialization
signals while leaving subsequent task fine-tuning unconstrained.


\noindent\textbf{Empirical confirmation.}
Proposition~\ref{prop:boundary_observability_main} predicts that
\emph{where} and \emph{how} probe information is delivered should matter
independently of the information itself.
Fixing all other stages and varying only the Stage~1 supervision,
Table~\ref{tab:supervision} isolates this effect.
Output-level distillation, which supervises the probe logits rather than the
intermediate pre-activations, is the weakest of the four targets, consistent
with the attenuation of boundary-relevant directions through the downstream
mapping formalized in Proposition~\ref{prop:boundary_observability_main}.
Moving supervision to the layer-wise pre-activations helps, and among
layer-wise targets, reproducing the probe's exact activation values is not
necessary: transferring only the activation side (sign, $m{=}0$) already
matches or exceeds exact value matching, and adding the margin ($m{>}0$)
gives a further gain.
Finally, the shuffled control is the weakest overall, confirming that the
gain stems from \emph{task-relevant} boundary structure rather than from
merely perturbing the fresh adapter for additional steps.
These observations support the central design choice of ABM: it is the
layer-wise \emph{boundary side}, not the exact activation value or the
output-level signal, that constitutes the transferable initialization
information.

\begin{table}[t]
\centering
\small
\caption{\textbf{Effect of the Stage~1 supervision target on downstream accuracy (MNLI, \%), with all other stages fixed.} \emph{Output} supervises the probe logits; the middle rows supervise layer-wise pre-activations by matching exact values, activation signs, or signs with a margin (ABM). \emph{Shuffled} permutes the boundary targets across coordinates as a control. \textbf{Bold} marks the proposed method.}
\label{tab:supervision}
\vspace{-3mm}
\begin{tabular*}{\columnwidth}{@{\extracolsep{\fill}} lc}
\toprule
Stage~1 supervision target & Acc. \\
\midrule
Output distillation (logits) & 84.19 \\
Value matching & 84.23 \\
Sign matching ($m{=}0$) & 84.32 \\
\textbf{Boundary matching ($m{>}0$, ABM)} & \textbf{84.65} \\
\midrule
Shuffled boundary (control) & 83.91 \\
\bottomrule
\end{tabular*}
\vspace{-5mm}
\end{table}

\subsection{Stage 2: Downstream Fine-Tuning}
\label{sec:downstream_finetuning}

Finally, we initialize the task adapter with the boundary-matched parameters,
\begin{equation}
    \phi_0^{\mathrm{task}}
    =
    \phi^{\mathrm{ABM}},
\end{equation}
and optimize only the original downstream objective:
\begin{equation}
    \phi^\star
    =
    \operatorname{FineTune}
    \left(
        \phi^{\mathrm{ABM}},
        \mathcal D,
        T_f
    \right).
    \label{eq:abm_finetuning}
\end{equation}
The probe and boundary-matching objective are not used in this stage.
Thus, ABM changes only the initialization of standard LoRA and introduces no
additional constraint during downstream adaptation. 

\section{Experiments}
\label{sec:exp}
 
We evaluate \method{} across both vision and language, spanning encoder--decoder, decoder-only, and hierarchical vision backbones, so as to test whether activation-boundary alignment transfers across architectures. Throughout, the comparison is controlled: within each benchmark all initialization methods share the same backbone, adapter placement, rank, optimizer, schedule, and step budget, and differ \emph{only} in how the adapter is initialized. For \method{}, the Stage~0 reference is a standard LoRA run under exactly this shared configuration, so that any difference relative to standard LoRA is attributable to the initialization alone.

\begin{table*}[t]
\centering
\caption{\textbf{Vision results on ConvNeXt-Tiny and Swin-Tiny across four fine-grained classification benchmarks (Flowers102, Oxford-IIIT Pets, Food-101, FGVC-Aircraft).} All adapters use rank $r{=}8$ and share the same placement, optimizer, and step budget, differing only in initialization. \textbf{Bold} and \underline{underline} mark the best and second-best result in each column among LoRA-based methods; Full FT is shown for reference and excluded from ranking. Cell shading is proportional to per-column accuracy and is provided only as a visual aid; the table remains fully legible in grayscale. Single-seed top-1 accuracy (\%).}
\label{tab:vision}
\setlength{\tabcolsep}{7pt}
\renewcommand{\arraystretch}{0.7}
\vspace{-3mm}
\scriptsize
\begin{tabular*}{\textwidth}{@{\extracolsep{\fill}} lcccccccccc}
\toprule
& \multicolumn{5}{c}{\textbf{ConvNeXt-T}} & \multicolumn{5}{c}{\textbf{Swin-T}} \\
\cmidrule(lr){2-6} \cmidrule(lr){7-11}
Method & Flwr & Pets & Food & Air & Avg & Flwr & Pets & Food & Air & Avg \\
\midrule
Full FT & 95.98 & 93.62 & 86.76 & 77.41 & 88.44 & 95.67 & 94.22 & 89.03 & 73.30 & 88.06 \\
\midrule
Normal & \cellcolor{evagreen!9}94.62 & \cellcolor{evagreen!34}\underline{93.98} & \cellcolor{evagreen!28}86.15 & \cellcolor{evagreen!8}74.17 & \cellcolor{evagreen!15}87.23 & \cellcolor{evagreen!15}93.14 & \cellcolor{evagreen!23}94.09 & \cellcolor{evagreen!34}\underline{85.99} & \cellcolor{evagreen!14}66.76 & \cellcolor{evagreen!8}85.00 \\
PiSSA & \cellcolor{evagreen!23}95.07 & \cellcolor{evagreen!31}93.89 & \cellcolor{evagreen!31}86.33 & \cellcolor{evagreen!22}75.43 & \cellcolor{evagreen!26}87.68 & \cellcolor{evagreen!8}92.94 & \cellcolor{evagreen!8}93.62 & \cellcolor{evagreen!32}85.93 & \cellcolor{evagreen!35}\underline{67.90} & \cellcolor{evagreen!15}85.10 \\
OLoRA & \cellcolor{evagreen!20}94.97 & \cellcolor{evagreen!8}93.02 & \cellcolor{evagreen!8}85.20 & \cellcolor{evagreen!13}74.65 & \cellcolor{evagreen!8}86.96 & \cellcolor{evagreen!38}\textbf{93.77} & \cellcolor{evagreen!10}93.68 & \cellcolor{evagreen!8}85.18 & \cellcolor{evagreen!27}67.48 & \cellcolor{evagreen!10}85.03 \\
EVA & \cellcolor{evagreen!8}94.60 & \cellcolor{evagreen!38}\textbf{94.14} & \cellcolor{evagreen!31}86.30 & \cellcolor{evagreen!9}74.26 & \cellcolor{evagreen!17}87.33 & \cellcolor{evagreen!33}\underline{93.64} & \cellcolor{evagreen!38}\textbf{94.58} & \cellcolor{evagreen!29}85.84 & \cellcolor{evagreen!8}66.40 & \cellcolor{evagreen!16}85.12 \\
LoRA-One & \cellcolor{evagreen!30}\underline{95.32} & \cellcolor{evagreen!31}93.89 & \cellcolor{evagreen!36}\underline{86.57} & \cellcolor{evagreen!37}\underline{76.75} & \cellcolor{evagreen!38}\textbf{88.13} & \cellcolor{evagreen!26}93.43 & \cellcolor{evagreen!29}\underline{94.28} & \cellcolor{evagreen!30}85.87 & \cellcolor{evagreen!38}\textbf{68.08} & \cellcolor{evagreen!38}\textbf{85.42} \\
\textbf{ABM} & \cellcolor{evagreen!38}\textbf{95.56} & \cellcolor{evagreen!13}93.21 & \cellcolor{evagreen!38}\textbf{86.66} & \cellcolor{evagreen!38}\textbf{76.84} & \cellcolor{evagreen!36}\underline{88.07} & \cellcolor{evagreen!31}93.59 & \cellcolor{evagreen!28}94.25 & \cellcolor{evagreen!38}\textbf{86.13} & \cellcolor{evagreen!23}67.24 & \cellcolor{evagreen!29}\underline{85.30} \\
\bottomrule
\end{tabular*}
\vspace{-5mm}
\end{table*}

\subsection{Settings}
\noindent\textbf{Benchmarks and Baselines.}
We study four settings: fine-grained visual classification with
ConvNeXt-T~\cite{liu2022convnet} and Swin-T~\cite{liu2021swin} on
Flowers102~\cite{nilsback2008automated},
Oxford-IIIT Pets~\cite{parkhi2012cats},
Food-101~\cite{bossard2014food}, and
FGVC-Aircraft~\cite{maji2013fine};
natural language understanding with T5-base~\cite{raffel2020exploring} on
GLUE~\cite{wang2019glue}; and instruction tuning with
Qwen2.5-1.5B~\cite{yang2024qwen25} on Alpaca~\cite{taori2023alpaca} and
LLaMA2-7B~\cite{touvron2023llama2} on WizardLM~\cite{xu2024wizardlm}.
Depending on the setting we compare against standard LoRA~\cite{hu2022lora}
and a set of initialization baselines drawn from Normal and Gaussian random
initialization, PiSSA~\cite{meng2024pissa}, OLoRA~\cite{buyukakyuz2024olora},
EVA~\cite{paischer2026parameter}, rsLoRA~\cite{kalajdzievski2023rank},
Orthogonal initialization, and LoRA-One~\cite{zhang2025lora}.
Full fine-tuning is reported for reference where available.
 
\noindent\textbf{Implementation.}
Unless noted, LoRA adapters use rank $r{=}8$ and $\alpha{=}16$. Vision (ConvNeXt-T, Swin-T): AdamW~\cite{loshchilov2019decoupled} ($1\times10^{-3}$, weight decay $1\times10^{-4}$, batch size $64$, one-epoch warm-up); adapters are inserted into attention projections. Training epochs are 10 (Flowers102, Pets), 5 (Food-101), and 20/25 (FGVC-Aircraft for ConvNeXt-T/Swin-T). All methods share identical settings except initialization. Stage~1 matches the last two pre-activation stages for 500 steps ($3\times10^{-4}$, batch size $32$, $m{=}1.0$). GLUE (T5-base): AdamW with cosine decay~\cite{loshchilov2017sgdr}, learning rate $1\times10^{-4}$, warm-up ratio $0.03$, sequence length $128$; Stage~1 matches six encoder/decoder layers for 100 steps ($3\times10^{-4}$, $m{=}0.5$). Instruction tuning: one epoch with cosine decay, sequence length $1024$, effective batch size $32$, and learning rates $2\times10^{-5}$ (LLaMA2-7B) or $2\times10^{-4}$ (Qwen2.5-1.5B). Stage~1 matches the pre-activation of \texttt{gate\_proj} (100 steps for LLaMA2, 500 for Qwen; $m{=}0.5$). Full hyperparameters are provided in the appendix.

For \method{}, the Stage~0 probe is a standard LoRA run in all settings
except where noted: on LLaMA2-7B we additionally report a probe trained with
orthogonal LoRA initialization (\emph{ABM (Ortho.)}) to test the sensitivity
of ABM to the reference it is distilled from.
Both LLaMA2-7B variants match the last $16$ decoder layers (layers
$16$--$31$) in Stage~1. All experiments were conducted on NVIDIA RTX 3090 GPUs.

 \subsection{Results}
\noindent\textbf{Vision Results.}
The takeaway is that \method{} is consistently competitive across backbones while relying only on forward passes, matching gradient-based initializers without their computational cost.

On both ConvNeXt-T and Swin-T in Table~\ref{tab:vision}, \method{} ranks second in average accuracy, behind LoRA-One and ahead of every other initializer. At the per-task level it is more often the best: \method{} attains the top accuracy in four of the eight backbone--dataset settings (Flowers102, Food-101, and Aircraft on ConvNeXt-T, and Food-101 on Swin-T). Its gains are largest on the harder, lower-accuracy tasks: on ConvNeXt-T Aircraft it reaches $76.84$ against $74.17$ for standard LoRA, a margin of $2.67$ points, and it improves over standard LoRA on Food-101 for both backbones.
 
It is worth reading the table by which methods stay near the top across \emph{both} backbones. Only \method{} and LoRA-One occupy the top two average positions on each, whereas the static initializers PiSSA, OLoRA, and EVA move between third and sixth depending on the backbone: OLoRA is strongest on Swin-T Flowers102 yet lowest on ConvNeXt-T, and EVA leads on Pets but trails on Aircraft. \method{} thus matches the cross-backbone stability of LoRA-One, which relies on an explicit gradient computation, using only forward passes over a reference adapter. This agrees with our analysis: a subspace aligned with the task-adapted activation geometry retains more of the useful gradient structure than one derived from pretrained weight statistics alone.
 
\noindent\textbf{Language: GLUE.}
We next fine-tune T5-base on the GLUE benchmark, comparing \method{} against Normal and Gaussian random initialization, PiSSA, EVA, and LoRA-One, all under an identical adapter and training configuration. We report four tasks (CoLA, SST-2, MNLI, QNLI), averaged over three seeds.
 
\begin{table}[t]
\centering
\small
\caption{\textbf{Language fine-tuning results on GLUE with T5-base, averaged over three seeds.} All adapters use rank $r{=}8$ and share the same training configuration, differing only in initialization. \textbf{Bold} and \underline{underline} mark the best and second-best per column. Metrics are Matthews correlation for CoLA and accuracy (\%) for the other tasks; per-seed standard deviations are reported in the appendix.}
\label{tab:glue}
\setlength{\tabcolsep}{7pt}
\renewcommand{\arraystretch}{0.7}
\vspace{-3mm}
\scriptsize
\begin{tabular*}{\columnwidth}{@{\extracolsep{\fill}} lccccc}
\toprule
Method & CoLA & SST-2 & MNLI & QNLI & Avg \\
\midrule
Normal & \cellcolor{evablue!8}69.86 & \cellcolor{evablue!33}93.85 & \cellcolor{evablue!11}84.31 & \cellcolor{evablue!8}92.83 & \cellcolor{evablue!8}85.21 \\
Gaussian & \cellcolor{evablue!22}73.60 & \cellcolor{evablue!8}93.35 & \cellcolor{evablue!8}84.23 & \cellcolor{evablue!12}92.89 & \cellcolor{evablue!18}86.02 \\
PiSSA & \cellcolor{evablue!36}\underline{77.53} & \cellcolor{evablue!29}93.77 & \cellcolor{evablue!38}\textbf{84.99} & \cellcolor{evablue!22}93.03 & \cellcolor{evablue!35}\underline{87.33} \\
EVA & \cellcolor{evablue!38}\textbf{78.11} & \cellcolor{evablue!34}\underline{93.88} & \cellcolor{evablue!37}84.96 & \cellcolor{evablue!38}\textbf{93.25} & \cellcolor{evablue!38}\textbf{87.55} \\
LoRA-One & \cellcolor{evablue!28}75.36 & \cellcolor{evablue!33}93.85 & \cellcolor{evablue!38}\textbf{84.99} & \cellcolor{evablue!18}92.97 & \cellcolor{evablue!28}86.79 \\
\textbf{ABM} & \cellcolor{evablue!35}77.15 & \cellcolor{evablue!38}\textbf{93.96} & \cellcolor{evablue!30}84.80 & \cellcolor{evablue!24}\underline{93.06} & \cellcolor{evablue!34}87.24 \\
\bottomrule
\end{tabular*}
\vspace{-5mm}
\end{table}
 
Table~\ref{tab:glue} shows that the initialization methods cluster tightly on GLUE: the top three initializers lie within about three tenths of a point in average score. Within this band \method{} attains the best result on SST-2 and the second-best on QNLI, and lands third on average, just behind EVA and PiSSA and ahead of LoRA-One. What distinguishes it is how it gets there: EVA and PiSSA build their subspace from a singular value decomposition and LoRA-One from a gradient estimate, whereas \method{} reaches the same band without either, from activation signs alone. The spread across methods is by far the largest on CoLA, the smallest-resource task ($8$ points between best and worst), while on the higher-resource tasks the methods are nearly indistinguishable; this is consistent with the view that initialization matters most when the downstream signal is limited.
 
\noindent\textbf{Language: Instruction Tuning.}
We further evaluate on generative instruction tuning, where LoRA is applied to two decoder-only backbones: Qwen2.5-1.5B fine-tuned on Alpaca and LLaMA2-7B fine-tuned on WizardLM. We report MT-Bench~\cite{zheng2023judging} scores and AlpacaEval~\cite{li2023alpacaeval} length-controlled (LC)~\cite{dubois2024length}
and raw win rate (WR). On LLaMA2 we additionally compare two choices of Stage~0 reference for \method{}, trained with standard or with orthogonal LoRA initialization, both matching the last 16 decoder layers.
 
\begin{table}[t]
\centering
\small
\caption{\textbf{Instruction-tuning results on Qwen2.5-1.5B with the Alpaca dataset, evaluated on MT-Bench (score) and AlpacaEval (length-controlled and raw win rate).} \textbf{Bold} and \underline{underline} mark the best and second-best per column. For \method{}, the Stage~0 reference is a standard LoRA run.}
\label{tab:qwen}
\setlength{\tabcolsep}{7pt}
\renewcommand{\arraystretch}{0.7}
\vspace{-3mm}
\scriptsize
\begin{tabular*}{\columnwidth}{@{\extracolsep{\fill}}lccc}
\toprule
Method & MT-Bench & LC(\%) & WR(\%) \\
\midrule
Normal & \cellcolor{evagold!10}3.58 & \cellcolor{evagold!23}43.49 & \cellcolor{evagold!14}39.01 \\
\midrule
rsLoRA & \cellcolor{evagold!30}\underline{4.26} & \cellcolor{evagold!37}\underline{44.66} & \cellcolor{evagold!38}\textbf{40.75} \\
Orthogonal & \cellcolor{evagold!16}3.80 & \cellcolor{evagold!8}42.27 & \cellcolor{evagold!16}39.18 \\
Gaussian & \cellcolor{evagold!8}3.52 & \cellcolor{evagold!20}43.24 & \cellcolor{evagold!12}38.87 \\
EVA & \cellcolor{evagold!13}3.70 & \cellcolor{evagold!30}44.04 & \cellcolor{evagold!22}39.57 \\
LoRA-One & \cellcolor{evagold!38}\textbf{4.51} & \cellcolor{evagold!21}43.36 & \cellcolor{evagold!8}38.57 \\
\midrule
\textbf{ABM} & \cellcolor{evagold!27}4.14 & \cellcolor{evagold!38}\textbf{44.73} & \cellcolor{evagold!33}\underline{40.37} \\
\bottomrule
\end{tabular*}
\vspace{-5mm}
\end{table}
 
\begin{table}[t]
\centering
\small
\caption{\textbf{Instruction-tuning results on LLaMA2-7B with the WizardLM dataset, evaluated on MT-Bench (score) and AlpacaEval (length-controlled and raw win rate).} \textbf{Bold} and \underline{underline} mark the best and second-best per column. \emph{Standard} and \emph{Orthogonal} denote whether the Stage~0 reference adapter is trained with standard or orthogonal LoRA initialization; both match the last $16$ decoder layers in Stage~1.}
\label{tab:llama}
\setlength{\tabcolsep}{7pt}
\renewcommand{\arraystretch}{0.7}
\vspace{-3mm}
\begin{tabular*}{\columnwidth}{@{\extracolsep{\fill}} lccc}
\toprule
Method & MT-Bench & LC(\%) & WR(\%) \\
\midrule
Normal & \cellcolor{evagold!24}5.89 & \cellcolor{evagold!8}42.16 & \cellcolor{evagold!8}46.27 \\
\midrule
rsLoRA & \cellcolor{evagold!33}\underline{5.98} & \cellcolor{evagold!25}44.02 & \cellcolor{evagold!21}48.51 \\
Orthogonal & \cellcolor{evagold!30}5.95 & \cellcolor{evagold!36}45.25 & \cellcolor{evagold!29}49.88 \\
Gaussian & \cellcolor{evagold!33}\underline{5.98} & \cellcolor{evagold!14}42.78 & \cellcolor{evagold!22}48.63 \\
EVA & \cellcolor{evagold!33}\underline{5.98} & \cellcolor{evagold!32}44.86 & \cellcolor{evagold!32}\underline{50.37} \\
LoRA-One & \cellcolor{evagold!8}5.74 & \cellcolor{evagold!37}\underline{45.43} & \cellcolor{evagold!38}\textbf{51.37} \\
\midrule
\textbf{ABM (Standard)} & \cellcolor{evagold!27}5.92 & \cellcolor{evagold!38}\textbf{45.53} & \cellcolor{evagold!27}49.51 \\
\textbf{ABM (Ortho.)} & \cellcolor{evagold!38}\textbf{6.03} & \cellcolor{evagold!24}43.92 & \cellcolor{evagold!21}48.51 \\
\bottomrule
\end{tabular*}
\vspace{-5mm}
\end{table}
 
On Qwen2.5-1.5B (Table~\ref{tab:qwen}), \method{} attains the best length-controlled win rate ($44.73$) and the second-best raw win rate, and improves over standard LoRA on all three metrics, raising MT-Bench from $3.58$ to $4.14$. LoRA-One leads on MT-Bench and rsLoRA on raw win rate, but neither matches \method{} once response length is controlled for.
 
On LLaMA2-7B (Table~\ref{tab:llama}), the picture is similar. \method{} takes the best MT-Bench score (ABM (Ortho.), $6.03$) and the best length-controlled win rate (ABM (Standard), $45.53$), while LoRA-One retains the highest raw win rate. Across both backbones, \method{} tops the length-controlled ranking, the metric least confounded by response length~\cite{dubois2024length}, and stays within the leading group on the others. The choice of reference initialization has a visible but modest effect: an orthogonal reference favors MT-Bench, whereas a standard reference gives the strongest length-controlled win rate.

\section{Conclusion}
\label{sec:conclusion}
We proposed ABM-LoRA, a task-informed LoRA initialization method based on the observation that task-induced activation boundaries emerge before activation values and low-rank updates converge. A short task probe extracts these boundaries, and margin-based layer-wise supervision transfers only their activation sides to a fresh adapter, without copying probe parameters or imposing additional constraints during fine-tuning.
We further characterized this process as boundary-supervision lifting, where task-output information is converted into intermediate supervision that reveals boundary-relevant LoRA directions. Across language and vision tasks, ABM improved the compute--performance trade-off over standard LoRA, demonstrating that early functional structure provides an effective source of LoRA initialization beyond pretrained weights and initial task gradients.

\bibliography{abmlora}

@inproceedings{hu2022lora,
  title={LoRA: Low-Rank Adaptation of Large Language Models},
  author={Hu, Edward J and Shen, Yelong and Wallis, Phillip and Allen-Zhu, Zeyuan and Li, Yuanzhi and Wang, Shean and Wang, Lu and Chen, Weizhu},
  booktitle={ICLR},
  year={2022}
}

@inproceedings{wang2024lora,
  title={LoRA-GA: Low-Rank Adaptation with Gradient Approximation},
  author={Wang, Shaowen and Yu, Linxi and Li, Jian},
  booktitle={NeurIPS},
  year={2024}
}

@article{kalajdzievski2023rank,
  title={A Rank Stabilization Scaling Factor for Fine-Tuning with LoRA},
  author={Kalajdzievski, Damjan},
  journal={arXiv preprint arXiv:2312.03732},
  year={2023}
}

@inproceedings{meng2024pissa,
  title={PiSSA: Principal Singular Values and Singular Vectors Adaptation of Large Language Models},
  author={Meng, Fanxu and Wang, Zhaohui and Zhang, Muhan},
  booktitle={NeurIPS},
  year={2024}
}

@article{chen2025hrp,
  title={HRP: High-Rank Preheating for Superior LoRA Initialization},
  author={Chen, Yuzhu and Wang, Yingjie and Fu, Shi and Shen, Li and Jing, Yongcheng and Tian, Xinmei and Tao, Dacheng},
  journal={arXiv preprint arXiv:2502.07739},
  year={2025}
}

@inproceedings{li2025beyond,
  title={Beyond Zero Initialization: Investigating the Impact of Non-Zero Initialization on LoRA Fine-Tuning Dynamics},
  author={Li, Shiwei and Luo, Xiandi and Tang, Xing and Wang, Haozhao and Chen, Hao and Luo, Weihong and Li, Yuhua and He, Xiuqiang and Li, Ruixuan},
  booktitle={ICML},
  year={2025}
}

@article{das2025consnotrainlora,
  title={ConsNoTrainLoRA: Data-Driven Weight Initialization of Low-Rank Adapters Using Constraints},
  author={Das, Debasmit and Park, Hyoungwoo and Hayat, Munawar and Choi, Seokeon and Yun, Sungrack and Porikli, Fatih},
  journal={arXiv preprint arXiv:2507.08044},
  year={2025}
}

@article{wang2025activation,
  title={Activation-Guided Low-Rank Parameter Adaptation for Efficient Model Fine-Tuning},
  author={Wang, Qingchen and Shen, Shengyu},
  journal={IEEE Access},
  year={2025}
}

@inproceedings{heo2019knowledge,
  title={Knowledge Transfer via Distillation of Activation Boundaries Formed by Hidden Neurons},
  author={Heo, Byeongho and Lee, Minsik and Yun, Sangdoo and Choi, Jin Young},
  booktitle={AAAI},
  year={2019}
}

@inproceedings{romero2015fitnets,
  title={FitNets: Hints for Thin Deep Nets},
  author={Romero, Adriana and Ballas, Nicolas and Kahou, Samira Ebrahimi and Chassang, Antoine and Gatta, Carlo and Bengio, Yoshua},
  booktitle={ICLR},
  year={2015}
}

@inproceedings{yim2017gift,
  title={A Gift from Knowledge Distillation: Fast Optimization, Network Minimization and Transfer Learning},
  author={Yim, Junho and Joo, Donggyu and Bae, Jihoon and Kim, Junmo},
  booktitle={CVPR},
  year={2017}
}

@inproceedings{zagoruyko2017paying,
  title={Paying More Attention to Attention: Improving the Performance of Convolutional Neural Networks via Attention Transfer},
  author={Zagoruyko, Sergey and Komodakis, Nikos},
  booktitle={ICLR},
  year={2017}
}

@inproceedings{montufar2014number,
  title={On the Number of Linear Regions of Deep Neural Networks},
  author={Mont{\'u}far, Guido and Pascanu, Razvan and Cho, Kyunghyun and Bengio, Yoshua},
  booktitle={NeurIPS},
  year={2014}
}

@article{pascanu2013number,
  title={On the Number of Response Regions of Deep Feed Forward Networks with Piece-Wise Linear Activations},
  author={Pascanu, Razvan and Montufar, Guido and Bengio, Yoshua},
  journal={arXiv preprint arXiv:1312.6098},
  year={2013}
}

@inproceedings{devlin2019bert,
  title={BERT: Pre-Training of Deep Bidirectional Transformers for Language Understanding},
  author={Devlin, Jacob and Chang, Ming-Wei and Lee, Kenton and Toutanova, Kristina},
  booktitle={NAACL-HLT},
  year={2019}
}

@inproceedings{brown2020language,
  title={Language Models Are Few-Shot Learners},
  author={Brown, Tom and Mann, Benjamin and Ryder, Nick and Subbiah, Melanie and Kaplan, Jared D and Dhariwal, Prafulla and Neelakantan, Arvind and Shyam, Pranav and Sastry, Girish and Askell, Amanda},
  booktitle={NeurIPS},
  year={2020}
}

@article{raffel2020exploring,
  title={Exploring the Limits of Transfer Learning with a Unified Text-to-Text Transformer},
  author={Raffel, Colin and Shazeer, Noam and Roberts, Adam and Lee, Katherine and Narang, Sharan and Matena, Michael and Zhou, Yanqi and Li, Wei and Liu, Peter J},
  journal={Journal of Machine Learning Research},
  volume={21},
  number={140},
  pages={1--67},
  year={2020}
}

@article{dosovitskiy2020image,
  title={An Image Is Worth 16x16 Words: Transformers for Image Recognition at Scale},
  author={Dosovitskiy, Alexey and Beyer, Lucas and Kolesnikov, Alexander and Weissenborn, Dirk and Zhai, Xiaohua and Unterthiner, Thomas and Dehghani, Mostafa and Minderer, Matthias and Heigold, Georg and Gelly, Sylvain and Uszkoreit, Jakob and Houlsby, Neil},
  journal={arXiv preprint arXiv:2010.11929},
  year={2020}
}

@article{touvron2023llama,
  title={LLaMA 2: Open Foundation and Fine-Tuned Chat Models},
  author={Touvron, Hugo and Martin, Louis and Stone, Kevin and Albert, Peter and Almahairi, Amjad and Babaei, Yasmine},
  journal={arXiv preprint arXiv:2307.09288},
  year={2023}
}

@inproceedings{chen2022adaptformer,
  title={AdaptFormer: Adapting Vision Transformers for Scalable Visual Recognition},
  author={Chen, Shoufa and Ge, Chongjian and Tong, Zhan and Wang, Jiangliu and Song, Yibing and Wang, Jue and Luo, Ping},
  booktitle={NeurIPS},
  year={2022}
}

@inproceedings{jia2022visual,
  title={Visual Prompt Tuning},
  author={Jia, Menglin and Tang, Luming and Chen, Bor-Chun and Cardie, Claire and Belongie, Serge and Hariharan, Bharath and Lim, Ser-Nam},
  booktitle={ECCV},
  year={2022}
}

@inproceedings{xu2024wizardlm,
  title={WizardLM: Empowering Large Pre-Trained Language Models to Follow Complex Instructions},
  author={Xu, Can and Sun, Qingfeng and Zheng, Kai and Geng, Xiubo and Zhao, Pu and Feng, Jiazhan and Tao, Chongyang and Jiang, Daxin},
  booktitle={ICLR},
  year={2024}
}

@misc{taori2023alpaca,
  title={Stanford Alpaca: An Instruction-Following LLaMA Model},
  author={Taori, Rohan and Gulrajani, Ishaan and Zhang, Tianyi and Dubois, Yann and Li, Xuechen and Guestrin, Carlos and Liang, Percy and Hashimoto, Tatsunori B},
  year={2023},
  howpublished={\url{https://github.com/tatsu-lab/stanford_alpaca}},
  note={Accessed: 2026-07-10}
}

@inproceedings{zheng2023judging,
  title={Judging LLM-as-a-Judge with MT-Bench and Chatbot Arena},
  author={Zheng, Lianmin and Chiang, Wei-Lin and Sheng, Ying and Zhuang, Siyuan and Wu, Zhanghao and Zhuang, Yonghao and Lin, Zi and Li, Zhuohan and Li, Dacheng and Xing, Eric P},
  booktitle={NeurIPS},
  year={2023}
}

@article{dubois2024length,
  title={Length-Controlled AlpacaEval: A Simple Way to Debias Automatic Evaluators},
  author={Dubois, Yann and Galambosi, Bal{\'a}zs and Liang, Percy and Hashimoto, Tatsunori B},
  journal={arXiv preprint arXiv:2404.04475},
  year={2024}
}

@article{buyukakyuz2024olora,
  title={Olora: Orthonormal low-rank adaptation of large language models},
  author={B{\"u}y{\"u}kaky{\"u}z, Kerim},
  journal={arXiv preprint arXiv:2406.01775},
  year={2024}
}

@article{paischer2026parameter,
  title={Parameter efficient fine-tuning via explained variance adaptation},
  author={Paischer, Fabian and Hauzenberger, Lukas and Schmied, Thomas and Alkin, Benedikt and Deisenroth, Marc and Hochreiter, Sepp},
  journal={Advances in Neural Information Processing Systems},
  volume={38},
  pages={46215--46267},
  year={2026}
}

@inproceedings{liu2022convnet,
  title     = {A ConvNet for the 2020s},
  author    = {Liu, Zhuang and Mao, Hanzi and Wu, Chao-Yuan and Feichtenhofer, Christoph and Darrell, Trevor and Xie, Saining},
  booktitle = {CVPR},
  pages     = {11976--11986},
  year      = {2022}
}

@inproceedings{liu2021swin,
  title     = {Swin Transformer: Hierarchical Vision Transformer using Shifted Windows},
  author    = {Liu, Ze and Lin, Yutong and Cao, Yue and Hu, Han and Wei, Yixuan and Zhang, Zheng and Lin, Stephen and Guo, Baining},
  booktitle = {ICCV},
  pages     = {10012--10022},
  year      = {2021}
}

@inproceedings{nilsback2008automated,
  title     = {Automated Flower Classification over a Large Number of Classes},
  author    = {Nilsback, Maria-Elena and Zisserman, Andrew},
  booktitle = {ICVGIP},
  pages     = {722--729},
  year      = {2008}
}

@inproceedings{parkhi2012cats,
  title     = {Cats and Dogs},
  author    = {Parkhi, Omkar M. and Vedaldi, Andrea and Zisserman, Andrew and Jawahar, C. V.},
  booktitle = {CVPR},
  pages     = {3498--3505},
  year      = {2012}
}

@inproceedings{bossard2014food,
  title     = {Food-101 -- Mining Discriminative Components with Random Forests},
  author    = {Bossard, Lukas and Guillaumin, Matthieu and Van Gool, Luc},
  booktitle = {ECCV},
  pages     = {446--461},
  year      = {2014}
}

@article{maji2013fine,
  title   = {Fine-Grained Visual Classification of Aircraft},
  author  = {Maji, Subhransu and Rahtu, Esa and Kannala, Juho and Blaschko, Matthew and Vedaldi, Andrea},
  journal = {arXiv preprint arXiv:1306.5151},
  year    = {2013}
}

@inproceedings{wang2019glue,
  title     = {{GLUE}: A Multi-Task Benchmark and Analysis Platform for Natural Language Understanding},
  author    = {Wang, Alex and Singh, Amanpreet and Michael, Julian and Hill, Felix and Levy, Omer and Bowman, Samuel R.},
  booktitle = {ICLR},
  year      = {2019}
}

@article{touvron2023llama2,
  title   = {Llama 2: Open Foundation and Fine-Tuned Chat Models},
  author  = {Touvron, Hugo and Martin, Louis and Stone, Kevin and others},
  journal = {arXiv preprint arXiv:2307.09288},
  year    = {2023}
}

@article{yang2024qwen25,
  title   = {Qwen2.5 Technical Report},
  author  = {Yang, An and Yang, Baosong and Zhang, Beichen and Hui, Binyuan and Zheng, Bo and Yu, Bowen and Li, Chengyuan and Liu, Dayiheng and Huang, Fei and Wei, Haoran and others},
  journal = {arXiv preprint arXiv:2412.15115},
  year    = {2024}
}

@article{zhang2025lora,
  title={Lora-one: One-step full gradient could suffice for fine-tuning large language models, provably and efficiently},
  author={Zhang, Yuanhe and Liu, Fanghui and Chen, Yudong},
  journal={arXiv preprint arXiv:2502.01235},
  year={2025}
}

@inproceedings{loshchilov2019decoupled,
  title     = {Decoupled Weight Decay Regularization},
  author    = {Loshchilov, Ilya and Hutter, Frank},
  booktitle = {ICLR},
  year      = {2019}
}

@inproceedings{loshchilov2017sgdr,
  title     = {{SGDR}: Stochastic Gradient Descent with Warm Restarts},
  author    = {Loshchilov, Ilya and Hutter, Frank},
  booktitle = {ICLR},
  year      = {2017}
}

@misc{li2023alpacaeval,
  title        = {{AlpacaEval}: An Automatic Evaluator of Instruction-Following Models},
  author       = {Li, Xuechen and Zhang, Tianyi and Dubois, Yann and Taori, Rohan and Gulrajani, Ishaan and Guestrin, Carlos and Liang, Percy and Hashimoto, Tatsunori B.},
  howpublished = {\url{https://github.com/tatsu-lab/alpaca_eval}},
  year         = {2023}
}

@inproceedings{dosovitskiy2021vit,
  title     = {An Image is Worth 16x16 Words: Transformers for Image Recognition at Scale},
  author    = {Dosovitskiy, Alexey and Beyer, Lucas and Kolesnikov, Alexander and
               Weissenborn, Dirk and Zhai, Xiaohua and Unterthiner, Thomas and
               Dehghani, Mostafa and Minderer, Matthias and Heigold, Georg and
               Gelly, Sylvain and Uszkoreit, Jakob and Houlsby, Neil},
  booktitle = {International Conference on Learning Representations (ICLR)},
  year      = {2021}
}

@article{zhai2019vtab,
  title   = {A Large-scale Study of Representation Learning with the
             Visual Task Adaptation Benchmark},
  author  = {Zhai, Xiaohua and Puigcerver, Joan and Kolesnikov, Alexander and
             Ruyssen, Pierre and Riquelme, Carlos and Lucic, Mario and
             Djolonga, Josip and Pinto, Andre Susano and Neumann, Maxim and
             Dosovitskiy, Alexey and Beyer, Lucas and Bachem, Olivier and
             Tschannen, Michael and Michalski, Marcin and Bousquet, Olivier and
             Gelly, Sylvain and Houlsby, Neil},
  journal = {arXiv preprint arXiv:1910.04867},
  year    = {2019}
}

@inproceedings{deng2009imagenet,
  title     = {ImageNet: A Large-Scale Hierarchical Image Database},
  author    = {Deng, Jia and Dong, Wei and Socher, Richard and Li, Li-Jia and
               Li, Kai and Fei-Fei, Li},
  booktitle = {IEEE Conference on Computer Vision and Pattern Recognition (CVPR)},
  year      = {2009}
}

@article{maji2013aircraft,
  title   = {Fine-Grained Visual Classification of Aircraft},
  author  = {Maji, Subhransu and Rahtu, Esa and Kannala, Juho and
             Blaschko, Matthew and Vedaldi, Andrea},
  journal = {arXiv preprint arXiv:1306.5151},
  year    = {2013}
}

@inproceedings{nilsback2008flowers,
  title     = {Automated Flower Classification over a Large Number of Classes},
  author    = {Nilsback, Maria-Elena and Zisserman, Andrew},
  booktitle = {Indian Conference on Computer Vision, Graphics and Image Processing (ICVGIP)},
  year      = {2008}
}

@inproceedings{bossard2014food101,
  title     = {Food-101 -- Mining Discriminative Components with Random Forests},
  author    = {Bossard, Lukas and Guillaumin, Matthieu and Van Gool, Luc},
  booktitle = {European Conference on Computer Vision (ECCV)},
  year      = {2014}
}

@inproceedings{parkhi2012pets,
  title     = {Cats and Dogs},
  author    = {Parkhi, Omkar M. and Vedaldi, Andrea and Zisserman, Andrew and
               Jawahar, C. V.},
  booktitle = {IEEE Conference on Computer Vision and Pattern Recognition (CVPR)},
  year      = {2012}
}

\clearpage
\newpage
\appendix

\setcounter{section}{0} 
\setcounter{figure}{0}
\renewcommand{\thefigure}{A\arabic{figure}}

\setcounter{table}{0}
\renewcommand{\thetable}{A\arabic{table}}

\setcounter{equation}{0}
\renewcommand{\theequation}{A\arabic{equation}}

\setcounter{proposition}{0}
\renewcommand{\theproposition}{A\arabic{proposition}}

\setcounter{corollary}{0}
\renewcommand{\thecorollary}{A\arabic{corollary}}

\section{Proofs for Boundary-Supervision Lifting}
\label{app:theoretical_proofs}

This section provides the complete proofs for the theoretical results in
Section~\ref{sec:boundary_analysis}.
We first derive the local sensitivity induced by the boundary objective and
then prove Proposition.

\paragraph{Local perturbations of a LoRA adapter.}
Holding all parameters other than $\phi_l$ fixed, we write the model output
locally as
\begin{equation}
    o(\phi_l)
    =
    h_l\!\left(z_l(\phi_l)\right),
    \label{eq:output_composition}
\end{equation}
where $h_l$ denotes the downstream computation from the layer-$l$
pre-activation to the final model output.
We define
\begin{equation}
    J_l
    =
    \frac{\partial z_l}{\partial\phi_l},
    \qquad
    H_l
    =
    \frac{\partial h_l}{\partial z_l}.
\end{equation}

Before stating our result, we clarify how a change in the LoRA parameters
affects an intermediate activation and the final model output.
Consider a small perturbation
$\phi_l \mapsto \phi_l+\epsilon v$,
where $v$ is an arbitrary direction in the parameter space of the LoRA
adapter at layer $l$, and $\epsilon>0$ is sufficiently small.
By first-order Taylor expansion,
\begin{equation}
    z_l(\phi_l+\epsilon v)
    =
    z_l(\phi_l)
    +
    \epsilon J_l v
    +
    o(\epsilon),
    \qquad
    J_l=\frac{\partial z_l}{\partial\phi_l}.
    \label{eq:local_activation_change}
\end{equation}
Thus, $J_lv$ represents the first-order change in the layer-$l$
pre-activation induced by moving the LoRA parameters along $v$.

This activation change reaches the model output only after passing through
the downstream network. Let
$H_l=\partial o/\partial z_l$ denote the downstream Jacobian. Then
\begin{equation}
    o(\phi_l+\epsilon v)
    =
    o(\phi_l)
    +
    \epsilon H_lJ_lv
    +
    o(\epsilon).
    \label{eq:local_output_change}
\end{equation}
Therefore, output-level task supervision observes the parameter direction
$v$ only through the transformed quantity $H_lJ_lv$.
A direction may substantially alter an intermediate activation
($J_lv\neq 0$) while producing little or no first-order change in the output
($H_lJ_lv\approx 0$).
This does not imply that the direction is globally irrelevant to the task;
it only means that it is weakly observable through the current downstream
mapping.

ABM attaches supervision directly to the intermediate pre-activation.
Consequently, it can observe whether the same perturbation moves an
unmatched activation toward its target side without requiring this change
to remain visible after propagation

through all subsequent layers.
We formalize this difference below.

\begin{proposition}[Direct observability of boundary-relevant directions]
\label{prop:boundary_observability}
Consider a monitored layer $l$ and a parameter perturbation direction $v$.
Let
\[
    D_l
    =
    \operatorname{Diag}
    \left(
        \indicator
        \left[
            m\mathbf 1-\tau_l\odot z_l>0
        \right]
    \right)
\]
denote the mask of boundary coordinates whose margin constraints are
currently violated.
If
\begin{equation}
    H_lJ_lv=0
    \qquad\text{and}\qquad
    D_lJ_lv\neq0,
    \label{eq:observability_condition}
\end{equation}
then $v$ is locally unobservable through the model output but observable
through the active boundary constraints.
\end{proposition}

\begin{proof}
For notational simplicity, we suppress the dependence on the input and
consider all monitored coordinates at layer $l$ as a single vector.
Let
\begin{equation}
    S_l
    =
    \operatorname{Diag}(\tau_l),
    \label{eq:sign_matrix}
\end{equation}
where each entry of $\tau_l$ belongs to $\{-1,+1\}$.
The signed margin residual associated with the boundary-matching objective is
\begin{equation}
    r_l(\phi_l)
    =
    \left[
        m\mathbf{1}
        -
        S_l z_l(\phi_l)
    \right]_+,
    \label{eq:boundary_residual}
\end{equation}
where $[\cdot]_+$ denotes the element-wise positive part.
A coordinate has a nonzero residual precisely when its target-side margin
constraint is violated.

Let
\begin{equation}
    D_l
    =
    \operatorname{Diag}
    \left(
        \indicator
        \left[
            m\mathbf{1}
            -
            S_lz_l(\phi_l)
            >
            0
        \right]
    \right)
    \label{eq:active_boundary_mask}
\end{equation}
be the diagonal mask of the currently active boundary constraints.
We assume that no coordinate lies exactly at the hinge point,
$m-\tau_{l,j}z_{l,j}=0$.
The same result can be stated using directional derivatives at such
nondifferentiable points.

Consider the perturbed adapter parameters
$\phi_l+\epsilon v$.
By the first-order expansion in
Equation~\eqref{eq:local_activation_change},
\begin{equation}
    z_l(\phi_l+\epsilon v)
    =
    z_l(\phi_l)
    +
    \epsilon J_lv
    +
    o(\epsilon).
    \label{eq:proof_activation_perturbation}
\end{equation}
For a sufficiently small $\epsilon$, the active set remains unchanged.
Applying the element-wise hinge function therefore gives
\begin{align}
    r_l(\phi_l+\epsilon v)
    &=
    \left[
        m\mathbf{1}
        -
        S_lz_l(\phi_l+\epsilon v)
    \right]_+
    \nonumber\\
    &=
    r_l(\phi_l)
    -
    \epsilon D_lS_lJ_lv
    +
    o(\epsilon).
    \label{eq:boundary_residual_perturbation}
\end{align}
Consequently, the first-order change in the boundary residual is
\begin{equation}
    \delta r_l[v]
    =
    -D_lS_lJ_lv.
    \label{eq:boundary_residual_direction}
\end{equation}

Because $D_l$ and $S_l$ are diagonal, they commute. Moreover,
$S_l^\top S_l=I$ because every diagonal entry of $S_l$ is either $-1$ or
$+1$. It follows that
\begin{align}
    \left\|
        D_lS_lJ_lv
    \right\|_2^2
    &=
    (J_lv)^\top
    S_l^\top D_l^\top D_lS_l
    (J_lv)
    \nonumber\\
    &=
    (J_lv)^\top D_l(J_lv)
    \nonumber\\
    &=
    \left\|
        D_lJ_lv
    \right\|_2^2.
    \label{eq:sign_preserves_norm}
\end{align}
Therefore,
\begin{equation}
    D_lJ_lv\neq0
    \quad\Longrightarrow\quad
    \delta r_l[v]\neq0.
    \label{eq:boundary_observable}
\end{equation}
Thus, the active boundary residuals are locally sensitive to the parameter
direction $v$.

We now consider the final model output.
From Equation~\eqref{eq:local_output_change}, the same parameter perturbation
produces
\begin{equation}
    o(\phi_l+\epsilon v)
    =
    o(\phi_l)
    +
    \epsilon H_lJ_lv
    +
    o(\epsilon).
    \label{eq:proof_output_perturbation}
\end{equation}
Under the condition $H_lJ_lv=0$, this becomes
\begin{equation}
    o(\phi_l+\epsilon v)
    =
    o(\phi_l)
    +
    o(\epsilon).
    \label{eq:output_unobservable}
\end{equation}
Hence, the output has no first-order sensitivity to $v$.

Equivalently, for an output-level task loss with
$q=\nabla_o\mathcal L_{\mathrm{task}}$, its directional derivative is
\begin{equation}
    \mathrm{d}\mathcal L_{\mathrm{task}}[v]
    =
    q^\top H_lJ_lv
    =
    0.
    \label{eq:task_directional_derivative_zero}
\end{equation}
The direction is therefore locally unobservable through the output-level
objective.

Combining Equations~\eqref{eq:boundary_observable} and
\eqref{eq:output_unobservable}, if
\[
    H_lJ_lv=0
    \qquad\text{but}\qquad
    D_lJ_lv\neq0,
\]
then the perturbation produces no first-order change in the final output
while producing a nonzero first-order change in at least one active boundary
residual.
Accordingly, $v$ is locally unobservable through the model output but
observable through the active boundary constraints.
\end{proof}

\begin{corollary}[Sensitivity under downstream attenuation]
\label{cor:downstream_attenuation}
Suppose that the downstream Jacobian is locally contractive on the active
boundary subspace, such that
\begin{equation}
    \|H_lu\|_2
    \leq
    \kappa_l\|u\|_2,
    \qquad
    0\leq\kappa_l<1,
\end{equation}
for every activation perturbation $u$ supported on the active boundary
coordinates.
Then, for any parameter direction $v$ satisfying
$D_lJ_lv=J_lv$,
\begin{equation}
    \|H_lJ_lv\|_2^2
    \leq
    \kappa_l^2
    \|D_lJ_lv\|_2^2.
    \label{eq:attenuated_sensitivity}
\end{equation}
\end{corollary}

\begin{proof}
Since $D_lJ_lv=J_lv$, the activation perturbation $J_lv$ lies entirely in
the active boundary subspace. Applying the contractivity assumption with
$u=J_lv$ gives
\begin{equation}
    \|H_lJ_lv\|_2
    \leq
    \kappa_l\|J_lv\|_2
    =
    \kappa_l\|D_lJ_lv\|_2.
\end{equation}
Squaring both sides yields
Equation~\eqref{eq:attenuated_sensitivity}.
\end{proof}

For the multi-layer analysis, we use the complete collection of LoRA
parameters $\phi$ as the common parameter space and define
\begin{equation}
    \widetilde J_l
    =
    \frac{\partial z_l}{\partial\phi}.
    \label{eq:global_activation_jacobian}
\end{equation}
Unlike the local Jacobian $J_l=\partial z_l/\partial\phi_l$ used in the
single-layer analysis, $\widetilde J_l$ includes the dependence of $z_l$ on
all LoRA modules that can affect it.

\begin{proposition}[Monotonic expansion of boundary observability]
\label{prop:multilayer_observability}
For a set of monitored layers $\mathcal M$, define
\begin{equation}
    K_{\mathcal M}^{\mathrm{bdry}}
    =
    \sum_{l\in\mathcal M}
    \lambda_l\widetilde J_l^\top D_l\widetilde J_l,
    \qquad
    \lambda_l>0.
    \label{eq:multilayer_boundary_matrix}
\end{equation}
Adding another monitored layer $k\notin\mathcal M$ yields
\begin{equation}
    K_{\mathcal M\cup\{k\}}^{\mathrm{bdry}}
    \succeq
    K_{\mathcal M}^{\mathrm{bdry}}.
    \label{eq:monotonic_observability}
\end{equation}
Moreover,
\begin{equation}
    \operatorname{Null}
    \left(
        K_{\mathcal M}^{\mathrm{bdry}}
    \right)
    =
    \bigcap_{l\in\mathcal M}
    \operatorname{Null}
    \left(
        D_l\widetilde J_l
    \right).
    \label{eq:boundary_nullspace}
\end{equation}
\end{proposition}

\begin{proof}
For any vector $v$,
\begin{equation}
    v^\top \widetilde J_l^\top D_l\widetilde J_lv
    =
    \|D_l\widetilde J_lv\|_2^2
    \geq0,
\end{equation}
where we used $D_l^\top D_l=D_l$.
Therefore, every matrix $\widetilde J_l^\top D_l\widetilde J_l$ is positive semidefinite.
It follows that
\begin{equation}
    K_{\mathcal M\cup\{k\}}^{\mathrm{bdry}}
    -
    K_{\mathcal M}^{\mathrm{bdry}}
    =
    \lambda_k\,\widetilde J_k^\top D_k\widetilde J_k
    \succeq0,
\end{equation}
which proves Equation~\eqref{eq:monotonic_observability}.

Furthermore,
\begin{align}
    v^\top K_{\mathcal M}^{\mathrm{bdry}}v
    &=
    \sum_{l\in\mathcal M}
    \lambda_l
    \left\|
        D_l\widetilde J_lv
    \right\|_2^2 .
\end{align}
Because every term is nonnegative and every $\lambda_l$ is strictly positive,
the sum equals zero if and only if
$D_l\widetilde J_lv=0$ for every $l\in\mathcal M$.
This proves Equation~\eqref{eq:boundary_nullspace}.
\end{proof}

\section{Additional Experiments}

%
%
%


\subsection{Reference Cost and Probe Length}
\label{sec:cost}

ABM trains a reference adapter before initializing the target adapter, which
raises two questions. First, are the reported gains simply a consequence of the
additional compute? Second, if the reference must be trained to convergence, is
ABM meaningfully an \emph{initialization} method at all? We address both with a
single set of experiments on T5-base.

\paragraph{Cost accounting.}
The total cost of ABM is $T_p + T_m + T_{ft}$, where $T_p$ is reference (probe)
training, $T_m$ is boundary matching, and $T_{ft}$ is downstream fine-tuning.
On GLUE, boundary matching consumes 100 examples with a single-example batch,
which amounts to between $0.14\%$ and $1.57\%$ of $T_{ft}$ in training FLOPs
across the tasks we consider; the vision experiments use a longer matching
schedule and a larger batch, so the corresponding fraction there is larger.
Matching also applies no label supervision: the objective
depends only on the sign structure of pre-activations, not on task labels. The
overhead of ABM on GLUE is therefore effectively $T_p$ alone, and the question
of cost reduces to a single controllable quantity---how long the probe must run.

\paragraph{The gain is not extra compute.}
Table~\ref{tab:cost-matched} compares ABM against LoRA and against a
compute-matched control (LoRA-2T) trained for twice as many steps. ABM
consumes $2.01\times$ the budget of LoRA and LoRA-2T consumes
$2.00\times$, so the two are matched to within one percent. Under this matched
budget ABM improves the average by $0.38$ points, exceeding LoRA-2T on CoLA
($+1.34$) and QNLI ($+0.22$), matching it on SST-2, and trailing by $0.04$ on
MNLI. Because the budgets are nearly identical, the improvement cannot be
attributed to additional training.

We note that the $2.29$-point average improvement over LoRA is
dominated by CoLA ($+8.14$); the remaining three tasks improve by $0.34$ points
on average. The gain is nonetheless positive on every task, ranging from $+0.22$
on QNLI to $+8.14$ on CoLA, and CoLA is both the smallest task and the one
scored by the most variance-sensitive metric.

\begin{table}[t]
\centering
\setlength{\tabcolsep}{4pt}
\caption{Compute-matched comparison on GLUE with T5-base. Cost is training
FLOPs normalized to LoRA. LoRA-2T is LoRA trained for twice
as many steps, matching ABM's budget to within one percent. Scores are Matthews
correlation for CoLA and accuracy elsewhere. \textbf{Bold} marks the best per
column; where the best value is tied, every tied row is set in bold.}
\label{tab:cost-matched}
\begin{tabular}{@{}lcccccc@{}}
\toprule
Method & Cost & CoLA & SST-2 & MNLI & QNLI & Avg \\
\midrule
LoRA & $1.00\times$ & 69.42 & 93.81 & 84.07 & 92.84 & 85.04 \\
LoRA-2T & $2.00\times$ & 76.22 & \textbf{94.04} & \textbf{84.69} & 92.84 & 86.95 \\
\rowcolor{evablue!12}
\textbf{ABM (ours)} & $2.01\times$ & \textbf{77.56} & \textbf{94.04} & 84.65 & \textbf{93.06} & \textbf{87.33} \\
\bottomrule
\end{tabular}
\end{table}

\paragraph{A short probe suffices.}
Table~\ref{tab:probe} varies the probe length $K$ while holding boundary
matching and downstream fine-tuning fixed. Performance saturates well before the
reference converges. At $K{=}50$, costing $1.11\times$, ABM already recovers
$97.7\%$ of the full-reference average; SST-2, MNLI, and QNLI are within $0.5$
points of their full-reference scores at this budget, and only CoLA continues to
improve with a longer probe. At $K{=}200$, costing $1.40\times$, ABM recovers
$99.4\%$ of the full-reference average and comes within $0.10$ points of
LoRA-2T while consuming $30\%$ less compute. Returns are essentially flat
between $K{=}150$ and $K{=}200$ ($86.78 \rightarrow 86.85$).

\begin{figure}[t]
\centering
\includegraphics[width=\columnwidth]{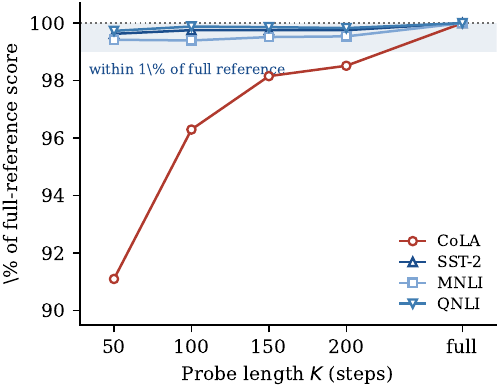}
\caption{Task-level recovery as a function of probe length. Each task is
normalized by its own full-reference score, so tasks on different scales are
comparable. Three of four tasks stay within one percent of the full-reference
result at every probe length, including the shortest; only CoLA benefits
materially from a longer reference.}
\label{fig:recovery}
\end{figure}

Figure~\ref{fig:recovery} shows where the residual gap lives once each task is
normalized by its own full-reference score. SST-2, MNLI, and QNLI remain above
$99\%$ at every probe length, including the shortest; CoLA alone climbs from
$91\%$ at $K{=}50$ to $98\%$ at $K{=}200$. The averaged column of
Table~\ref{tab:probe} is therefore driven almost entirely by CoLA, and on the
remaining three tasks a $50$-step probe---costing $1.11\times$---is already
sufficient. CoLA is the smallest of the four tasks and the only one scored by
Matthews correlation, which is more sensitive to small changes in the decision
boundary than accuracy; a longer probe evidently helps most where the boundary
itself is hardest to place.

This is the behavior the observation in Section~\ref{sec:motivation} predicts.
If boundary structure stabilizes early in adaptation, then a reference adapter
should become a useful source of that structure long before it converges, and
the marginal value of further reference training should decay quickly. Both hold
here. ABM's advantage therefore does not depend on a fully trained reference,
and the method retains its character as an initialization scheme rather than a
distillation pipeline.

\begin{table}[t]
\centering
\caption{Probe length ablation. $K$ is the number of steps of reference
training; boundary matching (100 examples) and downstream fine-tuning are
identical across rows. Cost is training FLOPs normalized to LoRA, and
\% full is the average as a fraction of the full-reference average. Accuracy
saturates by $K{=}200$, at $1.40\times$ cost.}
\label{tab:probe}
\setlength{\tabcolsep}{5pt}
\resizebox{\columnwidth}{!}{%
\begin{tabular}{@{}rccccccc@{}}
\toprule
$K$ & Cost & CoLA & SST-2 & MNLI & QNLI & Avg & \% full \\
\midrule
50 & $1.11\times$ & 70.66 & 93.69 & 84.16 & 92.81 & 85.33 & 97.7 \\
100 & $1.20\times$ & 74.69 & 93.81 & 84.14 & 92.95 & 86.40 & 98.9 \\
150 & $1.30\times$ & 76.13 & 93.81 & 84.24 & 92.93 & 86.78 & 99.4 \\
\rowcolor{evablue!12}
200 & $1.40\times$ & 76.41 & 93.81 & 84.26 & 92.90 & 86.85 & 99.4 \\
\midrule
Full & $2.01\times$ & 77.56 & 94.04 & 84.65 & 93.06 & 87.33 & 100.0 \\
\bottomrule
\end{tabular}%
}
\end{table}

\paragraph{Caveats.}
Cost is reported in training FLOPs. Boundary matching runs with a batch size of
one, so its wall-clock overhead exceeds this estimate because of low device
utilization; the ordering of the rows is unaffected. Reference checkpoints are
drawn from a single run capped at $200$ steps, so the cosine schedule is defined
over that horizon and the learning rate at early checkpoints differs slightly
from a full-horizon reference. Since the reference is used only as a source of
activation boundaries, we expect this effect to be secondary. Finally, we vary
the probe length $K$ only on GLUE; the appendix ablations use a fully trained
reference and vary the number of boundary-matching steps $S$ instead.

\section{Complexity Analysis}
\label{app:complexity}

Table~\ref{tab:abm_init_cost} reports the initialization-phase (Stage~1) cost of
ABM-LoRA on the GLUE dev sets.
Stage~2 corresponds to the LoRA fine-tuning process, which takes from 10 minutes
to 1.5 hours depending on the dataset. By contrast, Stage~1 is a lightweight
step performed once before fine-tuning: it completes in under 20 seconds on
every task, and its peak memory of $1.74$--$1.75$~GB stays below the $2.71$~GB
required during fine-tuning. Stage~1 therefore neither extends the runtime nor
raises the memory ceiling of the pipeline as a whole.

\begin{table*}[t]
\centering
\caption{\textbf{Initialization time and memory usage} on GLUE dev sets.
Init Time and Memory for ABM-LoRA are measured during Stage~1; the LoRA figure
is the peak memory of the fine-tuning stage, which both methods share.
LoRA has no separate initialization phase.}
\label{tab:abm_init_cost}
\small
\setlength{\tabcolsep}{6pt}
\begin{tabular*}{\textwidth}{@{\extracolsep{\fill}} llccccc}
\toprule
\multirow{2}{*}{\textbf{Model}} & \multirow{2}{*}{\textbf{Metric}} & \multicolumn{5}{c}{\textbf{Dataset}} \\
\cmidrule(lr){3-7}
 &  & \textbf{MNLI} & \textbf{SST-2} & \textbf{CoLA} & \textbf{QNLI} & \textbf{MRPC} \\
\midrule
LoRA & Memory (GB) & \multicolumn{5}{c}{2.71} \\
\midrule
\multirow{2}{*}{ABM-LoRA}
& Init Time (s) & 16.95 & 18.04 & 17.30 & 16.65 & 17.05 \\
& Memory (GB) & 1.75 & 1.74 & 1.74 & 1.75 & 1.75 \\
\bottomrule
\end{tabular*}
\end{table*}

\section{Per-Seed Variance on GLUE}
\label{app:seed_std}

Table~\ref{tab:glue_std} reports the per-seed standard deviations behind Table~3
of the main paper. Rows, columns, and highlighting follow that table so the two
can be read together. The Avg column is the unweighted mean of the four task
means and is reported without a deviation, since it is not recoverable from the
per-task deviations alone.

Most of the spread is small relative to the gaps between methods: on CoLA, where
the methods differ by up to $8.25$ points, every deviation is at most $0.51$. On
SST-2 and QNLI the whole column spans $0.61$ and $0.42$ points respectively,
which is comparable in size to the deviations within those columns.

\begin{table*}[t]
\centering
\caption{\textbf{Per-seed variance on GLUE with T5-base.} Each entry is the mean
over three seeds with the standard deviation as a subscript; means and
highlighting match Table~3 of the main paper. \textbf{Bold} and
\underline{underline} mark the best and second-best mean per column, with every
tied row marked. Metrics are Matthews correlation for CoLA and accuracy for the
other tasks.}
\label{tab:glue_std}
\small
\setlength{\tabcolsep}{6pt}
\begin{tabular*}{\textwidth}{@{\extracolsep{\fill}} lccccc}
\toprule
\textbf{Method} & \textbf{CoLA} & \textbf{SST-2} & \textbf{MNLI} & \textbf{QNLI} & \textbf{Avg} \\
\midrule
Normal   & $69.86_{\pm 0.33}$ & $93.85_{\pm 0.14}$ & $84.31_{\pm 0.17}$ & $92.83_{\pm 0.06}$ & 85.21 \\
Gaussian & $73.60_{\pm 0.36}$ & $93.35_{\pm 0.19}$ & $84.23_{\pm 0.07}$ & $92.89_{\pm 0.09}$ & 86.02 \\
PiSSA    & $\underline{77.53}_{\pm 0.09}$ & $93.77_{\pm 0.19}$ & $\mathbf{84.99}_{\pm 0.03}$ & $93.03_{\pm 0.05}$ & $\underline{87.33}$ \\
EVA      & $\mathbf{78.11}_{\pm 0.27}$ & $\underline{93.88}_{\pm 0.24}$ & $\underline{84.96}_{\pm 0.07}$ & $\mathbf{93.25}_{\pm 0.12}$ & $\mathbf{87.55}$ \\
LoRA-One & $75.36_{\pm 0.51}$ & $93.85_{\pm 0.11}$ & $\mathbf{84.99}_{\pm 0.08}$ & $92.97_{\pm 0.09}$ & 86.79 \\
\rowcolor{evablue!12}
\textbf{ABM (ours)} & $77.15_{\pm 0.30}$ & $\mathbf{93.96}_{\pm 0.05}$ & $84.80_{\pm 0.11}$ & $\underline{93.06}_{\pm 0.07}$ & 87.24 \\
\bottomrule
\end{tabular*}
\end{table*}

\section{Stage 1 ABM Loss Dynamics}
\label{app:stage1_loss}
\begin{figure*}[t]
  \centering
  \begin{minipage}[t]{0.32\linewidth}
    \centering
    \includegraphics[width=\linewidth]{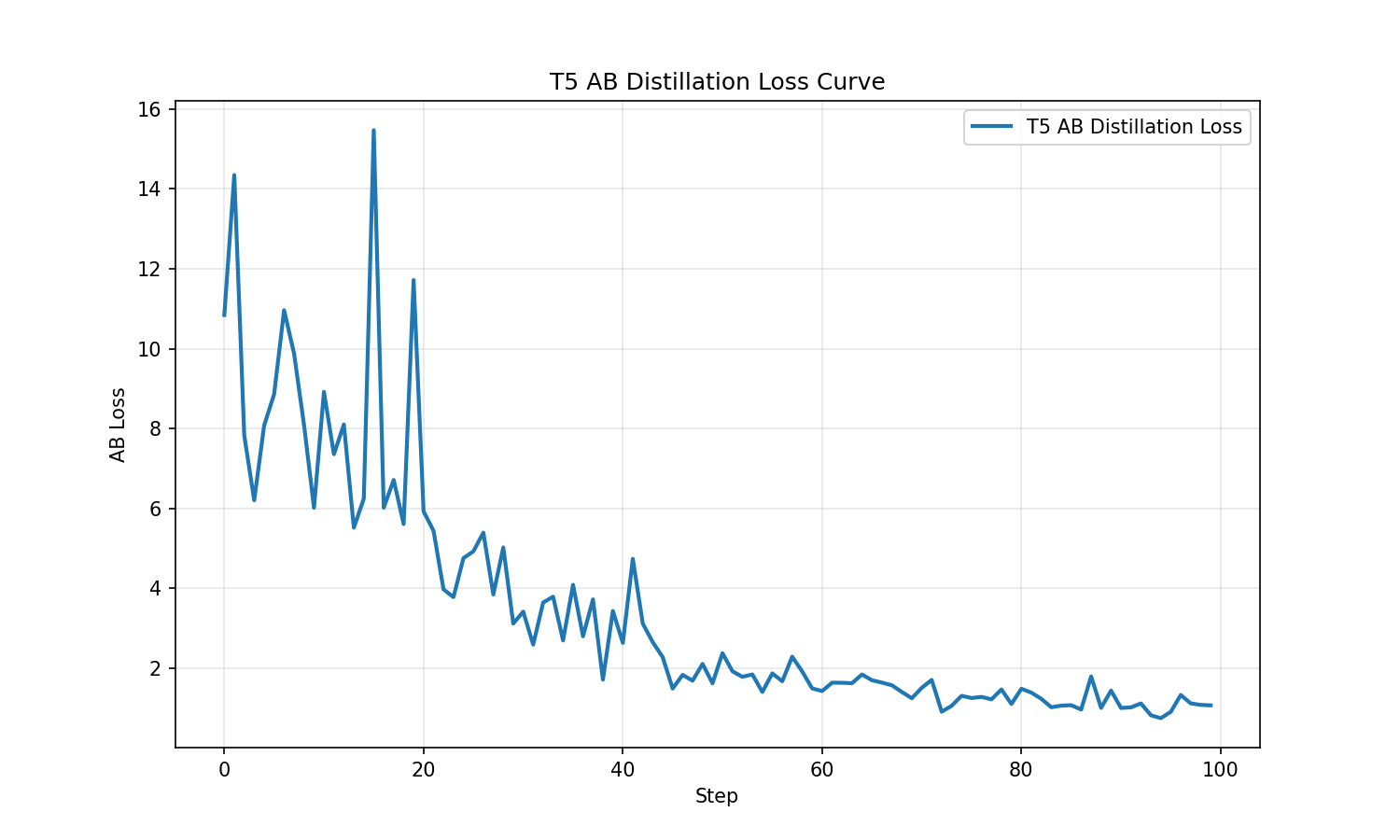}\\[2pt]
    \small MNLI
  \end{minipage}\hfill
  \begin{minipage}[t]{0.32\linewidth}
    \centering
    \includegraphics[width=\linewidth]{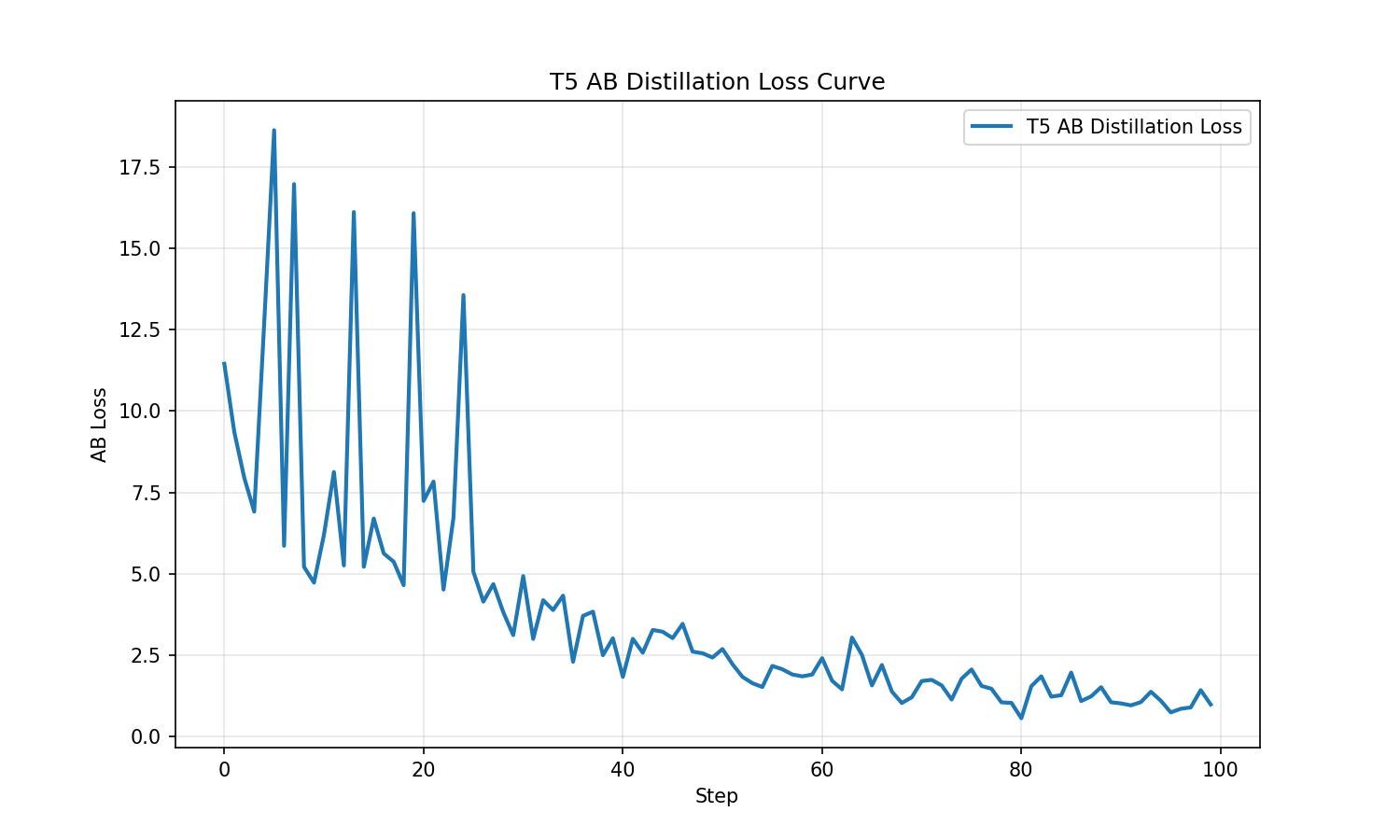}\\[2pt]
    \small SST-2
  \end{minipage}\hfill
  \begin{minipage}[t]{0.32\linewidth}
    \centering
    \includegraphics[width=\linewidth]{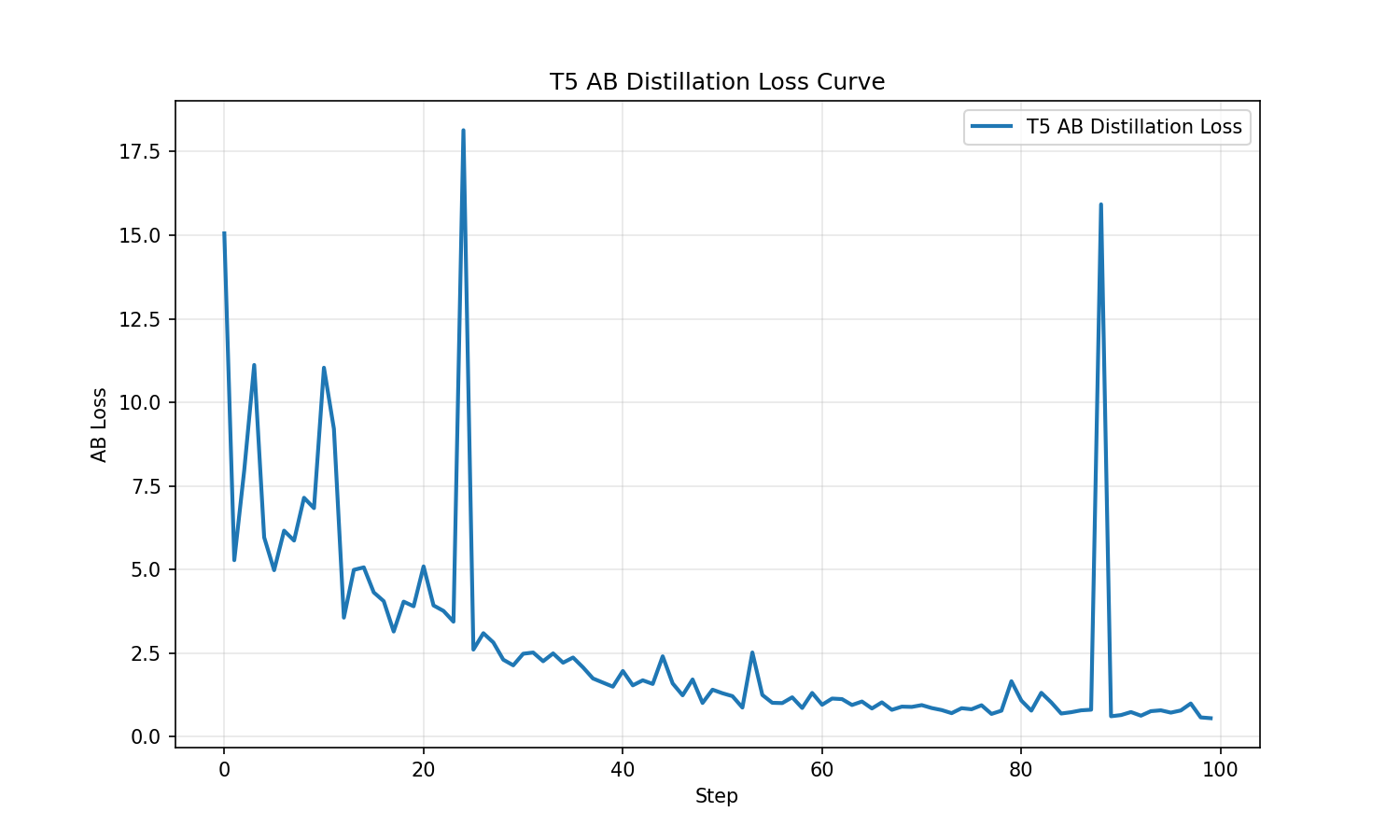}\\[2pt]
    \small CoLA
  \end{minipage}

  \vspace{3mm}

  \begin{minipage}[t]{0.32\linewidth}
    \centering
    \includegraphics[width=\linewidth]{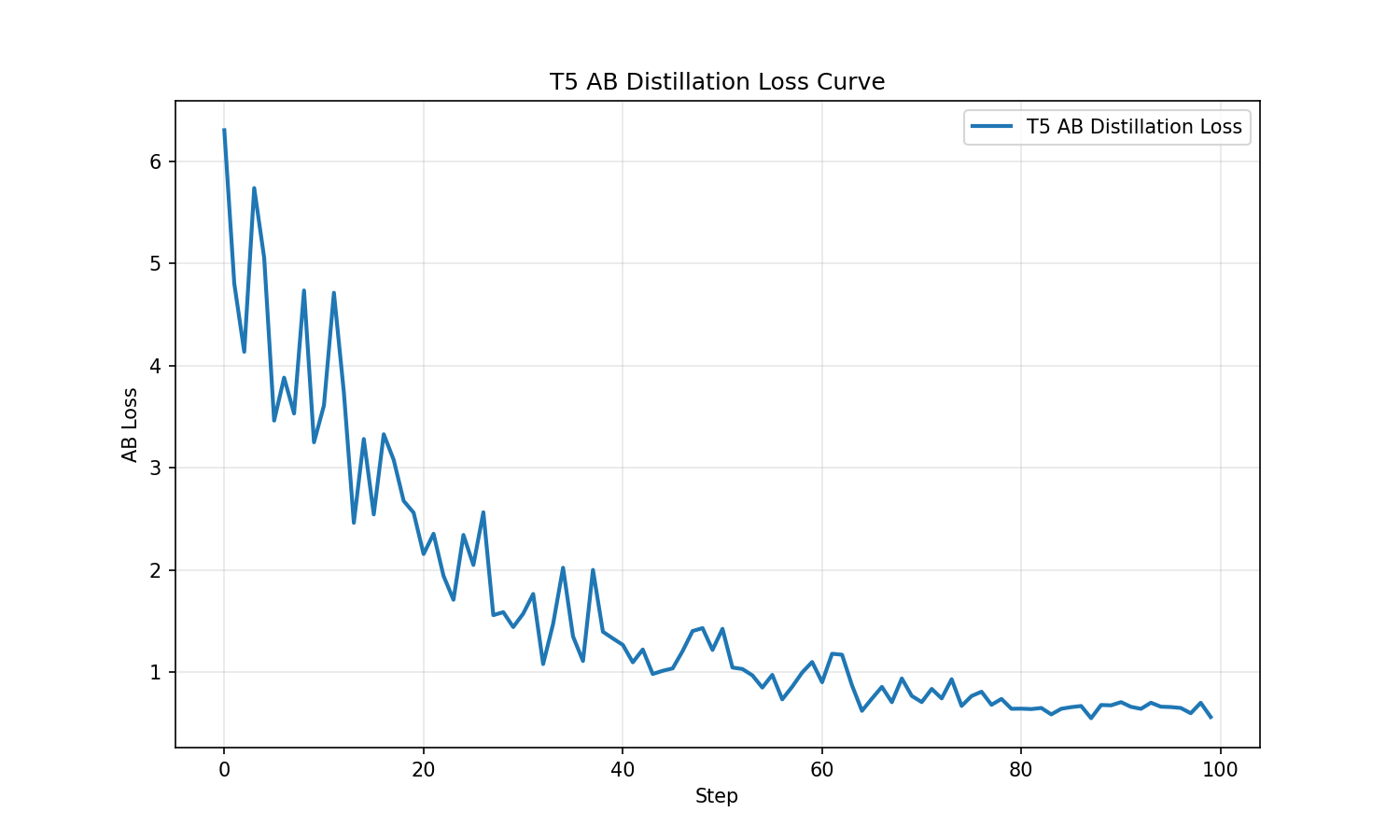}\\[2pt]
    \small QNLI
  \end{minipage}\hspace{0.02\linewidth}
  \begin{minipage}[t]{0.32\linewidth}
    \centering
    \includegraphics[width=\linewidth]{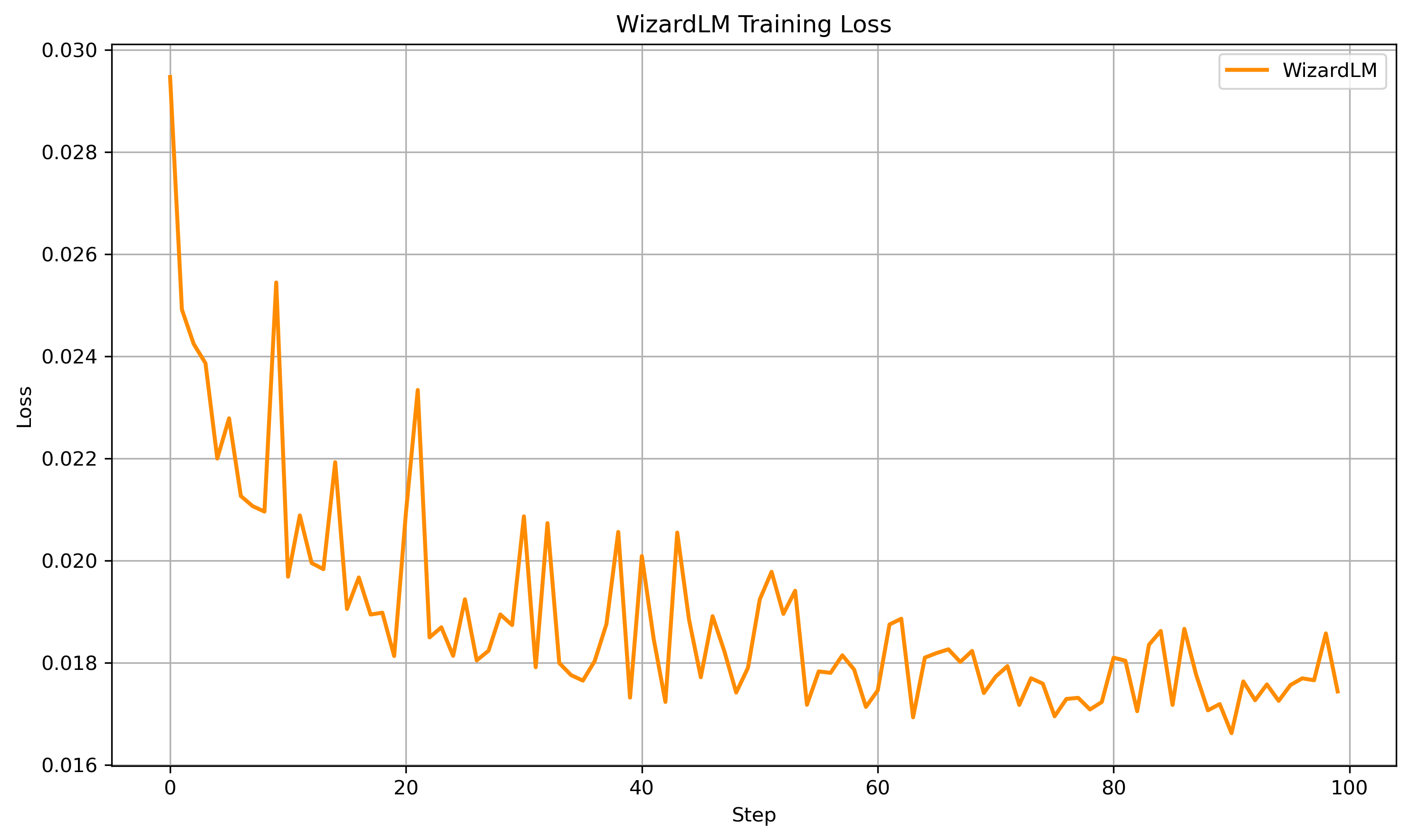}\\[2pt]
    \small WizardLM
  \end{minipage}

  \caption{\textbf{Stage 1 ABM loss evolution} on four GLUE subsets (T5-base:
    MNLI, SST-2, CoLA, QNLI) and the WizardLM dataset (LLaMA2-7B).
    In each case, the ABM loss exhibits a clear and steady downward trend throughout
    Stage 1, confirming that the activation boundary matching objective is being
    effectively optimized.}
  \label{fig:stage1_abm_all}
\end{figure*}
Figure~\ref{fig:stage1_abm_all} illustrates the evolution of the Stage~1 ABM loss
on four distinct subsets of the T5 GLUE benchmark and
the WizardLM dataset on LLaMA2-7B.
In each case, the ABM loss exhibits a clear and steady downward trend throughout
Stage~1, confirming that the activation boundary matching objective is being
effectively optimized.
Notably, CoLA shows a brief, transient upward spike around iteration 90, likely
due to sampling variance from CoLA's relatively small size, where
individual mini-batches may contain particularly ``hard'' examples whose activation
boundaries are harder to match.
However, the consistent reduction across all datasets demonstrates that our ABM
initialization successfully aligns the adapter's parameters with the pretrained
model's activation boundaries, enabling the adapter to learn informative gradients
quickly and setting the stage for faster convergence in later fine-tuning steps.

\section{Ablation Study on Margin and Layer Selection for LLaMA2-7B}
\label{app:llama_ablation}

To understand the impact of key hyperparameters in ABM initialization, we
conduct an ablation study on LLaMA2-7B, varying two critical factors: the
margin value $m$ and the number of layers where activation boundary matching
is applied.
Table~\ref{tab:llama_margin_layers} presents results for six configurations
combining two margin values ($m \in \{0.5, 1.0\}$) with three layer selection
strategies (12, 16, or 32 layers).

\begin{table*}[t]
\centering
\caption{\textbf{Ablation study on margin and layer selection for LLaMA2-7B.}
  We compare different combinations of margin values ($m$) and the number of
  layers where ABM initialization is applied. All rows use a fully trained
  reference and $S{=}100$ boundary-matching steps.
  LC = length-controlled win rate; WR = win rate.
  The best and second-best results are highlighted in \textbf{bold} and \underline{underline}, respectively.}
\label{tab:llama_margin_layers}
\small
\setlength{\tabcolsep}{6pt}
\begin{tabular*}{\textwidth}{@{\extracolsep{\fill}} cccccc}
\toprule
\textbf{Variant} & \textbf{Margin} & \textbf{\# Layers}
  & \textbf{MT-Bench}
  & \textbf{LC (\%)}
  & \textbf{WR (\%)} \\
\midrule
LoRA & -- & -- & 5.89 & 42.16 & 46.27 \\
\midrule
\multirow{6}{*}{ABM-LoRA}
  & 0.5 & 12  & 5.85 & 42.68 & 48.26 \\
  & 0.5 & 16  & \textbf{5.92} & \textbf{45.53} & \textbf{49.51} \\
  & 0.5 & 32  & \underline{5.91} & \underline{44.46} & \underline{49.50} \\
\cmidrule{2-6}
  & 1.0 & 12  & 5.84 & 43.44 & 48.82 \\
  & 1.0 & 16  & 5.88 & 43.06 & 49.32 \\
  & 1.0 & 32  & 5.85 & 42.40 & 47.27 \\
\bottomrule
\end{tabular*}
\end{table*}

\subsection{Experimental Setup}

All experiments use LLaMA2-7B~\citep{touvron2023llama} with the following
configuration.
\textbf{Base model:} \texttt{meta-llama/Llama-2-7b-hf}.
\textbf{LoRA:} rank $r=8$, alpha $\alpha=16$, dropout 0.1, task type
\texttt{CAUSAL\_LM}.
\textbf{Training:} learning rate $2\times10^{-5}$, 1 epoch, max sequence
length 1024, micro-batch size 1 with gradient accumulation 32, cosine decay
scheduler, warmup ratio 0.03, weight decay 0.0.
\textbf{ABM Stage 1:} fully trained reference; $S{=}100$ matching steps,
learning rate $2\times10^{-4}$, $m\in\{0.5,1.0\}$.
\textbf{Layer selection strategies:} (i)~12 layers: mid-level layers 12--23;
(ii)~16 layers: latter half, layers 16--31; (iii)~32 layers: all decoder layers.

\subsection{Results and Analysis}

Table~\ref{tab:llama_margin_layers} summarizes the impact of margin values and
layer selection strategies (rank~8, dropout~0.1).
Across all six configurations, ABM-LoRA consistently improves over the
LoRA baseline.
The best performance is achieved with $m=0.5$ and 16 layers, raising MT-Bench
from 5.89 to 5.92 (+0.03), boosting LC from 42.16\% to 45.53\% (+3.37 pp),
and increasing WR from 46.27\% to 49.51\% (+3.24 pp).

The smaller margin ($m=0.5$) leads on MT-Bench at every layer count, and on all
three metrics at 16 and 32 layers; the exception is 12 layers, where $m=1.0$
gives the higher LC (43.44\% vs.\ 42.68\%) and WR (48.82\% vs.\ 48.26\%).
Where the two orderings agree, the pattern suggests that tighter boundary
constraints better preserve structural knowledge encoded in the pretrained model.
Regarding layer selection, applying ABM to 16 layers (the deeper half) yields
the strongest results; both mid-level only (12 layers) and all 32 layers
show diminished gains.
Using only mid-level layers may miss critical structural information in the
deepest layers, whereas applying ABM to all 32 layers can over-constrain
the adapter, reducing its capacity for task-specific adaptation.
This finding confirms that selective layer matching focused on the latter half
provides the optimal balance between preserving pretrained knowledge and enabling
efficient task-specific fine-tuning.

\section{Ablation Study on VTAB-1K with ViT-B/16}
\label{app:vtab_ablation}

To systematically evaluate the impact of ABM initialization hyperparameters on
vision transformers, we conduct a comprehensive grid search over four factors on
VTAB-1K~\citep{zhai2019vtab} with ViT-B/16~\citep{dosovitskiy2021vit}: (1) layer
selection strategy (first 6, last 6, or all
12 attention layers), (2) margin $m\in\{0.5,1.0,2.0\}$, (3) layer weighting
(uniform vs.\ quadratic), and (4) the number of boundary-matching steps
$S\in\{500,1000\}$.
Here and in the remaining appendix sections the reference adapter is trained to
convergence, so $S$ is distinct from the probe length $K$ of
Table~\ref{tab:probe}.
This yields 36 configurations, each evaluated across all 19 VTAB-1K tasks.
Results are presented in Table~\ref{tab:vtab_ablation}.

\begin{table*}[t]
\centering
\caption{\textbf{VTAB-1K ablation study} with ViT-B/16 across 19 visual
  recognition tasks. Layer: F6=first\_6, L6=last\_6, A12=all\_12.
  $S$ is the number of boundary-matching steps; the reference is fully trained
  in every row. Weight: U=uniform, Q=quadratic.
  The best and second-best overall mean accuracies are highlighted in
  \textbf{bold} and \underline{underline}, respectively; where a value is tied,
  every tied row is marked.}
\label{tab:vtab_ablation}
\setlength{\tabcolsep}{3.5pt}
\renewcommand{\arraystretch}{0.9}
\resizebox{\textwidth}{!}{%
\begin{tabular}{cccc|ccccccc|cccc|cccccccc|c}
\toprule
& & & & \multicolumn{7}{c|}{Natural} & \multicolumn{4}{c|}{Specialized} & \multicolumn{8}{c|}{Structured} & \\
\cmidrule(lr){5-11} \cmidrule(lr){12-15} \cmidrule(lr){16-23}
\rotatebox{90}{Layer} & \rotatebox{90}{$S$} & \rotatebox{90}{M} & \rotatebox{90}{W}
& \rotatebox{90}{Cifar100} & \rotatebox{90}{Caltech101} & \rotatebox{90}{DTD}
& \rotatebox{90}{Flower102} & \rotatebox{90}{Pets} & \rotatebox{90}{SVHN}
& \rotatebox{90}{Sun397} & \rotatebox{90}{Camelyon} & \rotatebox{90}{EuroSAT}
& \rotatebox{90}{Resisc45} & \rotatebox{90}{Retinopathy}
& \rotatebox{90}{Clevr-Count} & \rotatebox{90}{Clevr-Dist} & \rotatebox{90}{DMLab}
& \rotatebox{90}{KITTI-Dst} & \rotatebox{90}{dSpr-Loc} & \rotatebox{90}{dSpr-Ori}
& \rotatebox{90}{sNORB-Azim} & \rotatebox{90}{sNORB-Ele} & \rotatebox{90}{Overall Mean} \\
\midrule
F6 & 500 & 0.5 & U & 71.0 & 92.2 & 62.3 & 98.2 & 88.3 & 84.5 & 50.3 & 84.6 & 95.4 & 78.9 & 71.7 & 81.2 & 61.9 & 43.9 & 77.4 & 85.6 & 49.8 & 28.4 & 42.6 & 70.9 \\
F6 & 500 & 0.5 & Q & 71.0 & 92.6 & 63.1 & 98.4 & 88.4 & 83.7 & 50.6 & 83.7 & 95.2 & 77.7 & 72.4 & 80.5 & 62.7 & 44.8 & 76.8 & 86.2 & 49.8 & 30.6 & 42.7 & 71.1 \\
F6 & 500 & 1.0 & U & 71.2 & 92.9 & 63.4 & 98.1 & 87.2 & 84.1 & 50.0 & 85.2 & 95.3 & 76.6 & 72.3 & 79.3 & 62.4 & 46.0 & 76.2 & 85.8 & 49.4 & 30.1 & 42.4 & 70.9 \\
F6 & 500 & 1.0 & Q & 71.7 & 93.4 & 62.3 & 98.0 & 88.1 & 84.5 & 50.5 & 85.0 & 95.1 & 78.4 & 71.7 & 80.4 & 62.8 & 45.8 & 77.4 & 85.6 & 50.9 & 30.2 & 42.9 & 71.3 \\
F6 & 500 & 2.0 & U & 70.2 & 92.2 & 62.4 & 98.4 & 88.4 & 83.6 & 47.9 & 83.9 & 95.6 & 77.4 & 71.5 & 79.9 & 62.0 & 46.9 & 74.1 & 85.9 & 50.5 & 31.4 & 38.0 & 70.5 \\
F6 & 500 & 2.0 & Q & 72.3 & 92.1 & 62.2 & 98.0 & 88.2 & 83.6 & 49.6 & 84.7 & 95.2 & 76.3 & 71.9 & 80.1 & 62.0 & 46.4 & 74.8 & 85.5 & 50.8 & 31.2 & 42.2 & 70.9 \\
F6 & 1000 & 0.5 & U & 69.9 & 92.3 & 63.2 & 98.1 & 87.0 & 85.0 & 49.5 & 84.6 & 95.3 & 77.2 & 71.8 & 80.3 & 62.4 & 45.2 & 76.8 & 84.5 & 49.0 & 29.8 & 42.8 & 70.8 \\
F6 & 1000 & 0.5 & Q & 70.9 & 92.1 & 63.1 & 98.3 & 86.9 & 84.3 & 49.5 & 85.1 & 95.5 & 77.1 & 71.8 & 80.2 & 62.4 & 45.7 & 75.7 & 85.3 & 48.0 & 28.0 & 42.7 & 70.7 \\
F6 & 1000 & 1.0 & U & 71.2 & 92.0 & 61.2 & 97.5 & 87.2 & 85.2 & 50.0 & 84.3 & 95.5 & 76.2 & 73.2 & 78.9 & 62.2 & 45.8 & 75.0 & 85.3 & 48.0 & 31.1 & 41.2 & 70.6 \\
F6 & 1000 & 1.0 & Q & 71.0 & 92.8 & 62.1 & 98.2 & 86.6 & 84.8 & 49.3 & 85.3 & 95.4 & 78.5 & 73.0 & 80.8 & 63.2 & 45.9 & 78.9 & 85.4 & 49.0 & 30.8 & 43.2 & 71.3 \\
F6 & 1000 & 2.0 & U & 71.5 & 91.3 & 63.1 & 97.3 & 87.7 & 84.7 & 48.8 & 84.7 & 95.6 & 77.6 & 72.0 & 79.4 & 62.1 & 48.1 & 74.7 & 84.9 & 50.3 & 33.5 & 40.3 & 70.9 \\
F6 & 1000 & 2.0 & Q & 71.3 & 93.0 & 63.8 & 97.8 & 87.9 & 84.6 & 48.5 & 83.8 & 95.6 & 76.9 & 71.5 & 79.3 & 62.9 & 47.5 & 77.1 & 85.4 & 50.2 & 31.6 & 41.6 & 71.1 \\
\midrule
\rowcolor{gray!20} L6 & 500 & 0.5 & U & 72.6 & 92.8 & 63.2 & 98.1 & 87.7 & 84.1 & 51.4 & 85.0 & 95.7 & 78.2 & 71.5 & 81.5 & 63.0 & 47.2 & 77.5 & 86.0 & 52.1 & 31.2 & 45.2 & \textbf{71.8} \\
L6 & 500 & 0.5 & Q & 74.2 & 92.8 & 65.6 & 97.4 & 87.6 & 84.8 & 51.4 & 85.5 & 95.2 & 77.8 & 72.3 & 81.5 & 63.6 & 47.6 & 74.5 & 83.4 & 51.8 & 29.1 & 44.4 & 71.6 \\
L6 & 500 & 1.0 & U & 72.8 & 93.0 & 63.0 & 98.1 & 88.0 & 85.4 & 50.5 & 85.5 & 95.0 & 78.7 & 71.7 & 81.4 & 62.9 & 47.4 & 75.5 & 86.0 & 51.8 & 29.8 & 43.6 & 71.6 \\
L6 & 500 & 1.0 & Q & 73.1 & 93.0 & 64.7 & 98.0 & 87.3 & 85.3 & 50.4 & 84.3 & 95.5 & 78.5 & 70.9 & 81.1 & 64.7 & 47.3 & 76.5 & 84.5 & 49.8 & 29.8 & 44.1 & 71.5 \\
L6 & 500 & 2.0 & U & 72.1 & 92.6 & 64.2 & 97.7 & 87.6 & 85.5 & 49.6 & 85.0 & 95.4 & 77.9 & 72.4 & 80.4 & 63.0 & 46.4 & 76.1 & 86.5 & 51.1 & 29.9 & 43.9 & 71.4 \\
\rowcolor{gray!10} L6 & 500 & 2.0 & Q & 73.8 & 93.0 & 64.2 & 97.7 & 87.7 & 84.8 & 49.7 & 85.0 & 95.1 & 78.8 & 72.0 & 81.6 & 64.0 & 47.4 & 75.5 & 85.1 & 50.6 & 31.9 & 44.7 & \underline{71.7} \\
\rowcolor{gray!10} L6 & 1000 & 0.5 & U & 72.2 & 93.3 & 62.7 & 98.6 & 87.8 & 84.8 & 49.1 & 87.0 & 95.5 & 77.9 & 72.2 & 81.4 & 62.8 & 46.8 & 77.5 & 86.4 & 51.9 & 31.4 & 43.5 & \underline{71.7} \\
L6 & 1000 & 0.5 & Q & 72.7 & 92.8 & 63.4 & 98.3 & 87.5 & 85.3 & 50.5 & 85.2 & 95.6 & 76.8 & 72.6 & 82.1 & 63.8 & 47.2 & 73.7 & 85.7 & 51.4 & 31.9 & 44.8 & 71.6 \\
L6 & 1000 & 1.0 & U & 71.8 & 92.4 & 63.6 & 97.8 & 87.2 & 84.6 & 49.4 & 85.1 & 95.2 & 77.9 & 72.5 & 81.2 & 62.8 & 47.4 & 74.4 & 86.5 & 50.6 & 29.8 & 42.6 & 71.2 \\
L6 & 1000 & 1.0 & Q & 71.8 & 93.1 & 65.0 & 97.8 & 88.1 & 84.5 & 49.3 & 83.9 & 95.5 & 77.8 & 72.5 & 81.7 & 63.5 & 46.5 & 77.2 & 85.5 & 51.3 & 30.4 & 44.3 & 71.6 \\
L6 & 1000 & 2.0 & U & 70.5 & 92.3 & 62.9 & 97.7 & 85.6 & 85.6 & 49.4 & 85.2 & 95.6 & 77.6 & 72.3 & 81.6 & 62.9 & 47.0 & 72.3 & 86.1 & 50.3 & 29.8 & 43.2 & 70.9 \\
L6 & 1000 & 2.0 & Q & 72.4 & 92.7 & 63.4 & 97.2 & 87.4 & 84.6 & 48.7 & 85.7 & 95.5 & 78.4 & 72.7 & 81.4 & 63.3 & 46.8 & 77.8 & 85.9 & 51.1 & 30.3 & 44.0 & 71.5 \\
\midrule
A12 & 500 & 0.5 & U & 71.9 & 93.8 & 62.9 & 97.8 & 87.2 & 83.1 & 50.3 & 85.7 & 95.9 & 78.4 & 72.0 & 81.2 & 63.0 & 45.7 & 75.2 & 85.7 & 50.5 & 28.5 & 44.4 & 71.2 \\
A12 & 500 & 0.5 & Q & 72.2 & 93.0 & 62.9 & 98.5 & 87.8 & 85.1 & 51.0 & 85.7 & 95.4 & 78.0 & 72.1 & 81.8 & 62.4 & 46.5 & 76.1 & 85.8 & 50.8 & 29.6 & 44.5 & 71.5 \\
A12 & 500 & 1.0 & U & 72.6 & 91.7 & 63.3 & 98.0 & 86.9 & 86.0 & 50.5 & 85.2 & 95.2 & 77.5 & 72.9 & 81.7 & 62.6 & 47.1 & 76.1 & 86.5 & 51.4 & 29.9 & 43.3 & 71.5 \\
A12 & 500 & 1.0 & Q & 72.9 & 91.9 & 62.9 & 98.0 & 86.6 & 84.6 & 50.3 & 84.2 & 95.6 & 78.5 & 72.0 & 82.0 & 63.2 & 46.4 & 77.5 & 85.1 & 50.1 & 30.6 & 43.8 & 71.4 \\
A12 & 500 & 2.0 & U & 72.9 & 92.0 & 62.9 & 97.8 & 87.5 & 85.5 & 50.1 & 85.2 & 95.8 & 77.4 & 72.0 & 80.1 & 62.1 & 47.2 & 76.8 & 85.4 & 52.0 & 31.2 & 44.5 & 71.5 \\
A12 & 500 & 2.0 & Q & 73.1 & 93.0 & 63.4 & 98.1 & 87.8 & 85.2 & 49.1 & 85.3 & 95.2 & 78.3 & 72.3 & 81.5 & 62.6 & 45.7 & 74.5 & 85.8 & 51.2 & 30.1 & 43.8 & 71.4 \\
A12 & 1000 & 0.5 & U & 71.5 & 92.8 & 63.0 & 98.2 & 86.8 & 84.4 & 49.2 & 85.3 & 95.4 & 77.0 & 71.9 & 81.2 & 63.6 & 46.9 & 75.2 & 85.7 & 50.8 & 30.7 & 45.2 & 71.3 \\
\rowcolor{gray!10} A12 & 1000 & 0.5 & Q & 72.7 & 93.6 & 63.3 & 98.2 & 87.5 & 84.3 & 49.2 & 85.5 & 95.0 & 78.4 & 72.8 & 81.2 & 63.3 & 46.8 & 78.1 & 86.4 & 52.0 & 30.9 & 43.2 & \underline{71.7} \\
A12 & 1000 & 1.0 & U & 71.3 & 92.7 & 61.8 & 97.8 & 86.6 & 85.0 & 49.4 & 84.8 & 95.8 & 77.2 & 72.0 & 80.5 & 63.3 & 46.6 & 75.8 & 86.5 & 50.6 & 30.1 & 44.2 & 71.2 \\
A12 & 1000 & 1.0 & Q & 71.6 & 93.7 & 62.3 & 97.8 & 87.6 & 85.7 & 50.1 & 85.4 & 95.4 & 78.8 & 72.2 & 81.0 & 63.1 & 46.6 & 77.1 & 86.6 & 50.6 & 31.1 & 44.3 & 71.6 \\
A12 & 1000 & 2.0 & U & 71.8 & 92.7 & 62.8 & 97.2 & 87.6 & 85.4 & 49.8 & 85.0 & 95.4 & 78.2 & 72.1 & 80.6 & 62.5 & 46.8 & 77.8 & 85.4 & 49.9 & 32.3 & 43.1 & 71.4 \\
A12 & 1000 & 2.0 & Q & 71.5 & 93.2 & 63.4 & 97.4 & 87.5 & 85.8 & 48.8 & 85.8 & 95.6 & 78.7 & 71.5 & 81.3 & 63.1 & 45.9 & 77.2 & 85.8 & 50.4 & 31.2 & 43.8 & 71.5 \\
\bottomrule
\end{tabular}
}
\end{table*}

\subsection{Experimental Setup}

All experiments use ViT-B/16 pretrained on ImageNet-21k~\citep{deng2009imagenet}.
\textbf{LoRA:} rank $r=8$, alpha $\alpha=16$, applied to query and value
projections in attention layers.
\textbf{Training:} learning rate varies by dataset (see main paper), batch
size 64, cosine decay scheduler.
\textbf{ABM Stage 1:} fully trained reference; matching steps $S\in\{500,1000\}$,
learning rate $5\times10^{-4}$, $m\in\{0.5,1.0,2.0\}$.
\textbf{Layer selection:} F6 (first 6, layers 0--5), L6 (last 6, layers 6--11),
A12 (all 12).
\textbf{Layer weighting:} uniform (U) assigns equal weight; quadratic (Q) uses
$w_l = \bigl(\frac{l+1}{L}\bigr)^2$ to emphasize deeper layers.

\subsection{Layer Selection Strategy}

Examining the overall mean accuracy across all 19 VTAB-1K tasks, L6 consistently
achieves the highest performance, with the best configuration reaching 71.8\%.
L6 accounts for 7 of the top 10 results; the per-strategy ranges are
L6 70.9--71.8\%, A12 71.2--71.7\%, and F6 70.5--71.3\%.

This finding aligns with our observations on LLaMA2-7B, where applying ABM to
the latter half of layers yielded optimal results.
The superiority of L6 suggests that deeper layers, which capture more abstract
and task-specific representations, benefit most from activation boundary
alignment.
Applying ABM to all 12 layers (A12) shows comparable but slightly lower
performance, likely due to over-constraining early layers.
Meanwhile, F6 misses critical structural information encoded in deeper layers,
resulting in consistently lower accuracy.

\section{Training Loss Evolution on VTAB-1K}
\label{app:vtab_loss}
\begin{figure*}[t]
  \centering
  \begin{subfigure}[b]{0.48\linewidth}
    \includegraphics[width=\linewidth]{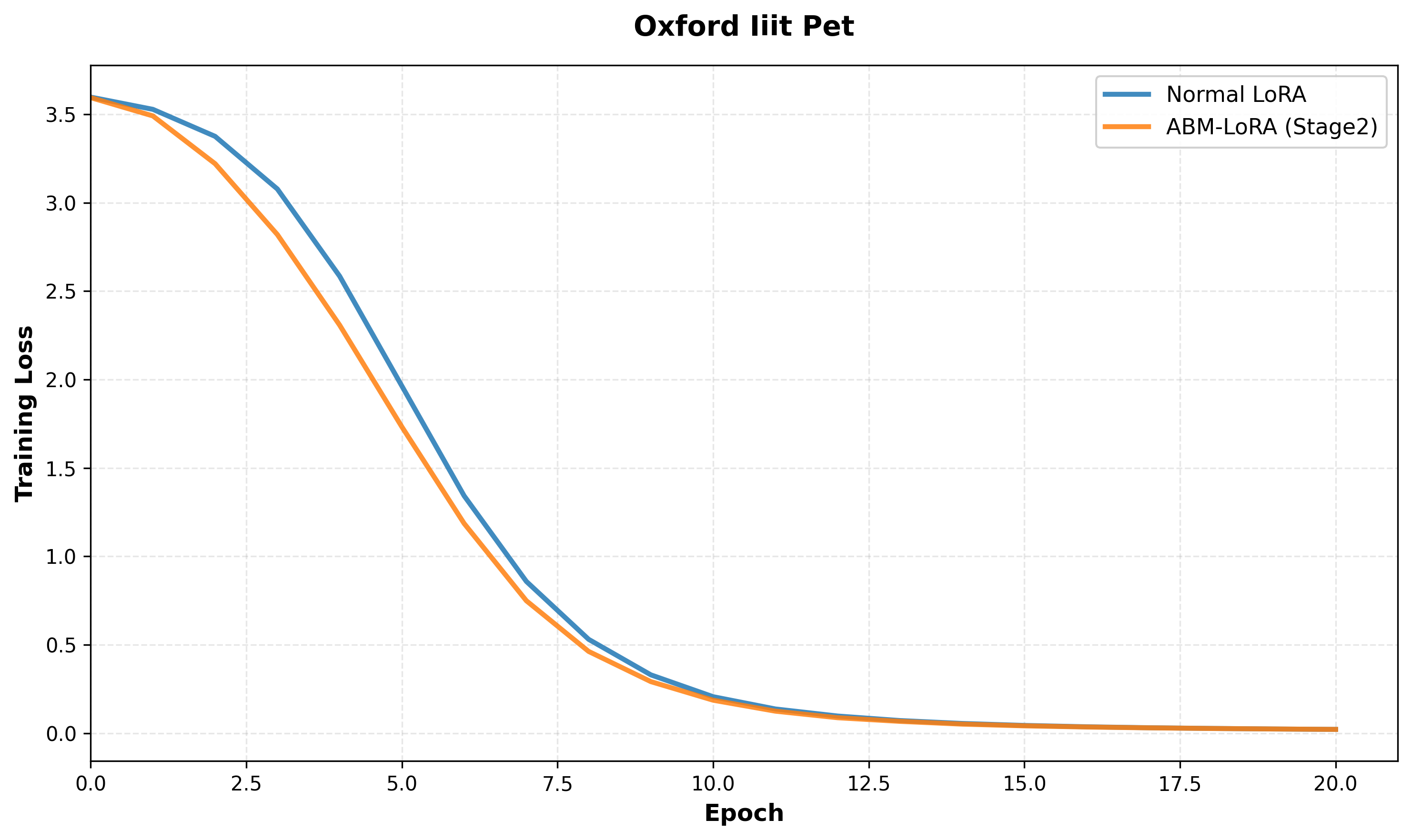}
    \caption{Oxford-IIIT Pet}
    \label{fig:loss_oxford}
  \end{subfigure}
  \hfill
  \begin{subfigure}[b]{0.48\linewidth}
    \includegraphics[width=\linewidth]{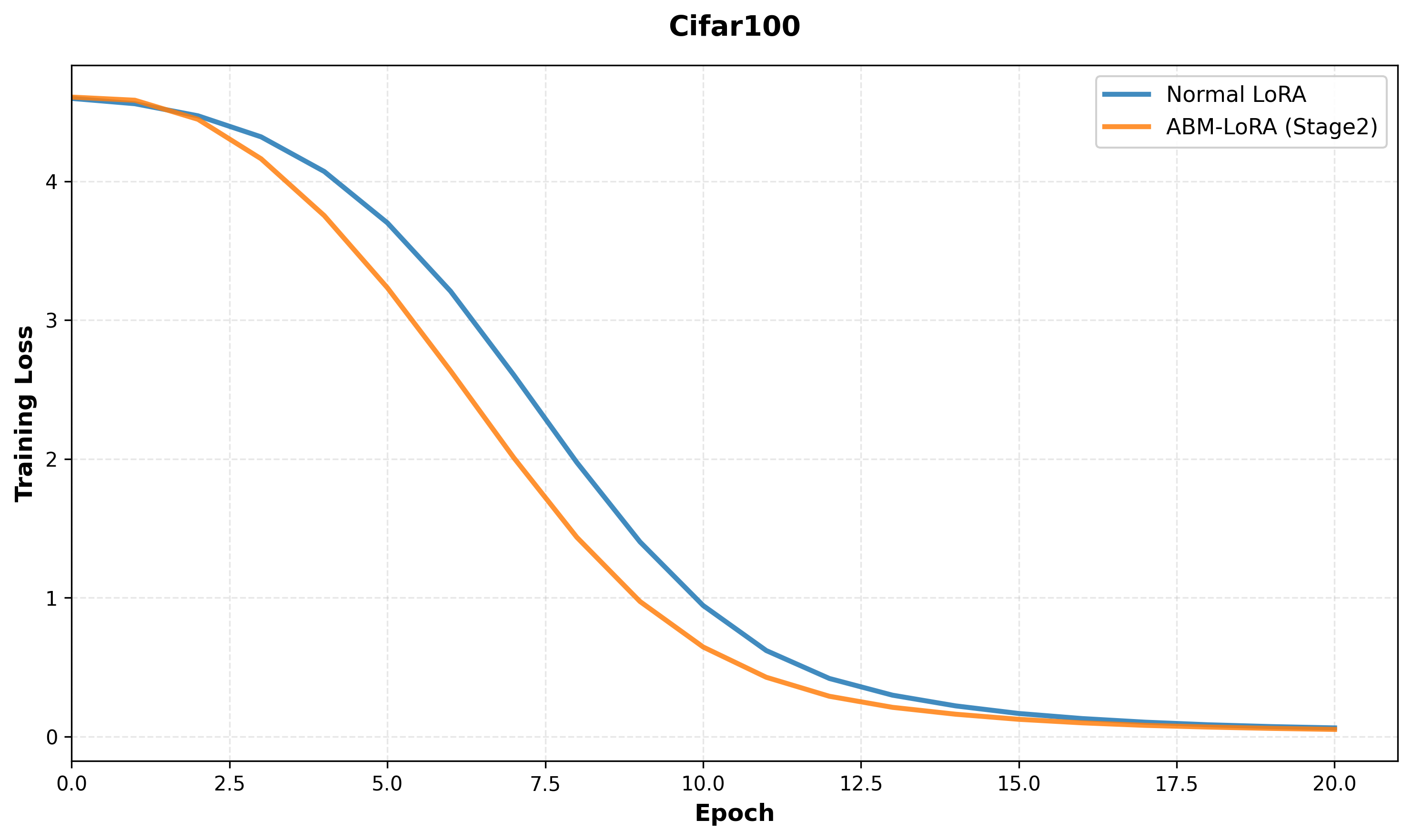}
    \caption{CIFAR-100}
    \label{fig:loss_cifar}
  \end{subfigure}

  \vspace{1ex}

  \begin{subfigure}[b]{0.48\linewidth}
    \includegraphics[width=\linewidth]{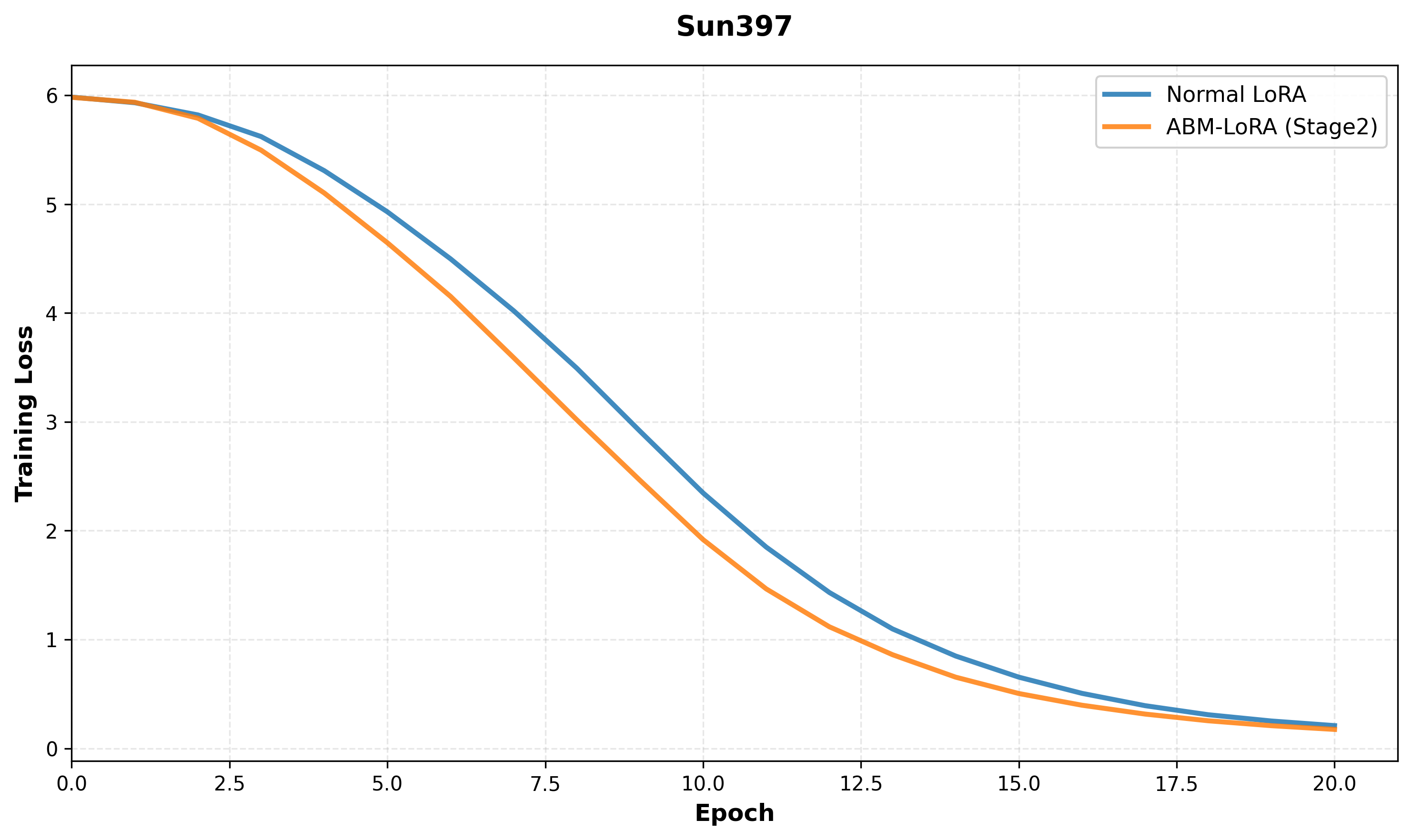}
    \caption{SUN397}
    \label{fig:loss_sun}
  \end{subfigure}
  \hfill
  \begin{subfigure}[b]{0.48\linewidth}
    \includegraphics[width=\linewidth]{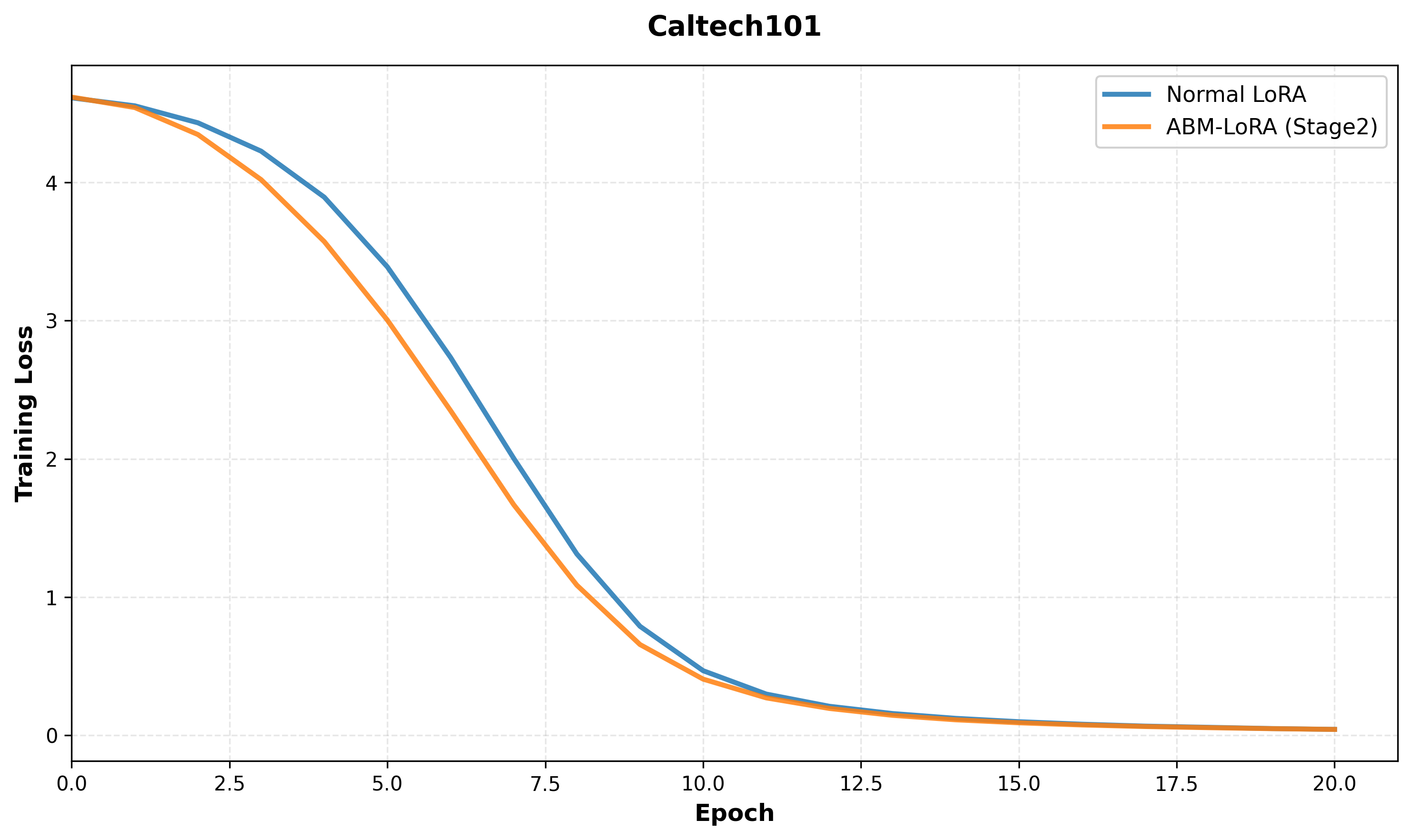}
    \caption{Caltech101}
    \label{fig:loss_caltech}
  \end{subfigure}
  \caption{\textbf{Training loss curves on VTAB-1K tasks.}
    Four representative examples from the 19-task benchmark illustrate
    ABM-LoRA's advantages in achieving lower initial training loss and
    faster early-stage convergence.}
  \label{fig:vit_natural_loss}
\end{figure*}

To demonstrate that ABM initialization extends to vision transformers, we
evaluate ViT-B/16 with rank-8 adapters across all 19 tasks in the VTAB-1K
benchmark.
Figure~\ref{fig:vit_natural_loss} presents training curves for four
representative tasks.
Across these examples, ABM-LoRA consistently demonstrates steeper loss reduction
in the early training phase.
On CIFAR-100, SUN397, and Caltech101, the improvement is particularly visible
within the first 5 epochs, where ABM-LoRA achieves substantially lower training
loss than LoRA.
Even on Oxford-IIIT Pet, where convergence patterns are more similar, ABM-LoRA
maintains a consistent advantage throughout training.
These curves illustrate how activation boundary matching enables vision adapters
to start from a better-informed initialization, leading to more efficient
optimization from the very beginning of fine-tuning.

\begin{table*}[t]
\centering
\caption{\textbf{Margin $m$ and matching length $S$ on ConvNeXt-T.}
  $S$ is the number of boundary-matching steps; the reference adapter is fully
  trained in every row.
  Sign is the fraction of pre-activation signs the student reproduces at the
  end of Stage~1; Acc is Stage~2 top-1 test accuracy. All runs use $r{=}8$,
  $\alpha{=}16$, seed 42, and matching on the last two stages; Stage~1 uses
  lr $3{\times}10^{-4}$ with batch 32, Stage~2 lr $10^{-3}$ with batch 64 for
  20 epochs. Shaded rows are $S{=}1000$.}
\label{tab:convnext_margin_k}
\small
\setlength{\tabcolsep}{9.5pt}
\begin{tabular}{cc cc cc cc cc c}
\toprule
& & \multicolumn{2}{c}{Aircraft} & \multicolumn{2}{c}{Flowers102}
  & \multicolumn{2}{c}{Food-101} & \multicolumn{2}{c}{Pets} & \\
\cmidrule(lr){3-4} \cmidrule(lr){5-6} \cmidrule(lr){7-8} \cmidrule(lr){9-10}
$m$ & $S$ & Sign & Acc & Sign & Acc & Sign & Acc & Sign & Acc & Mean \\
\midrule
\multirow{3}{*}{0.5}
  & 100  & 96.0 & 74.32 & 97.6 & 95.25 & 95.4 & 85.92 & 97.9 & \textbf{93.57} & 87.27 \\
  & 500  & 98.4 & 75.88 & 99.3 & 95.54 & 97.1 & 86.27 & 99.3 & 93.32 & 87.75 \\
\rowcolor{evagreen!10}
  & 1000 & 99.2 & 76.78 & 99.5 & \textbf{95.66} & 98.5 & 86.34 & 99.5 & 93.24 & 88.01 \\
\cmidrule{2-11}
\multirow{3}{*}{1.0}
  & 100  & 96.3 & 73.99 & 97.7 & 95.20 & 95.7 & 85.96 & 98.0 & 93.08 & 87.06 \\
  & 500  & 98.5 & 75.82 & 98.8 & 95.61 & 97.5 & 86.30 & 98.8 & 93.32 & 87.76 \\
\rowcolor{evagreen!10}
  & 1000 & 98.9 & \textbf{76.84} & 98.8 & 95.56 & 98.6 & \textbf{86.66} & 98.8 & 93.21 & \textbf{88.07} \\
\bottomrule
\end{tabular}
\end{table*}

\begin{table*}[t]
\centering
\caption{\textbf{Margin $m$ and matching length $S$ on Swin-T.}
  Columns are as in Table~\ref{tab:convnext_margin_k}, and all training
  settings are identical: $r{=}8$, $\alpha{=}16$, seed 42, matching on the last
  two stages, Stage~1 lr $3{\times}10^{-4}$ with batch 32, Stage~2 lr
  $10^{-3}$ with batch 64 for 20 epochs. Shaded rows are $S{=}1000$.
  The two best Mean values differ only past the second decimal
  ($85.375$ vs.\ $85.380$) and are both marked.}
\label{tab:swin_margin_k}
\small
\setlength{\tabcolsep}{9.5pt}
\begin{tabular}{cc cc cc cc cc c}
\toprule
& & \multicolumn{2}{c}{Aircraft} & \multicolumn{2}{c}{Flowers102}
  & \multicolumn{2}{c}{Food-101} & \multicolumn{2}{c}{Pets} & \\
\cmidrule(lr){3-4} \cmidrule(lr){5-6} \cmidrule(lr){7-8} \cmidrule(lr){9-10}
$m$ & $S$ & Sign & Acc & Sign & Acc & Sign & Acc & Sign & Acc & Mean \\
\midrule
\multirow{3}{*}{0.5}
  & 100  & 94.4 & 67.21 & 96.5 & 93.80 & 94.2 & 85.60 & 97.0 & 94.03 & 85.16 \\
  & 500  & 96.3 & \textbf{67.48} & 98.0 & 94.16 & 94.9 & 85.83 & 98.1 & 94.03 & \textbf{85.38} \\
\rowcolor{evagreen!10}
  & 1000 & 97.3 & 67.15 & 98.3 & \textbf{94.21} & 95.6 & 85.81 & 98.4 & 93.95 & 85.28 \\
\cmidrule{2-11}
\multirow{3}{*}{1.0}
  & 100  & 95.0 & 67.12 & 96.9 & 94.11 & 94.6 & 85.76 & 97.3 & 93.95 & 85.24 \\
  & 500  & 96.9 & 66.82 & 98.0 & 93.82 & 95.9 & 85.90 & 98.1 & 93.89 & 85.11 \\
\rowcolor{evagreen!10}
  & 1000 & 97.6 & 67.36 & 98.1 & 94.00 & 97.1 & \textbf{85.97} & 98.2 & \textbf{94.19} & \textbf{85.38} \\
\bottomrule
\end{tabular}
\end{table*}

\subsection{Margin Value Analysis}

Across all layer selection strategies, $m=0.5$ consistently outperforms larger
margins.
For L6 configurations with $S{=}500$, the overall mean is 71.6--71.8\%
($m=0.5$), 71.5--71.6\% ($m=1.0$), and 71.4--71.7\% ($m=2.0$).
The tighter margin enforces stricter activation boundary alignment, better
preserving the pretrained model's learned feature partitioning.
This trend holds across F6 and A12 as well, confirming that $m=0.5$ is the
most reliable choice for vision transformers.

\subsection{Impact of Matching Steps and Layer Weighting}

Comparing $S{=}500$ vs.\ $S{=}1000$, we observe only marginal differences; for the
best L6 configurations with $m=0.5$, $S{=}500$ yields 71.6--71.8\% and $S{=}1000$
yields 71.6--71.7\%.
This suggests that 500 matching steps are sufficient for effective activation
boundary alignment, and doubling the matching steps provides minimal additional
benefit while increasing computational cost.

Regarding layer weighting, both uniform and quadratic schemes achieve competitive
results with no clear overall winner.
For L6, $m=0.5$: $S{=}500$ uniform 71.8\%, quadratic 71.6\%; $S{=}1000$ uniform
71.7\%, quadratic 71.6\%.
Since ABM is already restricted to the last 6 layers, the uniform scheme performs
equally well without needing to re-weight contributions across layers.

\subsection{Task Group Analysis}

\noindent\textbf{Natural tasks.}
All configurations achieve competitive performance (77.7--79.1\% mean), as these
tasks align closely with ImageNet pretraining and require minimal adaptation.
Hyperparameter choice has limited impact here.

\noindent\textbf{Specialized tasks.}
L6 configurations maintain strong performance (82.3--83.1\%), with one
configuration achieving 87.0\% on Camelyon.
Medical and satellite imaging tasks benefit moderately from proper initialization.

\noindent\textbf{Structured tasks.}
This is where ABM initialization shows the most substantial gains.
The best L6 configurations reach a structured mean of 59.5--60.5\%, compared to
58.7\% for LoRA (main paper Table~2).
Notable improvements among top-performing L6 configurations include
sNORB-Ele (43.5--45.2\% vs.\ 39.2\%), dSpr-Ori (50.6--52.1\% vs.\ 49.5\%),
and Clevr-Count (81.4--82.1\% vs.\ 79.9\%).
These structured reasoning tasks—involving 3D viewpoint prediction, spatial
localization, and object counting—benefit significantly from the pretrained
model's preserved feature space partitioning.

\subsection{Selection of Configurations for Main Paper}

Based on this grid search, we selected four L6, $m=0.5$ configurations for
the main paper (Table~2):
(a)~$S{=}500$, uniform: 71.8\%;
(b)~$S{=}500$, quadratic: 71.6\%;
(c)~$S{=}1000$, uniform: 71.7\%;
(d)~$S{=}1000$, quadratic: 71.6\%.
These represent the optimal hyperparameter region, demonstrating that focusing
ABM initialization on the last 6 layers with $m=0.5$ consistently yields the
best vision transformer adaptation performance.


\section{Margin and Matching Length on ConvNeXt-T}
\label{app:convnext_margin_k}

The VTAB-1K study in Appendix~\ref{app:vtab_ablation} varies the number of
matching steps only between 500 and 1000. Here we widen that range and pair it
with the margin on a convolutional backbone, ConvNeXt-T,
using four fine-grained recognition datasets---FGVC-Aircraft
\citep{maji2013aircraft}, Flowers102 \citep{nilsback2008flowers}, Food-101
\citep{bossard2014food101} and Oxford-IIIT Pet \citep{parkhi2012pets}---and
reporting both the Stage~1 diagnostic---the fraction
of pre-activation signs the student reproduces---and the resulting Stage~2
accuracy. Writing $S$ for the number of boundary-matching steps,
Table~\ref{tab:convnext_margin_k} covers
$m \in \{0.5, 1.0\}$ and $S \in \{100, 500, 1000\}$.

\paragraph{Stage 1 optimizes what it is meant to.}
Sign agreement increases monotonically with $S$ in all twelve runs, as it must
if the objective is being optimized: on FGVC-Aircraft at $m{=}0.5$ it goes from
$96.0\%$ to $99.2\%$, and the Stage~1 loss falls by roughly a factor of six over
the same range (not shown). This is the image-domain counterpart of the loss
curves in Appendix~\ref{app:stage1_loss}. A larger margin produces a larger loss
at equal $S$, as the definition of the objective implies, but reaches comparable
agreement. The matching procedure therefore behaves as designed, independently
of what that agreement is worth downstream.

\paragraph{Most of the benefit arrives early.}
The step from $S{=}100$ to $S{=}500$ is worth $1.56$ points on FGVC-Aircraft at
$m{=}0.5$, while the step from $500$ to $1000$ adds a further $0.90$; on the
other three datasets the second step is worth at most $0.12$. This is
consistent with the VTAB-1K observation that 500 and 1000 matching steps
perform comparably (Appendix~\ref{app:vtab_ablation}). A short matching schedule
is sufficient in most cases; the exception is the task on which the adapter is
furthest from fitting its training data. Table~\ref{tab:probe} reports a
separate saturation, in the probe length $K$ on GLUE; the two are distinct
quantities and we do not claim that one predicts the other.

\begin{table*}[t]
\centering
\caption{\textbf{Performance across adapter ranks on T5-base GLUE benchmark.}
  We compare LoRA and ABM-LoRA at ranks $r \in \{8, 16, 32\}$
  (alpha $= 2r$) across five GLUE tasks.
  The best and second-best results are highlighted in \textbf{bold} and
  \underline{underline}, respectively; where the best value is tied, both rows
  are set in bold and no second-best is marked.}
\label{tab:rank_glue}
\small
\setlength{\tabcolsep}{18pt}
\begin{tabular}{ll ccccc}
\toprule
\textbf{Method} & \textbf{Rank} & \textbf{MNLI} & \textbf{SST-2}
  & \textbf{CoLA} & \textbf{QNLI} & \textbf{MRPC} \\
\midrule
\multirow{3}{*}{LoRA}
  & $r=8$  & 83.57 & 91.86 & 69.80 & 92.77 & 75.00 \\
  & $r=16$ & 84.27 & 92.09 & 72.48 & 92.88 & 75.74 \\
  & $r=32$ & \underline{84.86} & \underline{93.81} & 75.84 & \textbf{93.03} & 82.11 \\
\midrule
\multirow{3}{*}{ABM-LoRA}
  & $r=8$  & 84.35 & 92.43 & 76.51 & 92.90 & 83.09 \\
  & $r=16$ & 84.60 & 93.23 & \underline{79.19} & \textbf{93.03} & \underline{84.07} \\
  & $r=32$ & \textbf{85.10} & \textbf{94.27} & \textbf{80.25} & 92.95 & \textbf{85.29} \\
\bottomrule
\end{tabular}
\end{table*}

\begin{figure*}[t]
  \centering
  \begin{subfigure}[t]{0.24\textwidth}
    \centering
    \includegraphics[width=\linewidth]{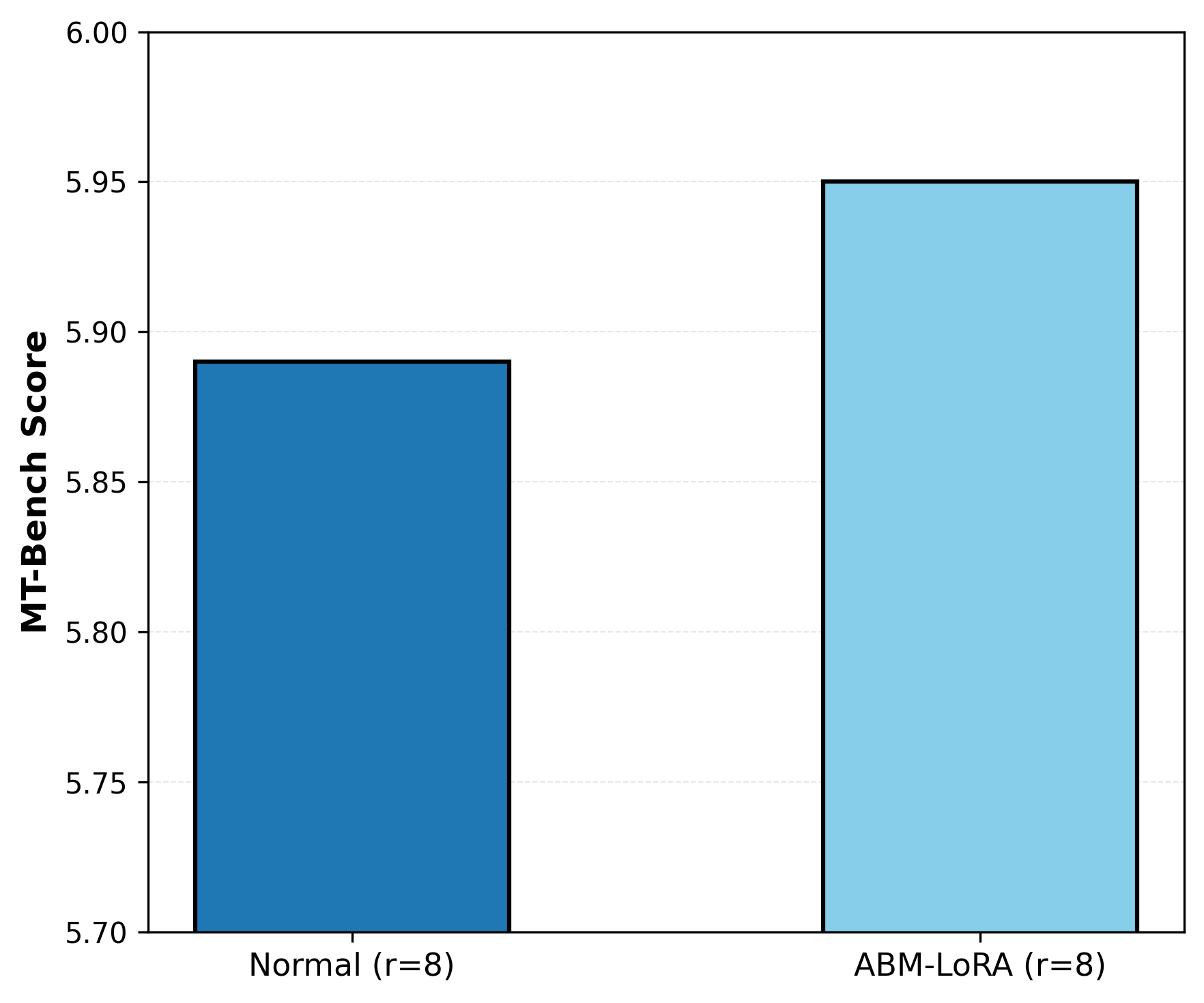}
    \caption{$r=8$: MT-Bench.}
    \label{fig:rank8_mtbench}
  \end{subfigure}%
  \hfill
  \begin{subfigure}[t]{0.24\textwidth}
    \centering
    \includegraphics[width=\linewidth]{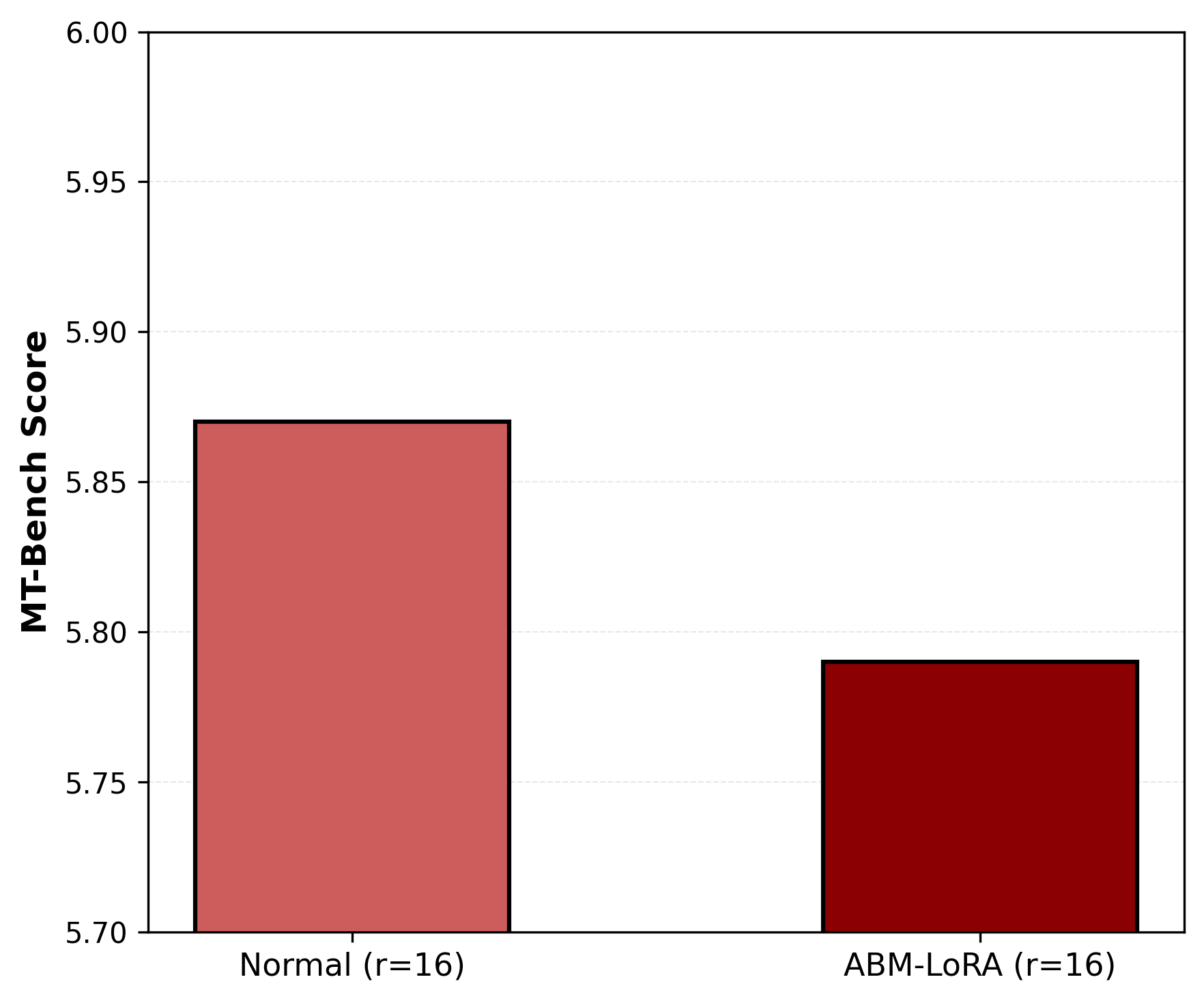}
    \caption{$r=16$: MT-Bench.}
    \label{fig:rank16_mtbench}
  \end{subfigure}%
  \hfill
  \begin{subfigure}[t]{0.24\textwidth}
    \centering
    \includegraphics[width=\linewidth]{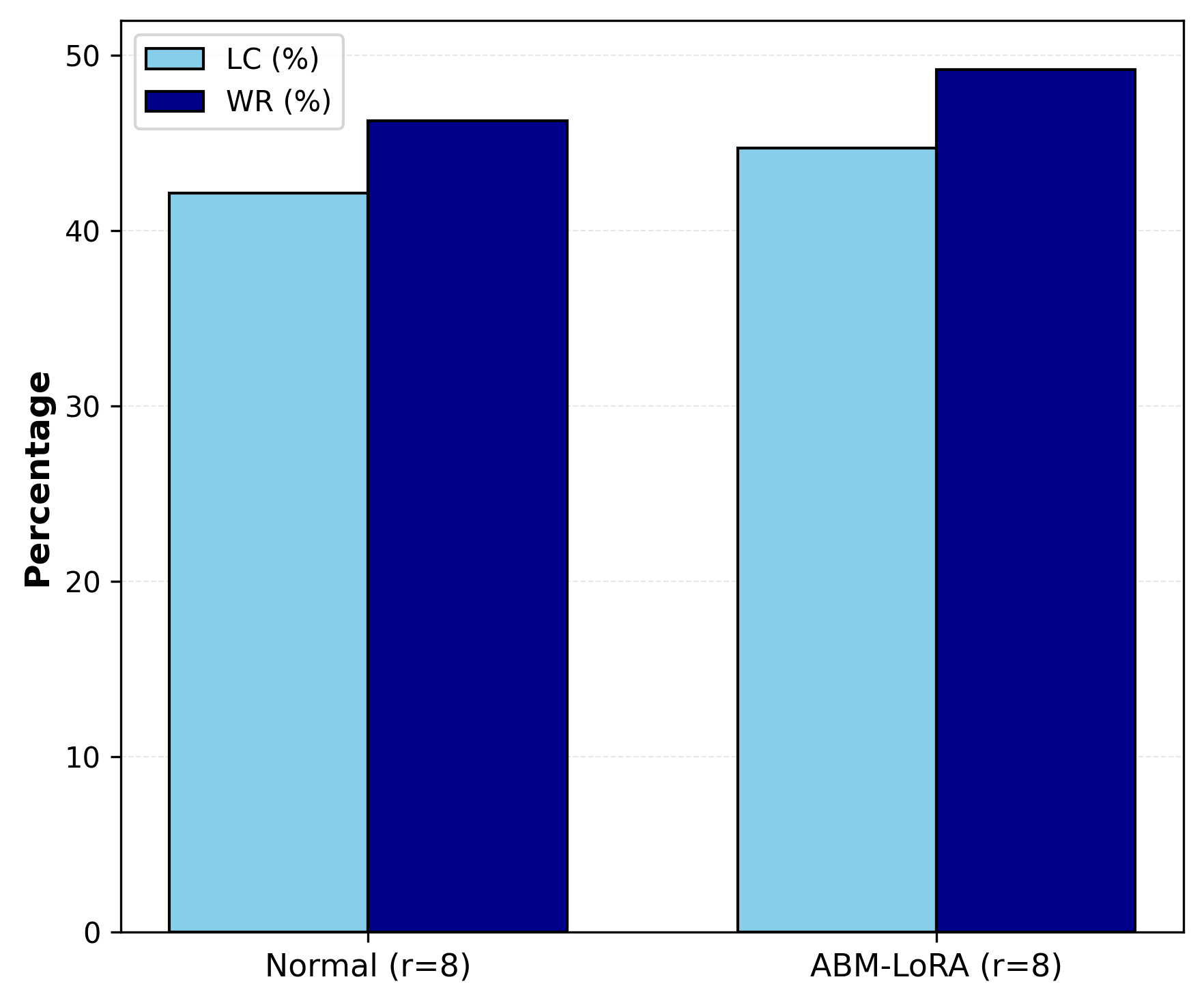}
    \caption{$r=8$: LC \& WR.}
    \label{fig:rank8_lc_wr}
  \end{subfigure}%
  \hfill
  \begin{subfigure}[t]{0.24\textwidth}
    \centering
    \includegraphics[width=\linewidth]{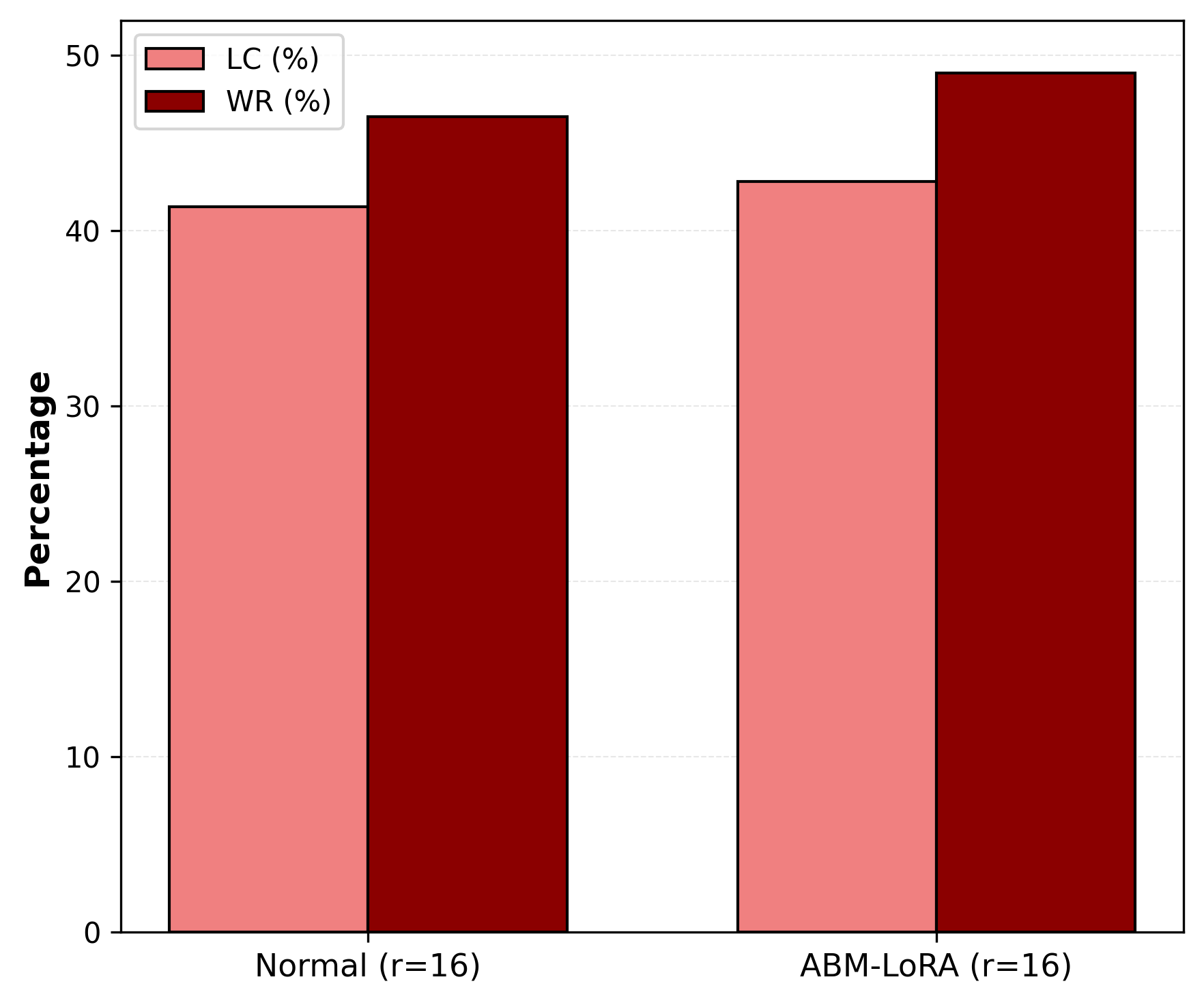}
    \caption{$r=16$: LC \& WR.}
    \label{fig:rank16_lc_wr}
  \end{subfigure}
  \caption{\textbf{Ablation study by adapter rank ($r=8$ vs.\ $r=16$)} on
    dialogue performance with dropout fixed at 0.1.
    (a,b)~MT-Bench scores with zoomed $y$-axis to highlight small differences.
    (c,d)~Length-controlled win rate (LC) and win rate (WR) percentages.}
  \label{fig:rank_ablation_full}
\end{figure*}

\paragraph{Where extra agreement stops paying.}
FGVC-Aircraft, the hardest of the four datasets, gains $2.46$ points across the
full range of $S$ at $m{=}0.5$ (and $2.85$ at $m{=}1.0$). The remaining three
vary by at most $0.74$ points, and on Pets the trend is slightly negative
($93.57 \rightarrow 93.24$) even as sign agreement improves from $97.9\%$ to
$99.5\%$. The datasets that stop responding
are those the adapter can already separate, which mirrors the diminishing
returns we observe on QNLI in Appendix~\ref{app:rank_glue}: once the task is
close to its ceiling under the given setup, initialization quality has less
room to act.

\paragraph{The margin is a second-order effect.}
At matched $S$ the two margins differ by at most $0.49$ points on any dataset,
and neither ordering is stable---$m{=}0.5$ leads on Flowers102 and Pets at
$S{=}1000$ and trails on FGVC-Aircraft and Food-101. We additionally ran
$m{=}1.5$ on FGVC-Aircraft and Food-101 (not shown), where it gives the best
$S{=}100$ accuracy on the former ($74.59$ vs.\ $74.32$ and $73.99$) and the best
$S{=}500$ accuracy on the latter ($86.43$ vs.\ $86.27$ and $86.30$). The
differences are the same size as those separating margins on VTAB-1K, where the
overall means span $71.4$--$71.8\%$ across all three margin values. These are
single-seed runs and we do not read an ordering into differences of a few tenths
of a point.

\section{Margin and Matching Length on Swin-T}
\label{app:swin_margin_k}

We repeat the sweep of Appendix~\ref{app:convnext_margin_k} on a hierarchical
transformer, Swin-T, matching on the pre-GELU activations of
the MLP block rather than
on the ConvNeXt inverted bottleneck. Table~\ref{tab:swin_margin_k} reports the
same twelve configurations on the same four datasets.

\paragraph{Boundary matching is not backbone-specific.}
As on ConvNeXt-T, sign agreement rises monotonically with $S$ in all twelve
runs---$94.4 \rightarrow 97.3\%$ on FGVC-Aircraft and $94.2 \rightarrow 95.6\%$
on Food-101 at $m{=}0.5$---with the Stage~1 loss falling by a factor of three to
four (not shown). Together with the ViT-B/16 results of
Appendix~\ref{app:vtab_ablation}, the encoder-decoder results of
Appendix~\ref{app:stage1_loss} and the decoder-only results of
Appendix~\ref{app:llama_ablation}, this covers convolutional, hierarchical
transformer, plain transformer, encoder-decoder and decoder architectures: the
objective optimizes in all five.

\paragraph{Downstream accuracy is flat on this backbone.}
Unlike ConvNeXt-T, no dataset here responds strongly to either hyperparameter.
The full spread across all twelve configurations is $0.66$ points on
FGVC-Aircraft, $0.41$ on Flowers102, $0.37$ on Food-101 and $0.30$ on Pets, and
neither $m$ nor $S$ orders consistently: $m{=}0.5$ gives the best
FGVC-Aircraft result at $S{=}500$ and trails $m{=}1.0$ at $S{=}1000$
($67.15$ vs.\ $67.36$). The two backbones
agree on three of four datasets; the FGVC-Aircraft sensitivity seen on
ConvNeXt-T does not reproduce here, which we report as a limitation of the
generality of that observation rather than as evidence against it.

\begin{figure*}[t]
  \centering
  \begin{minipage}[t]{0.3\textwidth}
    \centering
    \includegraphics[width=\linewidth]{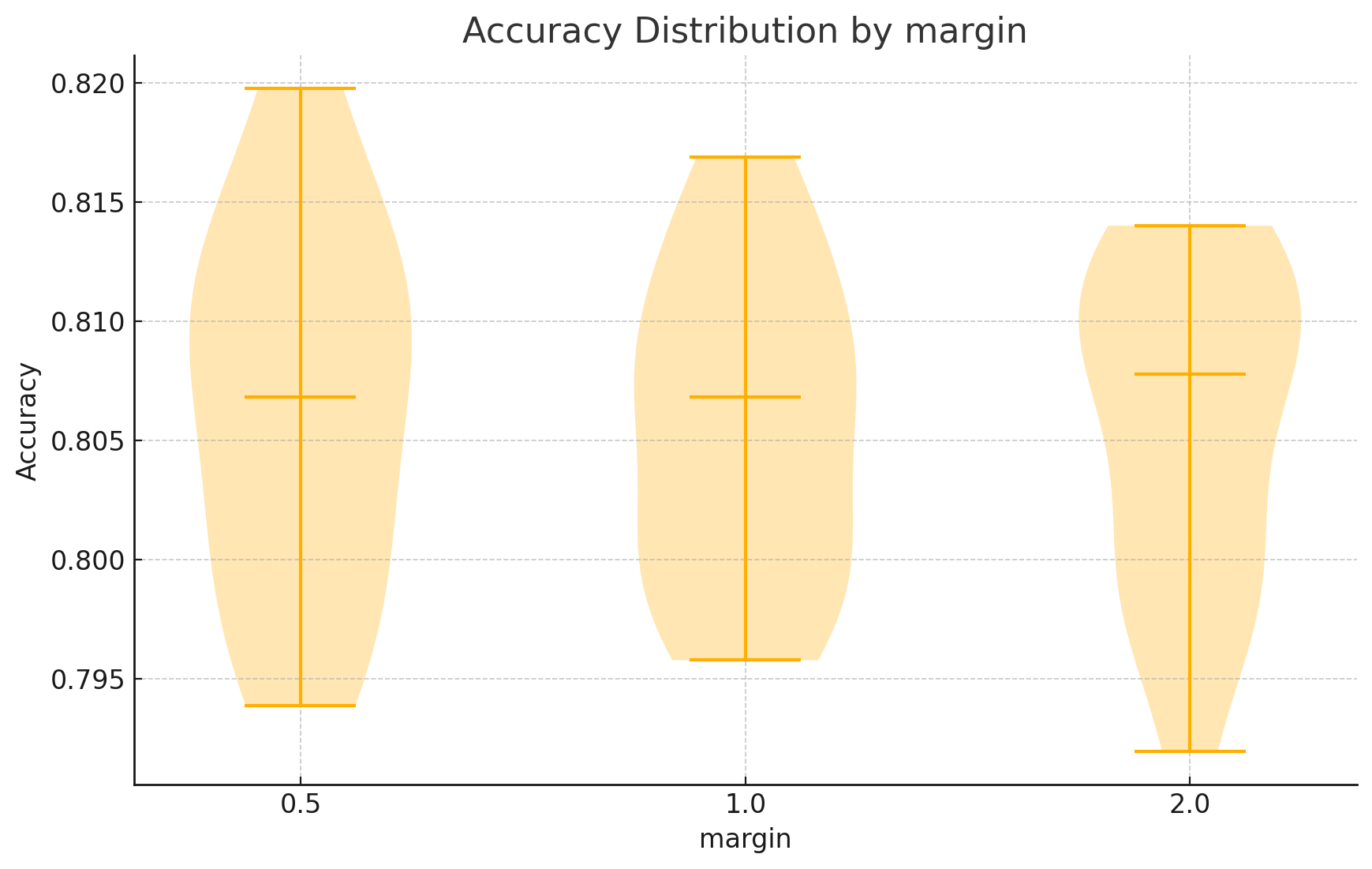}\\[1pt]
    \scriptsize (a) Margin \(m\)
  \end{minipage}\hfill
  \begin{minipage}[t]{0.3\textwidth}
    \centering
    \includegraphics[width=\linewidth]{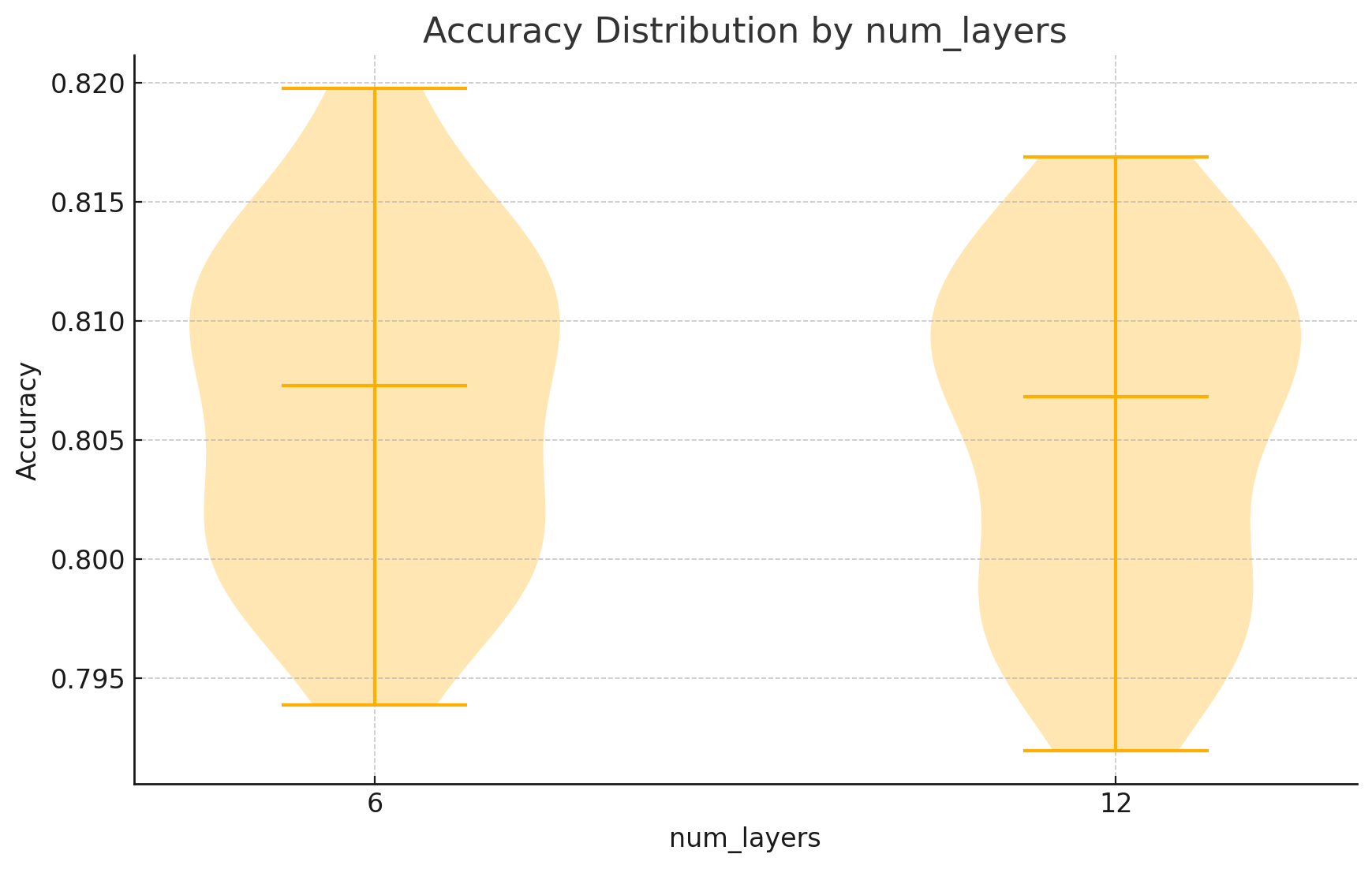}\\[1pt]
    \scriptsize (b) \# Layers
  \end{minipage}\hfill
  \begin{minipage}[t]{0.3\textwidth}
    \centering
    \includegraphics[width=\linewidth]{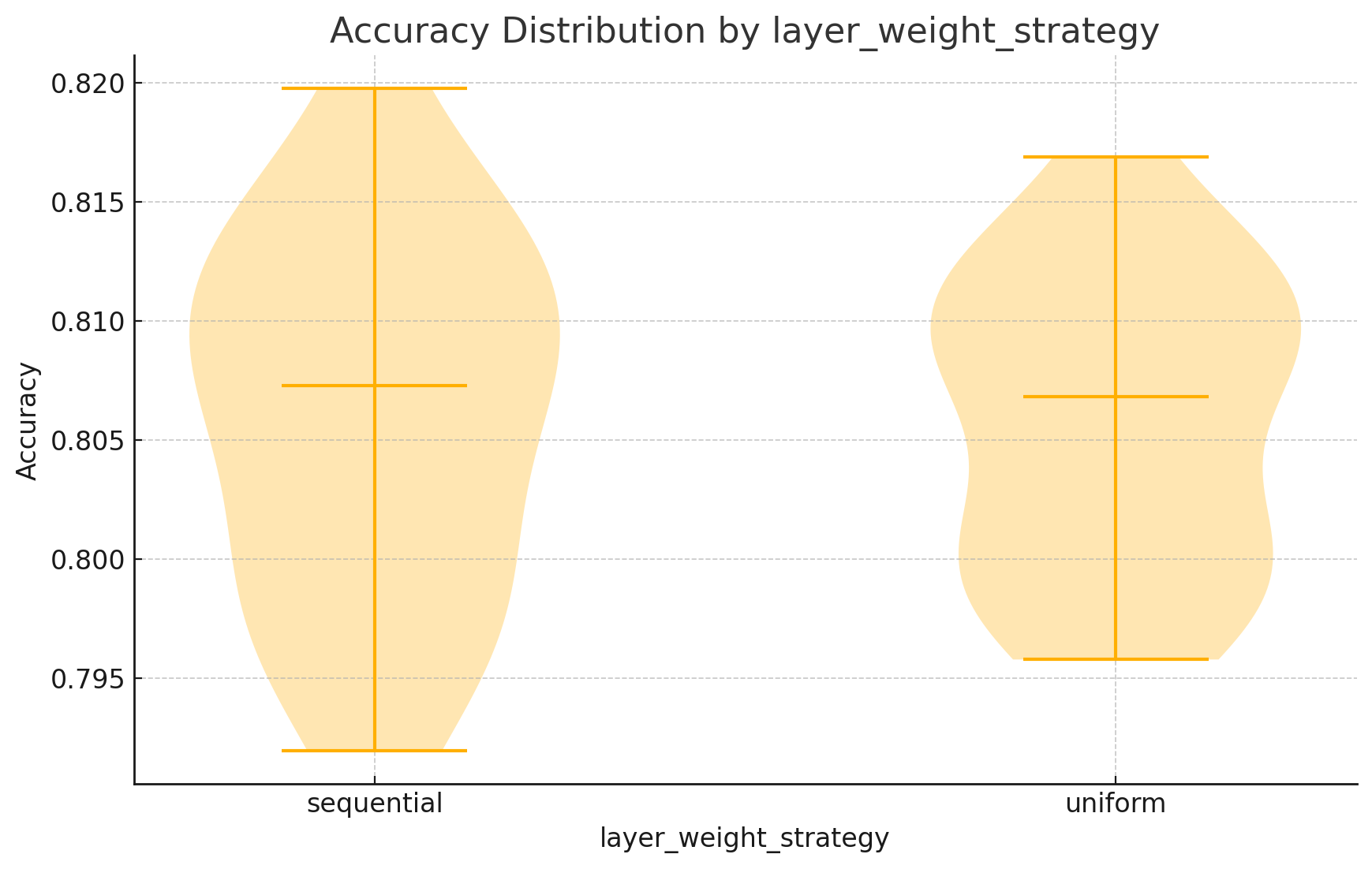}\\[1pt]
    \scriptsize (c) Weighting \(w_\ell\)
  \end{minipage}

  \vspace{3mm}

  \begin{minipage}[t]{0.3\textwidth}
    \centering
    \includegraphics[width=\linewidth]{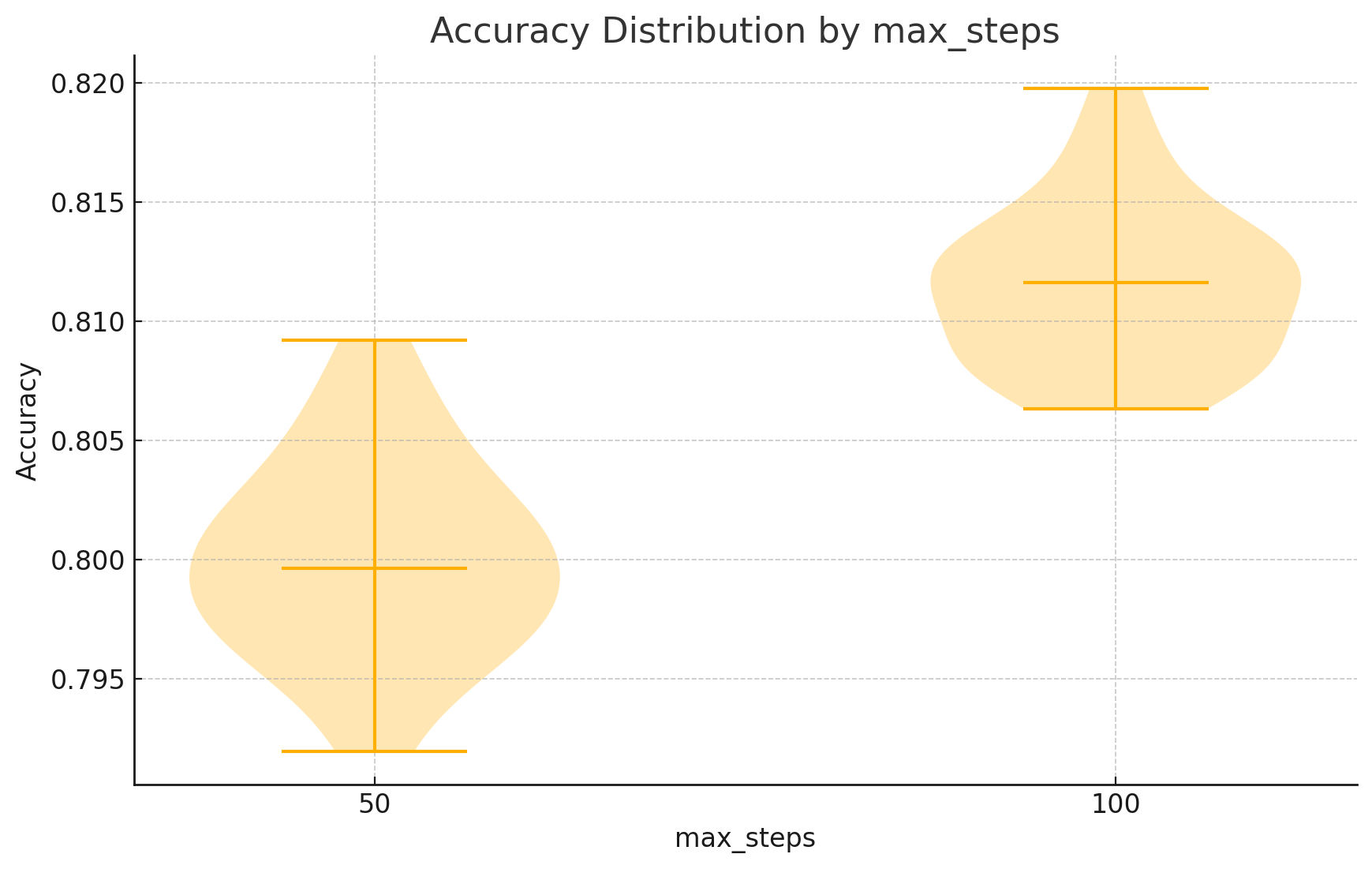}\\[1pt]
    \scriptsize (d) Matching steps \(S\)
  \end{minipage}\hspace{0.03\textwidth}
  \begin{minipage}[t]{0.3\textwidth}
    \centering
    \includegraphics[width=\linewidth]{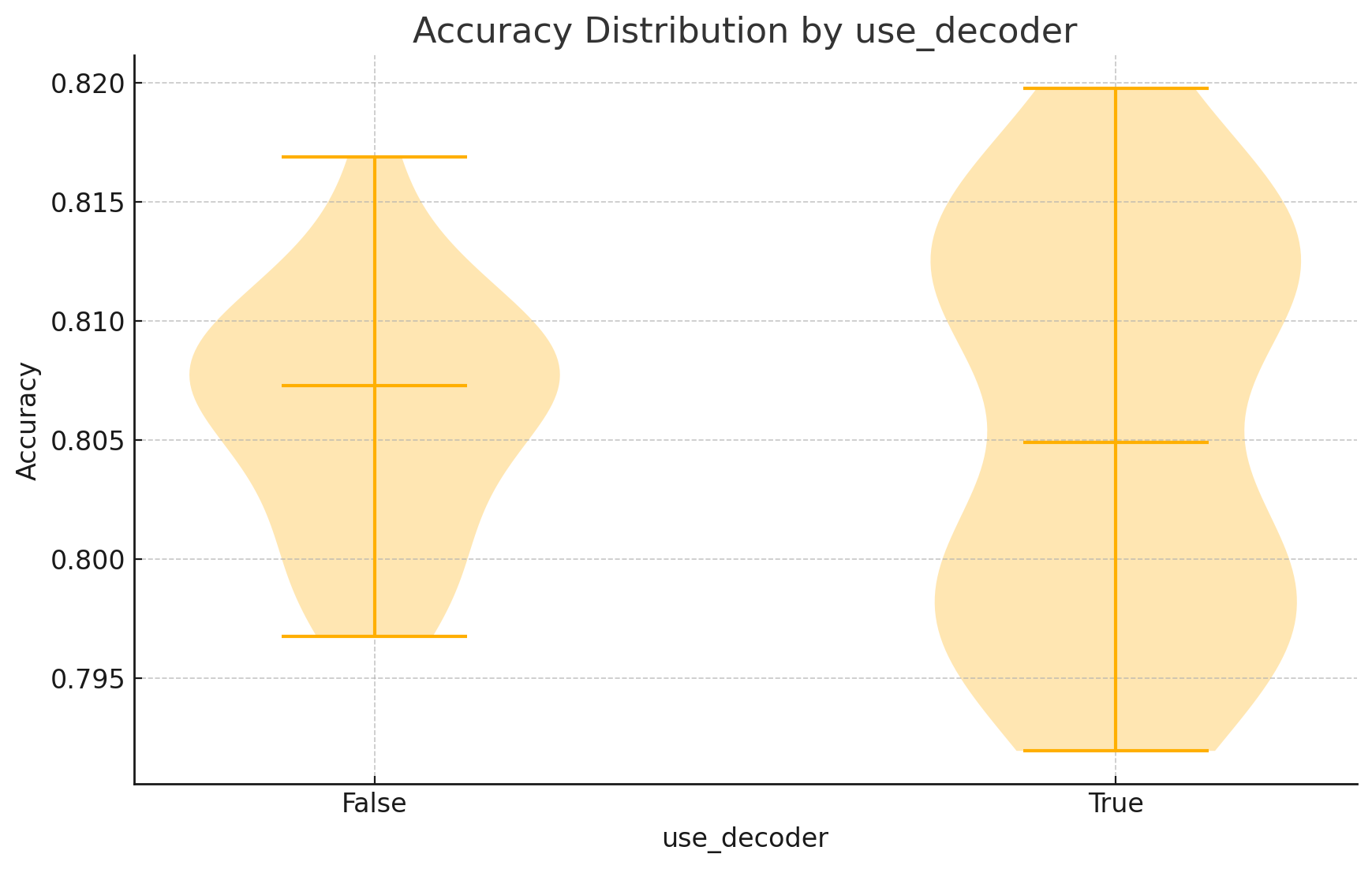}\\[1pt]
    \scriptsize (e) Scope
  \end{minipage}

  \caption{\textbf{Violin plots of accuracy distributions across ablation
  factors.}}
  \label{fig:ablation_violin}
\end{figure*}

\paragraph{Saturation on the easier datasets.}
Flowers102 and Pets reach $99.8$--$100\%$ training accuracy on Swin-T (not
shown), and on both the additional sign agreement bought by a longer matching
schedule slightly increases test loss ($0.2464 \rightarrow 0.2651$ on Pets from
$S{=}100$ to $S{=}1000$ at $m{=}0.5$) while leaving test accuracy unchanged to
within $0.1$ points. Food-101 and FGVC-Aircraft, at $94$--$97\%$ training
accuracy, show no such penalty. The pattern matches ConvNeXt-T, where Pets was
likewise the one dataset with a mildly negative trend, and supports reading the
effect as saturation of the downstream task: once the adapter separates the
training data, a more faithful copy of the reference's boundary structure has
little left to contribute.

\paragraph{Choice of defaults.}
No setting dominates across backbones, and the observed spreads are comparable
to the run-to-run variation expected from a single seed. We therefore do not
tune these hyperparameters per dataset, and adopt $m{=}0.5$ with $S{=}500$
throughout: $m{=}0.5$ is never worst on either backbone, and $S{=}500$ matches
$S{=}1000$ to within $0.2$ points on VTAB-1K and $0.1$ points on Swin-T at half
the matching cost.

\section{Impact of Adapter Rank on T5-base GLUE Benchmark}
\label{app:rank_glue}

Table~\ref{tab:rank_glue} reports performance of LoRA and ABM-LoRA across
adapter ranks $r \in \{8, 16, 32\}$ on five GLUE tasks using T5-base.
Alpha is set to $2r$ throughout.

\paragraph{Consistent gains across all ranks.}
ABM-LoRA outperforms LoRA on four of five tasks at every rank, demonstrating
that activation boundary matching provides a robust advantage regardless of the
adapter capacity; the exception is QNLI at $r{=}32$, discussed below.
The improvements are particularly pronounced on CoLA and MRPC, which are known
to be sensitive to initialization quality due to their smaller dataset sizes
and highly structured linguistic patterns.

\paragraph{Rank efficiency.}
A notable finding is that ABM-LoRA at rank 8 already surpasses LoRA at
rank 32 on CoLA (76.51\% vs.\ 75.84\%) and MRPC (83.09\% vs.\ 82.11\%).
This suggests that ABM initialization compensates for reduced adapter capacity
by aligning the low-rank subspace with the most informative gradient directions
from the outset, effectively making lower-rank adapters more parameter-efficient.

\paragraph{Scaling behavior.}
As rank increases, both methods improve, but ABM-LoRA maintains a consistent
lead. The gap on CoLA narrows from $+6.71$ pp at $r=8$ to $+4.41$ pp at $r=32$,
while on MRPC it narrows more sharply ($+8.09$ pp to $+3.18$ pp).
This indicates that LoRA can partially recover with higher rank,
but ABM-LoRA's initialization advantage persists even as capacity scales up.
On QNLI, both methods converge to near-identical performance across all ranks
(92.77\%--93.03\%), suggesting that this task approaches a performance ceiling
under the given training setup, where initialization strategy has diminishing
influence regardless of adapter capacity.
The slight drop of ABM-LoRA at $r=32$ (92.95\% vs.\ 93.03\%) is within
the range of training variance and does not represent a meaningful difference.

\section{Impact of Adapter Rank on LLaMA2-7B}
\label{app:rank_ablation}

We further examine how adapter rank affects
dialogue performance on LLaMA2-7B, extending
the rank analysis from T5-base in Appendix~\ref{app:rank_glue}
to a larger generative model. With dropout fixed at 0.1, we compare adapter ranks of 8 and 16 across
three evaluation metrics: MT-Bench, length-controlled win rate (LC), and win
rate (WR), under otherwise identical settings.
As shown in Figure~\ref{fig:rank_ablation_full}, blue-toned bars correspond to
rank-8 configurations and red-toned bars to rank-16.

For rank 8, replacing LoRA with ABM-LoRA improves MT-Bench from 5.89
to 5.95, raises LC from 42.16\% to 44.74\%, and increases WR from 46.27\% to
49.19\%.
At rank 16, ABM-LoRA shows a slight MT-Bench drop (5.87 to 5.79), yet still
boosts AlpacaEval metrics: LC rises from 41.37\% to 42.81\% and WR from 46.52\%
to 49.00\%.
This divergence suggests that while activation boundary matching continues to
improve pairwise win-rate metrics, its effect on MT-Bench may depend more
sensitively on rank and Stage~1 hyperparameters.
Therefore, careful adjustment of Stage~1 steps and hyperparameters according
to rank size is necessary to maintain overall performance balance.

\subsection{Ablation Study}
To analyze the impact of Stage~1 ABM initialization, we performed a grid-search
ablation over five factors: (1) margin $m$,
$\{0.5,\,1.0,\,2.0\}$; (2) number of layers, $\{6,\,12\}$\footnote{T5 has 12
encoder and 12 decoder layers; ``6'' refers to the 6 deepest layers in each.};
(3) layer weighting $w_\ell$, sequential vs.\
uniform; (4) matching steps $S \in \{50,\,100\}$; and (5) scope, encoder+decoder
vs.\ encoder only.
We evaluated 48 configurations on the GLUE dev set using ABM-LoRA (T5-base) and
report the average accuracy.
As shown in Figure~\ref{fig:ablation_violin}, the setup with $m{=}0.5$, the six
deepest layers, and sequential weighting yields the highest median accuracy.
Increasing $S$ from 50 to 100 gives modest gains, and adding both encoder and
decoder layers provides a slight boost.
The optimal configuration ($m{=}0.5$, 6 layers, sequential $w_\ell$, $S{=}100$,
encoder+decoder) achieves 81.98\% average accuracy.
Using all layers or applying uniform weighting led to lower accuracy,
indicating that focusing on a subset of layers with gradually increasing
weights provides the most effective initialization.

\end{document}